\documentclass[acmsmall, screen]{acmart}
\usepackage{amsmath,amsthm,mathtools}
\usepackage{booktabs,tabularx,array,longtable}
\usepackage{colortbl}  
\usepackage{pifont}
\usepackage{float}
\usepackage{wrapfig}
\usepackage{subcaption}
\usepackage{algorithm}
\usepackage{algpseudocode}
\algrenewcommand\algorithmicwhile{\textbf{While}}
\algrenewcommand\algorithmicdo{\textbf{Do}}
\algrenewcommand\algorithmicif{\textbf{If}}
\algrenewcommand\algorithmicthen{\textbf{Then}}
\algrenewcommand\algorithmicelse{\textbf{Else}}
\algrenewcommand\algorithmicend{\textbf{End}}
\algrenewcommand\algorithmicfor{\textbf{For}}
\algrenewcommand\algorithmicforall{\textbf{For All}}
\algrenewcommand\algorithmicreturn{\textbf{Return}}
\usepackage{listings}
\usepackage{tikz}
\usetikzlibrary{arrows.meta,positioning,fit,calc,backgrounds}
\usepackage{xspace}
\usepackage{enumitem}
\usepackage{xr}
\usepackage[normalem]{ulem}
\newif\ifproseparagraph
\proseparagraphtrue
\AddToHook{env/algorithmic/begin}{\proseparagraphfalse}
\AddToHook{env/tikzpicture/begin}{\proseparagraphfalse}
\AddToHook{env/tabular/begin}{\proseparagraphfalse}
\AddToHook{env/tabular*/begin}{\proseparagraphfalse}
\AddToHook{para/begin}{%
  \ifproseparagraph
    \parfillskip=0pt plus .4\linewidth\relax
    \finalhyphendemerits=1000000\relax
  \else
    \parfillskip=0pt plus 1fil\relax
  \fi}
\AtBeginDocument{\finalhyphendemerits=1000000\relax}
\ifdefined\issupplement
\fi
\definecolor{structural}{RGB}{45,91,132}
\definecolor{numerical}{RGB}{174,86,20}
\definecolor{lightstruct}{RGB}{233,240,246}
\definecolor{lightnum}{RGB}{253,240,226}
\definecolor{scepgold}{HTML}{E0A91B}  
\newcommand{\sconly}{\nosupport}  
\newcommand{\neither}{\textcolor{black!45}{---}}  
\newcommand{\nopair}{\multicolumn{1}{c}{\sconly}}  
\newcommand{\nopairdiv}{\multicolumn{1}{c?}{\sconly}}  
\newcommand{\nosys}{\multicolumn{1}{c}{\neither}}  
\newcommand{\nosysdiv}{\multicolumn{1}{c?}{\neither}}  
\newcommand{\nopairnotediv}{\multicolumn{1}{c?}{\sconly}}  
\newcolumntype{?}{!{\color{black!30}\vrule width 0.4pt}}  
\newcommand{\grpline}{\arrayrulecolor{black!25}\midrule[0.3pt]\arrayrulecolor{black}}  
\newcommand{\sys}{\textsc{Scepter}\xspace}
\newcommand{\neuton}{\textsc{Neuton}\xspace}
\newcommand{\cf}{\textsc{ConstraintFlow}\xspace}
\newcommand{\gb}{g-BCSR\xspace}

\newcommand{\EReal}{\overline{\mathbb R}}
\newcommand{\R}{\mathbb R}

\newcommand{\pheader}[1]{%
  \par\vspace{0.8ex}%
  \noindent\textbf{#1.}\ %
}
\newtheorem{theorem}{Theorem}[section]
\newtheorem{lemma}[theorem]{Lemma}

\newtheorem{corollary}[theorem]{Corollary}
\theoremstyle{definition}
\newtheorem{definition}[theorem]{Definition}
\usepackage{stmaryrd,multirow}
\graphicspath{{figures/}}
\tikzset{alt/.code={}}
\definecolor{gbcsrcall}{HTML}{CC5500}
\newcommand{\irnode}[1]{\textsf{#1}}
\newcommand{\gbcsr}{\mbox{g-BCSR}\xspace}
\newcommand{\sparse}{\gbcsr}
\newcommand{\compiler}{\neuton}
\newcommand{\scepter}{\sys}
\newcommand{\stblock}{\irnode{StencilBlock}\xspace}
\newcommand{\wblock}{\irnode{WindowBlock}\xspace}
\newcommand{\rblock}{\irnode{BroadcastBlock}\xspace}
\newcommand{\cblock}{\irnode{UniformBlock}\xspace}
\newcommand{\dblock}{\irnode{DenseBlock}\xspace}
\newcommand{\diagblock}{\irnode{DiagonalBlock}\xspace}
\newcommand{\tmul}{\mathbin{@}}
\newcommand{\descr}{\pi}
\newcommand{\payload}{\varrho}
\newcommand{\sem}[1]{\llbracket #1 \rrbracket}
\newcommand{\boxof}{\mathrm{box}}
\newcommand{\supp}{\mathrm{supp}}
\newcommand{\ibp}{IBP\xspace}
\newcommand{\deeppoly}{DeepPoly\xspace}
\newcommand{\deepz}{DeepZ\xspace}
\newcommand{\crownibp}{CROWN-IBP\xspace}

\newcommand{\reusecert}{\textsc{ReuseCert}\xspace}
\newcommand{\zid}{\textsc{ZID}\xspace}
\newcommand{\relu}{ReLU\xspace}
\newcommand{\traverse}{\textnormal{\texttt{traverse}}\xspace}

\newcommand{\CommentRight}[1]{\hfill{$\triangleright$ \textit{#1}}}
\newsavebox{\fitwidthbox}
\newcommand{\fitwidth}[1]{%
  \sbox{\fitwidthbox}{#1}%
  \ifdim\wd\fitwidthbox>\linewidth
    \resizebox{\linewidth}{!}{\usebox{\fitwidthbox}}%
  \else\usebox{\fitwidthbox}\fi}

\newcommand{\true}{\mathsf{true}}
\newcommand{\false}{\mathsf{false}}

\definecolor{diagramcolor}{rgb}{0.0,0.35,0.7}

\newcommand{\skippoly}{\textsc{SkipPoly}\xspace}
\newcommand{\boxpoly}{\textsc{BoxPoly}\xspace}

\newcommand{\nosupport}{\textcolor{red!75!black}{\ding{55}}}

\usepackage[table]{xcolor}

\title{Fast and Scalable Abstract Interpretation for Deep Neural Networks}
\author{Avaljot Singh}
\authornote{Equal Contribution}
\email{avaljot2@illinois.edu}
\affiliation{%
  \institution{University of Illinois Urbana-Champaign}
  \country{USA}
}

\author{Dushyant Bharadwaj}
\authornotemark[1]
\affiliation{%
  \institution{University of Illinois Urbana-Champaign}
  \country{USA}
}
\email{db50@illinois.edu}

\author{Yasmin Chandini Sarita}
\affiliation{%
  \institution{University of Illinois Urbana-Champaign}
  \country{USA}
}
\email{ysarita2@illinois.edu}

\author{Heng Zhong}
\affiliation{%
  \institution{University of Illinois Urbana-Champaign}
  \country{USA}
}
\email{hengz2@illinois.edu}
\author{Charith Mendis}
\affiliation{%
  \institution{University of Illinois Urbana-Champaign}
  \country{USA}
}
\email{charithm@illinois.edu}
\author{Gagandeep Singh}
\affiliation{%
  \institution{University of Illinois Urbana-Champaign}
  \country{USA}
}
\email{ggnds@illinois.edu}

\begin{document}
\begin{abstract}
Abstract interpretation can certify properties of Deep Neural Networks
(DNNs), but precise analyses remain expensive. General implementations
repeat work when analyzing the same architecture under different inputs
and properties. Interleaving this work with numerical operations limits
GPU utilization and restricts how much of the analysis a tensor compiler
can process together.

We present \sys, a framework for specializing DNN abstract interpreters
through \emph{structural semantics}. The semantics determines the
structure of intermediate abstract values while deferring their numerical
evaluation. Structural computations from successive operations can
therefore be performed together before certification, generating an
executable tensor program for the numerical computation. With the
interpreter, architecture, and input and batch shapes fixed, the program
can be reused with different weights and input regions. We prove that
it computes the same bounds as the general interpreter in exact
arithmetic. The generated program also exposes operations and their
dependencies for tensor compilation and memory planning. We implement
\sys for the tensor programs that \neuton generates from \cf
specifications.

We evaluate \sys on eight abstract interpreters and 44 DNNs against
six baseline systems on GPUs and CPUs. Across the evaluated analyses,
\sys handles networks with up to $153\times$ as many neurons as the
largest completed by the corresponding baselines. It completes
\deeppoly on ResNet50 and VGG-16, whereas none of the evaluated
\deeppoly baselines completes either network. On configurations completed
by both systems, GPU geometric-mean speedups over \neuton range from
$11.9\times$ to $44.4\times$ across eight analyses, with a
$2.60\times$ geometric-mean reduction in peak memory. In an input-splitting
case study, \sys proves 68 properties for which the baselines
time out.

Our code is available at \url{https://github.com/uiuc-focal-lab/constraintflow.git}
\end{abstract}
\maketitle
\section{Introduction}\label{sec:intro}
Abstract interpretation~\cite{AI} is widely used for formal certification
of Deep Neural Networks (DNNs). A DNN abstract interpreter propagates
constraints on neuron values to establish properties over large,
potentially infinite sets of inputs. As for programs, there is a tradeoff between the cost and precision of a DNN abstract interpreter.  Interval Bound Propagation (\ibp) maintain
independent bounds on each neuron, whereas polyhedral analyzers such as  \deepz~\cite{deepz} and
\deeppoly~\cite{deeppoly} capture relationships between neurons. These
relationships can establish properties that intervals alone cannot.
However, maintaining and combining them requires substantial computation
and memory. On larger networks, users may therefore have to resort to
simpler analyses that lack the precision needed to prove the desired
properties. Making precise analyses practical at scale remains a
central challenge in DNN certification.

Substantial research and engineering effort has made DNN abstract
interpreters increasingly efficient, with hand-written implementations
such as auto\_LiRPA~\cite{lirpagithub}, IVAN~\cite{incremental1}, LUNA~\cite{luna} and
GPUPoly~\cite{gpupoly}, and frameworks such as \neuton~\cite{neuton}.
These implementations exploit the dependencies encoded in abstract
values to avoid unnecessary storage and computation. However, these
dependencies evolve as the abstract interpreter combines and refines bounds.
The implementation must therefore determine how each intermediate
abstract value is computed from the preceding ones.
This introduces work beyond the numerical evaluation of the bounds.
General implementations perform both kinds of work together and repeat
the decisions about these relationships during every analysis. Further, when the
same DNN is certified for different properties or the subregions produced
by input splitting, much of this non-numerical work remains unchanged, although the numerical bounds differ.
The challenge is to scale precise abstract interpreters to larger DNNs
without repeatedly computing information that can be determined from the
network architecture.

\textbf{This work.}
We present \sys, a framework that scales precise DNN abstract interpreters
to networks beyond the reach of existing implementations. \sys
specializes the interpreter to a given architecture, resolving the
computation determined by that architecture ahead of certification and
generating executable certifier code for that architecture. For instance, \sys executes \deeppoly~\cite{deeppoly}, one of the most widely
used DNN certifiers, on ResNet50~\cite{robustness} with 2.94 million
neurons and VGG-16~\cite{vgg16-vnncomp} with 13.6 million neurons.
None of the compared \deeppoly baselines completes either network. The largest
network in our suite completed by any of them is ResNet34 with about
968K neurons. \sys
therefore executes the same analysis on networks with $3.0\times$ and
$14\times$ as many neurons, respectively. Across other networks and
certifiers in our evaluation, we observe runtime speedups of one to two
orders of magnitude (\S\ref{sec:spec-eval}).

The key insight is that, for a fixed abstract interpreter, the structure
of intermediate abstract values often depends on the DNN architecture
rather than the weights or input regions. We call the work that determines
these patterns and the operations on them structural computation. The
coefficients and bounds change with the weights and input regions, but
this structural computation can be reused. However,
separating the two computations is nontrivial because the abstract interpreter
interleaves them throughout its execution. Each operation can change
the structure of the abstract values used by subsequent operations.
We must determine how these intermediate structures evolve without
computing their numerical values, identify which structural computations
can be performed in advance, and bring them together before numerical
execution.

We develop novel \emph{structural semantics} for DNN abstract interpreters
to perform this separation. The semantics makes the structural information
in abstract values explicit and defines how each operation transforms
this information. We define the structural semantics for composable operations that hold across abstract domains and abstract transformers. Its rules derive each result's structure from the
structures of its inputs, while recording code required for the numerical
computation. We also
reformulate coefficient-dependent control so that structural execution
can proceed through the abstract interpreter without knowing the bounds.
Executing these rules for a fixed DNN architecture brings together the
structural computation before numerical evaluation. The output is
executable certifier code specialized to that architecture, reusable with
different weights, input regions, and perturbation specification. We prove that
specialization preserves the general interpreter's bounds. Decisions that still require numerical values remain in the
generated code, with the computations they invoke specialized separately.

This separation is particularly useful on GPUs. Numerical operations
can execute quickly, making the intervening CPU work a bottleneck.
When a CPU decision reads a GPU result, the next GPU operation must
wait for that decision even if transferring the result takes negligible
time. Specialization also exposes larger
sequences of numerical operations to a tensor compiler, allowing it to
combine operations and eliminate intermediate storage. We further use
the dependencies in the generated program to process independent groups
of neurons separately and release their large intermediate tensors
earlier.

The structural patterns captured by our semantics arise across DNN
certifier implementations. For example, auto\_LiRPA uses
convolutional patches~\cite{lirpagithub}, IVAN applies diagonal ReLU
transformations through neuronwise coefficients~\cite{incremental1},
and GPUPoly stores coefficients within dependence sets determined by
the network architecture~\cite{gpupoly}. These implementations use
different data structures to exploit similar structural patterns,
combining computation on these structures with arithmetic on coefficient
values. The generalized block compressed sparse representation (\gb),
introduced by \neuton~\cite{neuton}, provides a unified representation
of these patterns. We build our structural semantics on this
representation and implement \sys for the tensor programs generated by
\neuton from \cf~\cite{constraintflowsas} specifications, providing one
specialization procedure for different abstract domains and
certification algorithms.

\begin{wrapfigure}{r}{0.48\linewidth}
    \centering
    \includegraphics[width=\linewidth]{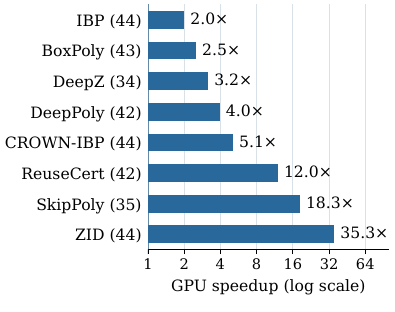}
    \caption[GPU speedup over the fastest evaluated baseline]{\sys's
        geometric-mean GPU speedup over the fastest evaluated baseline on
        each network. Parentheses indicate the number of networks completed
        by at least one baseline. }
    \Description{Eight horizontal blue bars show geometric-mean GPU speedups
        of 2.0 for IBP, 2.5 for BoxPoly, 3.2 for DeepZ, 4.0 for DeepPoly,
        5.1 for CROWN-IBP, 12.0 for ReuseCert, 18.3 for SkipPoly, and 35.3
        for ZID. The baseline is selected separately for each network.
        The axis is logarithmic and network counts appear beside the labels.}
    \label{fig:intro-best-baseline}
\end{wrapfigure}
We evaluate \sys on eight abstract interpreters and 44 DNNs against
six baseline systems on GPUs and CPUs. Across the analyses, \sys
handles networks with up to $153\times$ as many neurons as the largest
completed by the corresponding baselines, and completes 20
certifier--network pairs that no corresponding baseline completes.
On GPUs, speedups over \neuton range from $11.9\times$ to $44.4\times$
across all eight analyses. For \deeppoly, \sys is
$5.69\times$ faster than auto\_LiRPA, $6.57\times$ faster than GPUPoly,
and $22.3\times$ faster than IVAN. The gains also extend to CPU
execution, where \sys is $87.4\times$ faster than ERAN for \deeppoly
and $522\times$ faster for \deepz.
\sys also reduces peak GPU memory by a factor of $2.60\times$ relative
to \neuton over shared configurations. The runtime gains have a direct
benefit for input splitting, where the same network is repeatedly
analyzed over smaller regions. Using \sys and auto\_LiRPA as \deeppoly
backends for the same splitting procedure, \sys proves 68 properties
within a 25-second budget per instance for which the auto\_LiRPA backend
times out.
Fig.~\ref{fig:intro-best-baseline} shows \sys's geometric-mean GPU speedup
for each abstract interpreter. We divide the fastest evaluated baseline's
runtime by \sys's runtime separately on each network, then take the
geometric mean over networks where at least one baseline completes.

\textbf{Main contributions.}
\begin{itemize}[noitemsep, nolistsep]
    \item We develop novel structural semantics for DNN abstract interpreters
    that separates structural computation from numerical evaluation
    and generates executable certifier code specialized to a
    fixed architecture. We prove that specialization preserves the original
    output.
    \item We implement these semantics in \sys for abstract interpreters
    generated by \neuton from \cf specifications. \sys provides a common
    specialization procedure for different abstract domains and certification
    algorithms, and exposes the generated numerical code to tensor
    compilation and memory planning.
    \item We evaluate \sys on eight abstract interpreters and 44 DNNs.
    \sys completes 20 certifier--network pairs that no corresponding baseline
    completes and enables analyses to process networks with up to $153\times$
    as many neurons as the largest completed by their baselines. We also
    observe runtime speedups of one to two orders of magnitude.
\end{itemize}

\section{Background}\label{sec:background}
We introduce DNN abstract interpreters, tensor computation, and the
\sparse representation.

\pheader{DNNs and Certification}\label{sec:background-cert}
A feedforward DNN is a Directed Acyclic Graph (DAG) whose nodes are
\emph{neurons}. Each neuron applies an operation to the values of its
predecessors. For example, an affine neuron computes
$n=b+\sum_j w_j n_j$, and a ReLU neuron computes $n=\max(0,n_1)$.
Neurons are grouped into layers. Fully connected and convolutional layers
are both affine layers: their outputs are weighted sums of inputs,
possibly with a bias. Activation layers apply functions such as ReLU.
The \emph{architecture} specifies these operations, their connectivity,
and their dimensions. Learned weights and biases supply their numerical
parameters. A connection remains in the architecture even when its weight is zero.

A certification problem consists of a DNN $\mathcal N$, an input region
$\mathcal X$, and an output property $\psi$, represented as a set of
permitted outputs. The goal is to establish
$\mathcal N(\mathcal X)\subseteq\psi$. For local robustness, every input in a perturbation region around
$\mathbf x_0$ must receive the same classification $t$. 

\pheader{Abstract Values and Tensor Computation}
\label{sec:background-abstract}
Abstract interpretation~\cite{AI} reasons about possibly infinite sets of
neuron values using \emph{abstract values}. An \emph{abstract transformer}
computes an abstract value for an operation's outputs from those of its
inputs. A sound transformer over-approximates the operation, retaining
every result possible for the input abstraction. Starting from an
abstraction containing $\mathcal X$, an abstract interpreter therefore
computes an output abstraction $a_{\mathrm{out}}$ such that
$\mathcal N(\mathcal X)\subseteq\gamma(a_{\mathrm{out}})$,
where $\gamma$ maps an abstract value to the concrete values it represents.
The property is certified if $\gamma(a_{\mathrm{out}})\subseteq\psi$.
Otherwise, the approximation may be too coarse to establish the property.

Abstract interpreters use different representations for these constraints.
Intervals retain a lower and upper bound for each neuron independently.
\deeppoly~\cite{deeppoly} uses affine lower and upper bounds that relate
neurons to other neurons in the network. \deepz~\cite{deepz} uses
zonotopes, whose shared symbolic variables capture relationships between
neuron values. These representations lead to different numerical
computations and different storage requirements.

For example, consider affine lower bounds for the network's neurons $n_p$:
$n_p\ \geq\ b_p+\sum_q L[p,q]n_q$, where coefficients form a tensor $L$, with rows and columns indexed by
network neurons. Constants are stored separately.
Operations on these bounds become tensor operations on their coefficients.
If a bound does not refer to a neuron, its coefficient is zero. For a
convolution, each output refers to a local window of inputs, and the same
weights are used at different positions. So, coefficient tensors contain
both sparse regions and repeated values.

\pheader{Existing Specification and Execution Frameworks}
\cf~\cite{constraintflowsas} is a domain-specific language for specifying
abstract interpreters at the neuron level. A specification describes
abstract values, transformers, and how they are applied across a DNN.
\neuton~\cite{neuton} translates these specifications into tensor programs
that process entire layers. We call such an executable a \emph{general
program}: its code supports different DNN architectures and determines
the required tensor operations from the DNN supplied at run time.

\pheader{g-BCSR Tensors}\label{sec:background-gbcsr}
\neuton introduced the generalized block compressed sparse representation
(\sparse) for the tensors used by these programs~\cite{neuton}. A
\sparse tensor is a tuple $\mathcal T=(d,S,B,Z)$, where $d$ is the background
value, $Z$ is its shape, and $B=\{B_k\}_{k=1}^m$ is a list of blocks at
start indices $S=\{s_k\}_{k=1}^m$. The blocks occupy pairwise disjoint
boxes. The dense tensor represented by $\mathcal T$ is
\[
  \sem{\mathcal T}[\mathbf i]=
  \begin{cases}
    \sem{B_k}[\mathbf i-s_k], & \mathbf i\text{ lies in the box of }B_k,\\
    d, & \mathbf i\text{ lies in no block}.
  \end{cases}
\]
Here $\sem{B_k}$ denotes the dense block represented by $B_k$.
The background can be zero, or a value such as $+\infty$ or $-\infty$
used to initialize bounds. A block's type determines how its core is
interpreted. For example, a \stblock stores the kernel weights shared
across convolution windows, a \diagblock stores diagonal entries, and a
\wblock stores coefficients within windows without requiring them to be
shared. The represented tensor can therefore be much larger than the
stored numerical data.
\neuton proved that operations on \sparse match their dense counterparts.
\S\ref{sec:gb-format} refines the block model to separate structure from
numerical values and justify specialization.

\section{Overview}\label{sec:spec-overview}
We illustrate \sys on a small convolutional network, describing the
computation of output bounds, its execution using \sparse, and its
specialization.

\subsection{Computing Output Bounds}\label{sec:overview-computation}

\makeatletter\@nobreakfalse\everypar{}\makeatother
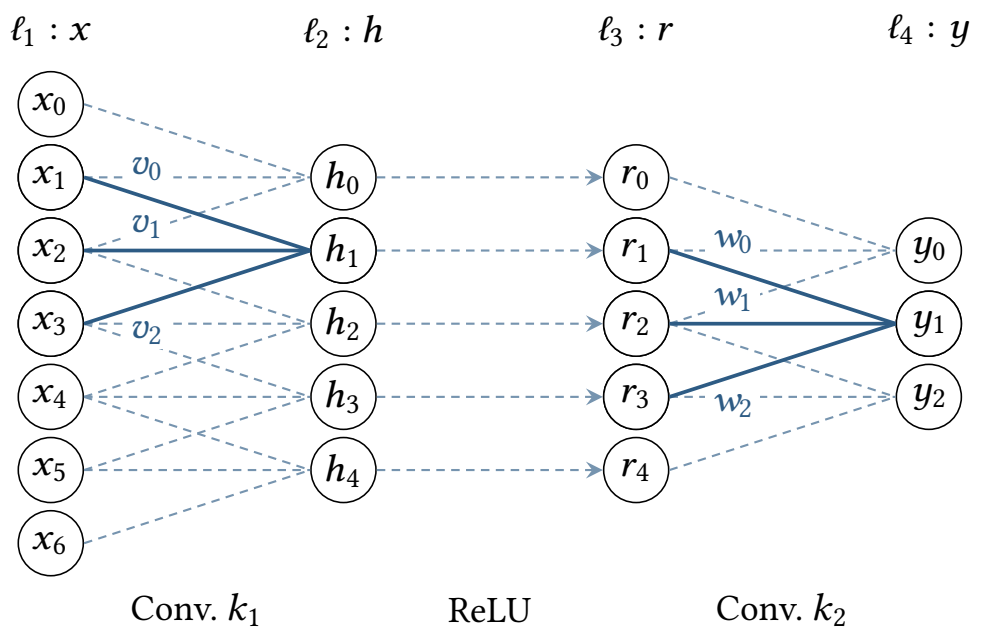
\begin{wrapfigure}[16]{r}{0.48\linewidth}
    \centering
    \fitwidth{
\begin{tikzpicture}[text=black,x=.88cm,y=.88cm,
    alt={Four columns contain seven input neurons x, five convolution
      neurons h, five ReLU neurons r, and three output neurons y.
      Light blue dashed edges show the other connections. Three solid blue edges labelled v0, v1,
      and v2 connect one input window to h1. Three solid blue edges labelled
      w0, w1, and w2 connect one ReLU window to y1. Both convolution
      kernels have shape one by one by three, stride one, and no padding.},
    neuron/.style={circle,draw=black,fill=white,minimum size=4.3mm,
      inner sep=0pt,font=\scriptsize,line width=.3pt},
    wire/.style={draw=structural!65,line width=.4pt,
      dash pattern=on 1.6pt off 1.2pt},
    emphwire/.style={draw=structural,line width=.7pt,solid},
    layer/.style={font=\scriptsize,align=center},
    op/.style={font=\scriptsize,align=center},
    note/.style={font=\scriptsize,text=black,align=center}]
  \def\pitch{0.55}
  \foreach \i in {0,...,6}
    \node[neuron] (x\i) at (0.5,{-\i*\pitch}) {$x_{\i}$};
  \foreach \i in {0,...,4}{
    \node[neuron] (h\i) at (2.7,{-0.55-\i*\pitch}) {$h_{\i}$};
    \node[neuron] (r\i) at (4.9,{-0.55-\i*\pitch}) {$r_{\i}$};}
  \foreach \i in {0,...,2}
    \node[neuron] (y\i) at (7.1,{-1.1-\i*\pitch}) {$y_{\i}$};
  \foreach \j in {0,...,4}{
    \foreach \t in {0,...,2}{
      \pgfmathtruncatemacro{\xi}{\j+\t}
      \ifnum\j=1\else\draw[wire] (x\xi.east) -- (h\j.west);\fi}
    \draw[wire,->,>=stealth] (h\j.east) -- (r\j.west);}
  \foreach \j in {0,...,2}
    \foreach \t in {0,...,2}{
      \pgfmathtruncatemacro{\ri}{\j+\t}
      \ifnum\j=1\else\draw[wire] (r\ri.east) -- (y\j.west);\fi}
  \foreach \i in {1,2,3}
    \node[neuron,draw=black,fill=white] at (x\i) {$x_{\i}$};
  \node[neuron,draw=black,fill=white] at (h1) {$h_1$};
  \draw[emphwire] (x1.east) -- (h1.west)
    node[pos=.28,above=2pt,font=\scriptsize,text=structural,fill=white,inner sep=1pt] {$v_0$};
  \draw[emphwire] (x2.east) -- (h1.west)
    node[pos=.28,above=1pt,font=\scriptsize,text=structural,fill=white,inner sep=1pt] {$v_1$};
  \draw[emphwire] (x3.east) -- (h1.west)
    node[pos=.28,below=2pt,font=\scriptsize,text=structural,fill=white,inner sep=1pt] {$v_2$};
  \foreach \i in {1,2,3}
    \node[neuron,draw=black,fill=white] at (r\i) {$r_{\i}$};
  \node[neuron,draw=black,fill=white] at (y1) {$y_1$};
  \draw[emphwire] (r1.east) -- (y1.west)
    node[pos=.28,above=2pt,font=\scriptsize,text=structural,fill=white,inner sep=1pt] {$w_0$};
  \draw[emphwire] (r2.east) -- (y1.west)
    node[pos=.28,above=1pt,font=\scriptsize,text=structural,fill=white,inner sep=1pt] {$w_1$};
  \draw[emphwire] (r3.east) -- (y1.west)
    node[pos=.28,below=2pt,font=\scriptsize,text=structural,fill=white,inner sep=1pt] {$w_2$};
  \node[layer] at (0.5,0.55) {$\ell_1:x$};
  \node[layer] at (2.7,0.55) {$\ell_2:h$};
  \node[layer] at (4.9,0.55) {$\ell_3:r$};
  \node[layer] at (7.1,0.55) {$\ell_4:y$};
  \node[op] at (1.6,-3.8) {Conv. $k_1$};
  \node[op] at (3.8,-3.8) {ReLU};
  \node[op] at (6.0,-3.8) {Conv. $k_2$};
\end{tikzpicture}}
    \caption[The running network]{The running network. Solid blue edges
        mark one window per convolution. Other connections are dashed.}
    \Description{A network with seven input neurons, five first-convolution
        neurons, five ReLU neurons, and three output neurons. Solid dark
        blue edges identify one three-input window in each convolution.
        Other connections use lighter blue dashed edges.}
    \label{fig:gb-network}
\end{wrapfigure}

\pheader{The running example}
Consider the network in Fig.~\ref{fig:gb-network}: a convolution, a ReLU,
and a second convolution. Its layers $1$, $2$, $3$, and $4$ contain the neurons
$x=(x_0,\ldots,x_6)$, $h=(h_0,\ldots,h_4)$,
$r=(r_0,\ldots,r_4)$, and $y=(y_0,y_1,y_2)$, respectively. We use
one-dimensional convolutions, so each layer is a row of values. Both
convolutions have one input channel and one output channel. Their kernels
$k_1,k_2\in\R^{1\times1\times3}$ have three weights each, where the
dimensions are output channels, input channels, and kernel width. Write
these weights as $(v_0,v_1,v_2)$ and $(w_0,w_1,w_2)$. With stride $1$,
no padding, and biases omitted, the network computes
  $h_j=\sum_{t=0}^{2}v_t x_{j+t}$,
  $r_j=\max(0,h_j)$,
  $y_a=\sum_{t=0}^{2}w_t r_{a+t}$,
where $j\in\{0,\ldots,4\}$ and $a\in\{0,1,2\}$.

\pheader{Bounds as coefficient tensors}
A common computation in DNN abstract interpretation is to express bounds
on output neurons in terms of input neurons. The input region can then be
used to obtain numerical output bounds. We illustrate this computation
using affine lower bounds.
Similar bound computation arises in several abstract interpreters.
\deeppoly and \crownibp use affine bounds over network
neurons~\cite{deeppoly,crownibp}. \deepz instead expresses neuron values
using shared noise variables~\cite{deepz}. Its affine transformers also
update coefficient tensors through matrix multiplication, although the
coefficient columns denote noise variables.
With arbitrary weight signs, polyhedral interpreters use numerical tensor
operations to select lower or upper bounds when substituting neuron
terms. We assume non-negative weights in the second convolution to
illustrate the resulting tensor computations using only lower bounds.

Suppose the abstract interpreter has chosen slopes $\lambda_j$ such that
$\lambda_j h_j$ is a lower bound for the ReLU output $r_j$. Combining
these bounds with the affine layers gives $y_a\ \geq\ \sum_{t=0}^{2}\sum_{q=0}^{2}
          w_t\lambda_{a+t}v_q x_{a+t+q}.$
Tensor implementations compute such lower bounds through matrix
multiplications. We write $L_i^j\in\R^{m_i\times m_j}$ for the
coefficients of lower bounds on layer $i$ in terms of layer $j$, where
$m_i$ is the number of neurons in layer $i$. Thus the subscript identifies
the layer being bounded, and the superscript identifies the layer of the neurons used in the bounds. In our example, $L_2^1\in\R^{5\times7}$ and
$L_4^3\in\R^{3\times5}$ contain the convolution coefficients, while
$L_3^2=\operatorname{diag}(\lambda_0,\ldots,\lambda_4)$ contains the ReLU
bound coefficients. Constants are stored separately and omitted here.
Starting with $L_4^3$, we compute the output bounds in terms of
successively earlier layers using $L_4^{i-1}=L_4^i\tmul L_i^{i-1}$
for $i=3,2$. Thus we first compute $L_4^2=L_4^3\tmul L_3^2$ over
layer $2$, then $L_4^1=L_4^2\tmul L_2^1$ over the input layer.
The final block $L_4^1$ contains the coefficients relating the output bounds to the input neurons, $ w_t\lambda_{a+t}v_q$.

The tensor program performs these computations on whole sparse tensors.
The full coefficient tensor $L_i\in\R^{m_i\times n}$ stores the
coefficients of lower polyhedral bounds on layer $i$ and has columns for
\emph{all} $n$ neurons in the network ($n=20$ here). Its block $L_i^j$
occupies the columns for layer $j$. Columns not used by the bounds contain
zeros.
Stacking the initial lower-bound tensors for all layers gives the
network-wide tensor $L\in\R^{n\times n}$, which remains fixed during
this computation.

At the step for layer $i$, the current output-layer tensor $L_4$ contains
$L_4^i$ in the columns for layer $i$, and zeros elsewhere. The whole-tensor
call $L_4\tmul L$ selects the block product $L_4^i\tmul L_i^{i-1}$ and
returns a tensor containing $L_4^{i-1}$ in the columns for layer $i-1$.
The result is used at the next step.
Thus the recurrence above
describes the matrix multiplications that compute the changing
coefficient block, while $L_4$ denotes the full tensor that stores it.
Each operation has operand shapes $3\times n$ and $n\times n$.
The sparse representation stores the blocks directly without allocating
the dense $n\times n$ tensor.

\subsection{Interleaved Computation}\label{sec:overview-problem}
\begin{figure}[t]
    \centering
    \begin{subfigure}[t]{0.40\linewidth}
        \centering
        \vspace{0pt}
        \fitwidth{
\begin{tikzpicture}[x=.88cm,y=.88cm,
    gridline/.style={draw=structural!30,line width=.16pt},
    bandline/.style={draw=structural,line width=.3pt},
    blockbox/.style={draw=structural,line width=.4pt,densely dashed},
    frame/.style={draw=black,line width=.35pt},
    note/.style={font=\scriptsize,align=left,text=black},
    callout/.style={draw=black,line width=.3pt,->,>=stealth}]
  \definecolor{boundconvsecond}{RGB}{170,55,130}
  \def\cell{.22}
  \def\inputcell{.065}
  \def\inputheight{.455}
  \def\gridheight{3.315}
  \node[font=\small] at (2.2,.63) {$L\in\mathbb R^{20\times20}$};

  \foreach \v in {0,...,20}{
    \draw[gridline] (\v*\cell,0) -- (\v*\cell,-\gridheight);
    \pgfmathsetmacro{\rowy}{-min(\v,7)*\inputcell-max(\v-7,0)*\cell}
    \draw[gridline] (0,\rowy) -- (4.4,\rowy);}

  \foreach \row in {0,...,4}
    \foreach \tap/\shade in {0/25,1/55,2/90}{
      \fill[structural!\shade]
        ({(\row+\tap)*\cell},{-\inputheight-\row*\cell}) rectangle ++(\cell,-\cell);
      \draw[gridline]
        ({(\row+\tap)*\cell},{-\inputheight-\row*\cell}) rectangle ++(\cell,-\cell);}
  \foreach \row/\shade in {0/30,1/45,2/60,3/75,4/90}{
    \fill[teal!\shade] ({(7+\row)*\cell},{-\inputheight-(5+\row)*\cell}) rectangle ++(\cell,-\cell);
    \draw[gridline] ({(7+\row)*\cell},{-\inputheight-(5+\row)*\cell}) rectangle ++(\cell,-\cell);}
  \foreach \row in {0,...,2}
    \foreach \tap/\shade in {0/25,1/55,2/90}{
      \fill[boundconvsecond!\shade]
        ({(12+\row+\tap)*\cell},{-\inputheight-(10+\row)*\cell}) rectangle ++(\cell,-\cell);
      \draw[gridline]
        ({(12+\row+\tap)*\cell},{-\inputheight-(10+\row)*\cell}) rectangle ++(\cell,-\cell);}
  \draw[frame] (0,0) rectangle (4.4,-\gridheight);
  \foreach \v in {7,12,17}{
    \draw[bandline] (\v*\cell,0) -- (\v*\cell,-\gridheight);
    \pgfmathsetmacro{\rowy}{-\inputheight-(\v-7)*\cell}
    \draw[bandline] (0,\rowy) -- (4.4,\rowy);}
  \foreach \mid/\name in {3.5/x,9.5/h,14.5/r,18.5/y}{
    \node[font=\scriptsize,anchor=south] at (\mid*\cell,.08) {$\name$};
    \pgfmathsetmacro{\rowy}{-min(\mid,7)*\inputcell-max(\mid-7,0)*\cell}
    \node[font=\scriptsize,anchor=east] at (-.10,\rowy) {$\name$};}
  \draw[blockbox] (0,-.455) rectangle (1.54,-1.555);
  \draw[blockbox] (1.54,-1.555) rectangle (2.64,-2.655);
  \draw[blockbox,draw=boundconvsecond] (2.64,-2.655) rectangle (3.74,-3.315);
  \node[note,fill=white,inner sep=2pt] (convone) at (2.80,-.965)
    {$L_2^1$ : $5\times7$};
  \draw[callout] (convone.west) -- (1.57,-.965);
  \node[note,fill=white,inner sep=2pt,align=center] (relu) at (3.57,-2.085)
    {$L_3^2$\\$5\times5$};
  \draw[callout] (relu.west) -- (2.67,-2.085);
  \node[note,fill=white,inner sep=1pt,align=center] (convtwo) at (1.27,-2.965)
    {$L_4^3$ : $3\times5$};
  \draw[callout] (convtwo.east) -- (2.61,-2.965);

  \draw[blockbox] (.05,-3.805) rectangle (.38,-3.585);
  \node[note,anchor=west] at (.46,-3.695) {Stored block};
  \fill[structural!55] (2.25,-3.805) rectangle (2.58,-3.585);
  \node[note,anchor=west] at (2.66,-3.695) {Possible nonzero};
\end{tikzpicture}}
        \caption{Initial layer bounds in $L$.}
        \Description{A twenty-by-twenty tensor with blocks for the bounds
            on layer two over one, layer three over two, and layer four
            over three. The seven unused input-neuron rows form a vertically
            compressed, unfilled grid.}
        \label{fig:gb-bound-tensor}
    \end{subfigure}\hfill
    \begin{subfigure}[t]{0.58\linewidth}
        \centering
        \vspace{0pt}
        \fitwidth{
\begin{tikzpicture}[x=.88cm,y=.88cm,
    gridline/.style={draw=structural!30,line width=.16pt},
    bandline/.style={draw=structural,line width=.3pt},
    blockbox/.style={line width=.4pt,densely dashed},
    frame/.style={draw=black,line width=.35pt},
    note/.style={font=\scriptsize,align=left,text=black},
    callout/.style={draw=black,line width=.3pt,->,>=stealth}]
  \definecolor{boundconvsecond}{RGB}{170,55,130}
  \definecolor{boundfinal}{RGB}{97,52,166}
  \def\cell{.24}
  \node[font=\small] at (2.4,.69)
    {Output-layer tensors ($3\times20$)};
  \foreach \mid/\name in {3.5/x,9.5/h,14.5/r,18.5/y}
    \node[font=\scriptsize,anchor=south] at (\mid*\cell,.08) {$\name$};

  \newcommand{\fullboundgrid}[2]{%
    \foreach \row in {0,...,2}
      \foreach \col in {0,...,19}
        \draw[gridline] (\col*\cell,{#1-\row*\cell}) rectangle ++(\cell,-\cell);
    \draw[frame] (0,#1) rectangle ++({20*\cell},{-3*\cell});
    \foreach \v in {7,12,17}
      \draw[bandline] (\v*\cell,#1) -- ++(0,{-3*\cell});
    \node[note,anchor=east,align=right] at (-.12,{#1-1.5*\cell})
      {bounds\\over $#2$};}

  \fullboundgrid{0}{r}
  \foreach \row in {0,...,2}
    \foreach \tap/\shade in {0/25,1/55,2/90}{
      \fill[boundconvsecond!\shade]
        ({(12+\row+\tap)*\cell},{-\row*\cell}) rectangle ++(\cell,-\cell);
      \draw[gridline]
        ({(12+\row+\tap)*\cell},{-\row*\cell}) rectangle ++(\cell,-\cell);}
  \draw[blockbox,draw=boundconvsecond] ({12*\cell},0)
    rectangle ++({5*\cell},{-3*\cell});
  \node[note,anchor=west] (overr) at (5.03,-.36) {block $L_4^3$};
  \draw[callout] (overr.west)--({17*\cell+.025},-.36);

  \draw[->,>=stealth,structural,line width=.4pt] (2.4,-.88)--(2.4,-1.38);
  \node[note,anchor=east] at (2.18,-1.13) {match $r$};
  \node[note,anchor=west] at (2.62,-1.13) {multiply by $L$};

  \fullboundgrid{-1.55}{h}
  \foreach \row in {0,...,2}
    \foreach \tap in {0,...,2}{
      \pgfmathtruncatemacro{\shade}{20+9*mod(3*\tap+\row,9)}
      \fill[teal!\shade]
        ({(7+\row+\tap)*\cell},{-1.55-\row*\cell}) rectangle ++(\cell,-\cell);
      \draw[gridline]
        ({(7+\row+\tap)*\cell},{-1.55-\row*\cell}) rectangle ++(\cell,-\cell);}
  \draw[blockbox,draw=teal] ({7*\cell},-1.55)
    rectangle ++({5*\cell},{-3*\cell});
  \node[note,anchor=west] (overh) at (5.03,-1.91) {block $L_4^2$};
  \draw[callout] (overh.west)--({12*\cell+.025},-1.91);

  \draw[->,>=stealth,structural,line width=.4pt] (2.4,-2.43)--(2.4,-2.93);
  \node[note,anchor=east] at (2.18,-2.68) {match $h$};
  \node[note,anchor=west] at (2.62,-2.68) {multiply by $L$};

  \fullboundgrid{-3.10}{x}
  \foreach \row in {0,...,2}
    \foreach \tap in {0,...,4}{
      \pgfmathtruncatemacro{\shade}{15+6*mod(9*\tap+4*\row,15)}
      \fill[boundfinal!\shade]
        ({(\row+\tap)*\cell},{-3.10-\row*\cell}) rectangle ++(\cell,-\cell);
      \draw[gridline]
        ({(\row+\tap)*\cell},{-3.10-\row*\cell}) rectangle ++(\cell,-\cell);}
  \draw[blockbox,draw=boundfinal] (0,-3.10)
    rectangle ++({7*\cell},{-3*\cell});
  \node[note,anchor=west] (overx) at (5.03,-3.46) {block $L_4^1$};
  \draw[callout] (overx.west)--({7*\cell+.025},-3.46);

\end{tikzpicture}}
        \caption{Output-layer tensor $L_4$ at successive stages.}
        \Description{Three full three-by-twenty output-layer tensors. Their
            highlighted blocks express bounds over r, then h, then x.
            The block labels identify the successive reference layers.}
        \label{fig:gb-bound-progression}
    \end{subfigure}
    \caption[Full tensors and successive output bounds]{The initial tensor
        $L$ and successive full output-layer tensors. In (b), $L_4^3$, $L_4^2$, and $L_4^1$ label the highlighted
        blocks within the full tensors. The two convolution blocks in (a)
        use different hues, with repeated shades indicating shared
        weights within each block. Distinct colors in the result blocks show different values. }
    \label{fig:gb-running}
\end{figure}
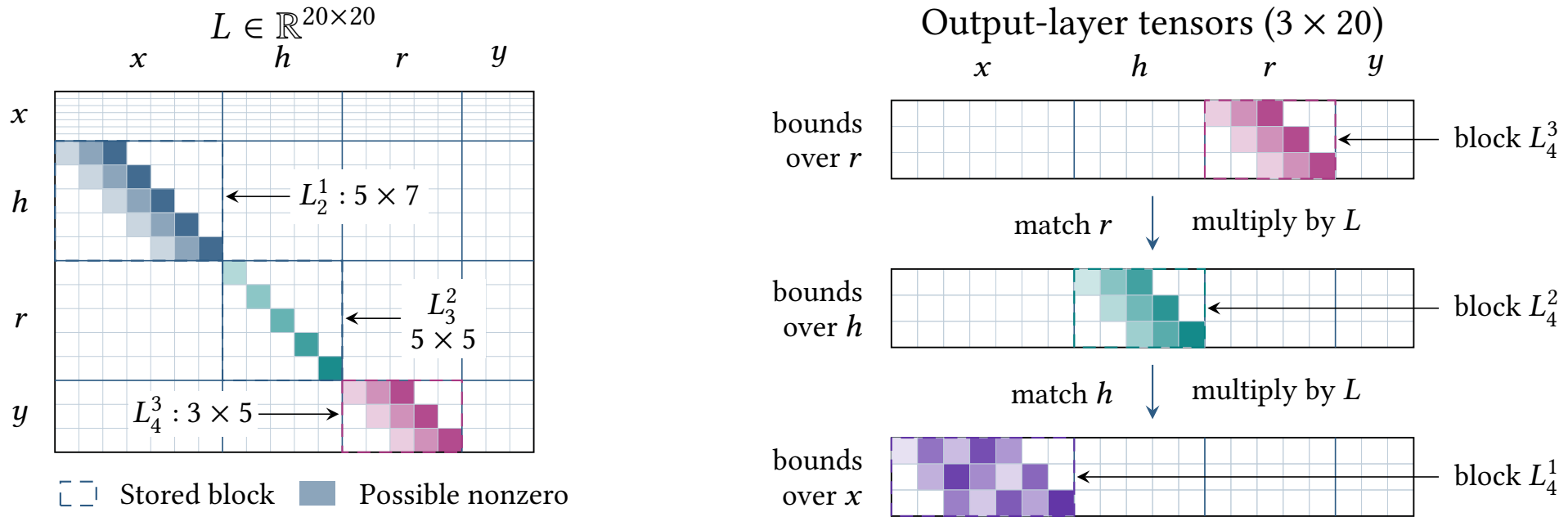
Fig.~\ref{fig:gb-bound-tensor} shows $L$ and
Fig.~\ref{fig:gb-bound-progression} shows the full output-layer tensor
at three stages, containing $L_4^3$, $L_4^2$, and $L_4^1$, respectively. Each surrounding grid has columns
for all $20$ network neurons.
Most entries are zero. Within each convolution block, a row contains three
potentially nonzero coefficients, and the same kernel weights are shared
across rows. The ReLU block is diagonal. The \sparse representation described in \neuton~\cite{neuton} exploits the sparsity and repeated values in the $L$ tensors to avoid materializing the dense coefficient tensors.
It records
the sparsity patterns: $L_2^1$ and $L_4^3$ use the \stblock type, while $L_3^2$
uses the \diagblock type.
The result block $L_4^2$ still has windows of width $3$, but its entries
$w_t\lambda_{a+t}$ need not be shared across rows. It therefore uses a
\wblock. The final tensor contains the block $L_4^1$, a
\wblock of width $5$ in the columns of $x$.

While the \sparse representation reduces the storage and computation in \S\ref{sec:overview-computation}, the calls $L_4\tmul L$ must determine which blocks contribute and how to represent each
result. The first call matches the columns for $r$ in $L_4$ with the rows
for $r$ in $L$, selects $L_4^3$ and $L_3^2$, and determines that their
product has windows of width three in the columns for $h$. The next call
receives a different intermediate representation. It matches the columns
for $h$, selects $L_4^2$ and $L_2^1$, and determines the width-five windows
of $L_4^1$. These decisions are made during execution, alongside computing
the coefficient values.

Making these decisions during execution introduces dependencies between
the CPU and GPU. The CPU matches blocks and selects the numerical
operations that the GPU will perform. When a decision also reads
computed coefficients, the CPU must wait for the corresponding GPU
result before selecting and issuing the next operation. The GPU must
then wait for the CPU to complete this work. These dependencies can
leave the GPU idle even when transferring the result takes negligible
time. They also restrict the sequence of tensor operations available to
a compiler, limiting its ability to combine operations across such
decisions.

A further inefficiency is that these decisions are repeated across
certification runs. The same abstract interpreter often analyzes one
DNN for different properties or for the subregions produced by input
splitting. New input regions can change the ReLU slopes and coefficient
values. Nevertheless, whenever an operation executes, the program again
matches the same layer ranges and determines the same convolution
windows and admissible result layouts. Thus, compact coefficient storage
still requires repeated computation of information that remains valid
across runs.
We now show how \sys specializes the interpreter to a fixed DNN
architecture to separate these representation decisions from coefficient
computation. Their results can then be computed once and reused across
certification runs.




\subsection{Specializing to DNN Architecture}\label{sec:overview-solution}

\begin{figure}[t]
    \centering
    \fitwidth{
\begingroup
\definecolor{opteal}{RGB}{0,111,118}
\definecolor{opbluewash}{RGB}{239,247,255}
\definecolor{optealwash}{RGB}{232,249,247}
\begin{tikzpicture}[x=.95cm,y=.88cm,>=Latex,font=\footnotesize,text=black,
    steptitle/.style={font=\footnotesize,align=center},
    detail/.style={font=\scriptsize,align=center},
    sarr/.style={->,draw=structural,line width=.4pt},
    narr/.style={->,draw=opteal,line width=.4pt}]
  \draw[structural,fill=opbluewash,dashed,line width=.3pt]
    (0,1.95) rectangle (12.0,3.60);
  \draw[opteal,fill=optealwash,dashed,line width=.3pt]
    (0,-.46) rectangle (12.0,1.45);
  \node[anchor=west,font=\footnotesize] at (.18,3.32)
    {Structural computation};
  \node[steptitle] at (1.85,2.90) {1. Match blocks};
  \node[detail] at (1.85,2.39)
    {$L_4^3$ with $L_3^2$\\Columns $r$ match rows $r$};
  \node[steptitle] at (5.95,2.90) {2. Determine result layout};
  \node[detail] at (5.95,2.39)
    {Columns $h$; windows of width $3$\\\wblock};
  \node[steptitle] at (10.15,2.90) {3. Select a routine};
  \node[detail] at (10.15,2.39)
    {Scale columns by\\the diagonal entries of $L_3^2$};
  \draw[sarr] (3.40,2.60)--(4.03,2.60);
  \draw[sarr] (7.95,2.60)--(8.62,2.60);
  \node[anchor=west,font=\footnotesize] at (.18,1.16)
    {Numerical computation};
  \node[detail] at (1.85,.43)
    {Weights $w_0,w_1,w_2$\\[2pt]
      Slopes $\lambda_0,\ldots,\lambda_4$};
  \node[draw=opteal,fill=white,line width=.3pt,
      minimum width=3.15cm,minimum height=.66cm,inner sep=3pt,
      align=center] (scale) at (6.55,.43)
    {Scale columns\\[2pt]
      $L_4^2[a,a+t]=w_t\lambda_{a+t}$};
  \draw[narr] (3.35,.43)--(scale.west);
  \draw[sarr] (10.15,2.07)--(10.15,1.70)--(6.55,1.70)--(scale.north);
  \node[detail,fill=white,inner sep=2pt] at (8.43,1.70)
    {Routine and layout};
  \begin{scope}[shift={(9.72,.76)},x=.10cm,y=-.10cm]
    \foreach \row in {0,1,2}
      \foreach \tap in {0,1,2}{
        \pgfmathtruncatemacro{\shade}{30+7*(3*\row+\tap)}
        \fill[opteal!\shade] ({7+\row+\tap},\row) rectangle ++(1,1);}
    \foreach \i in {0,...,20}
      \draw[structural!35,line width=.15pt] (\i,0)--(\i,3);
    \foreach \i in {0,...,3}
      \draw[structural!35,line width=.15pt] (0,\i)--(20,\i);
    \draw[black,line width=.3pt] (0,0) rectangle (20,3);
    \foreach \i in {7,12,17}
      \draw[black,line width=.25pt] (\i,0)--(\i,3);
    \draw[opteal,densely dashed,line width=.35pt] (7,0) rectangle (12,3);
    \foreach \mid/\name in {3.5/x,9.5/h,14.5/r,18.5/y}
      \node[font=\scriptsize,anchor=south] at (\mid,-.10) {$\name$};
  \end{scope}
  \node[detail,anchor=north] at (10.72,.35)
    {Result $L_4$\\Stored block $L_4^2$};
  \draw[narr] (scale.east)--(9.57,.43);
\end{tikzpicture}
\endgroup}
    \caption[Structural and numerical computation in a tensor operation]{Computing
        $L_4^2$ with the whole-tensor call $L_4\tmul L$. }
    \label{fig:gb-matmul}
\end{figure}
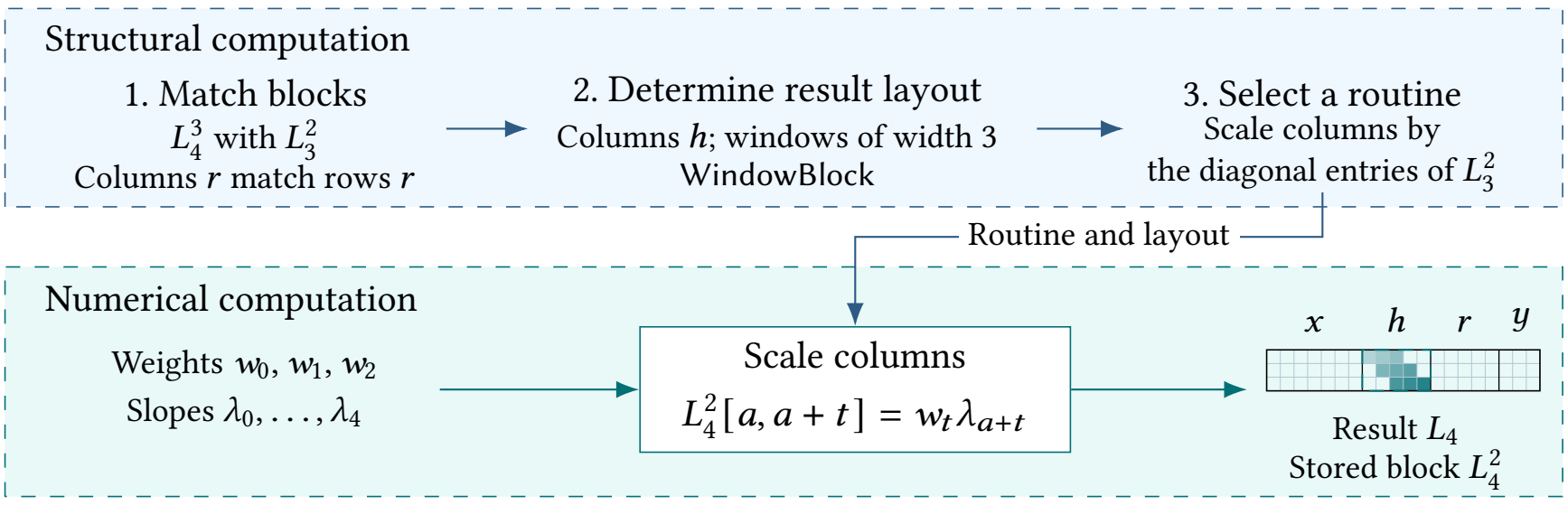

We distinguish two kinds of computation in the running example.
First, \textit{structural computation} involves matching the operand
blocks, determining the result's positions and internal representation,
and selecting the arithmetic to perform. Second, \textit{numerical
computation} evaluates the arithmetic on weights, bounds, and relaxation
parameters. For the first multiplication, determining that $L_4^2$ uses
width-three windows is structural computation, while evaluating its
entries $w_t\lambda_{a+t}$ is numerical computation.
Fig.~\ref{fig:gb-matmul} illustrates this distinction.
In the general GPU implementation, structural computation executes on
the CPU, while numerical computation operates on coefficient arrays
using GPU kernels.

Our solution is based on two observations. First, for a fixed abstract
interpreter, the DNN architecture determines
the structural patterns used by these tensor operations. In the example,
the layer dimensions determine the block positions, while the convolution
and ReLU operations determine the value-sharing patterns.
Changing the weights or input region changes the coefficients, but the
dimensions of $L_4^2$ and $L_4^1$ remain constant. 
%
Second, the structure of an intermediate result can determine the
structure needed by the next operation without evaluating the
intermediate coefficients. Knowing the dimensions of
$L_4^2$ and the convolution kernel size of $L_2^1$ is sufficient to
determine the width-five representation of $L_4^1$. We can therefore
delay the numerical computation of $L_4^2$ and first perform the
structural computation for both multiplications. This structural computation can be reused across runs with different numerical inputs.

\begin{figure}[t]
    \centering
    \begin{subfigure}{\linewidth}
        \centering
        \fitwidth{
\begin{tikzpicture}[x=.86cm,y=.82cm,
    >=Latex,font=\footnotesize,
    cpu/.style={draw=structural,fill=tlBlueWash,line width=.35pt},
    gpu/.style={draw=tlTeal,fill=tlTealWash,line width=.4pt},
    dependency/.style={->,draw=structural,line width=.4pt},
    lane/.style={anchor=east,font=\footnotesize,text=black},
    note/.style={font=\scriptsize,text=black,align=center}]
  \definecolor{tlTeal}{RGB}{0,105,112}
  \definecolor{tlBlueWash}{RGB}{223,238,255}
  \definecolor{tlTealWash}{RGB}{218,245,245}
  \path[use as bounding box] (-.05,-1.56) rectangle (13.65,.98);
  \def\xzero{1.15}
  \def\sw{2.10}
  \def\kw{1.70}
  \def\lh{.44}
  \node[lane] at (.95,.22) {CPU};
  \node[lane] at (.95,-.83) {GPU};
  \draw[->,black,line width=.3pt] (\xzero,-1.40)--(13.02,-1.40)
    node[anchor=west,font=\scriptsize,text=black] {Time};

  \node[note,anchor=south] at (6.85,.70) {Repeated for each certification run};
  \foreach \i in {0,1,2}{%
    \pgfmathsetmacro{\sx}{\xzero+\i*(\sw+\kw)}
    \pgfmathsetmacro{\gx}{\sx+\sw}
    \pgfmathtruncatemacro{\k}{\i+1}
    \draw[cpu] (\sx,0) rectangle ++(\sw,\lh);
    \node[note] at ({\sx+\sw/2},.22) {Structure $\k$};
    \draw[gpu] (\gx,-1.05) rectangle ++(\kw,\lh);
    \node[text=tlTeal] at ({\gx+\kw/2},-.83) {$\kappa_{\k}$};
    \draw[structural,densely dashed,line width=.3pt]
      (\sx,-.83)--(\gx,-.83)
      node[midway,above=2pt,font=\scriptsize,text=black] {Waiting};
    \draw[dependency] (\gx,0)--(\gx,-.61);
    \ifnum\i<2
      \draw[dependency,densely dashed] ({\gx+\kw},-.61)--({\gx+\kw},0);
    \fi
  }
\end{tikzpicture}}
        \caption{General execution interleaves structural and numerical computation.}
        \Description{Three structural computations on the CPU alternate with
            three numerical GPU operations. Arrows show the dependencies
            between them.}
        \label{fig:gb-timeline-general}
    \end{subfigure}
    \par\medskip
    \begin{subfigure}{\linewidth}
        \centering
        \fitwidth{
\begin{tikzpicture}[x=.86cm,y=.82cm,
    >=Latex,font=\footnotesize,
    cpu/.style={draw=structural,fill=tlBlueWash,line width=.35pt},
    gpu/.style={draw=tlTeal,fill=tlTealWash,line width=.4pt},
    dependency/.style={->,draw=structural,line width=.4pt},
    lane/.style={anchor=east,font=\footnotesize,text=black},
    note/.style={font=\scriptsize,text=black,align=center}]
  \definecolor{tlTeal}{RGB}{0,105,112}
  \definecolor{tlBlueWash}{RGB}{223,238,255}
  \definecolor{tlTealWash}{RGB}{218,245,245}
  \path[use as bounding box] (-.05,-1.56) rectangle (13.65,.98);
  \def\xzero{1.15}
  \def\sw{2.10}
  \def\kw{1.70}
  \def\lh{.44}
  \node[lane] at (.95,.22) {CPU};
  \node[lane] at (.95,-.83) {GPU};
  \draw[->,black,line width=.3pt] (\xzero,-1.40)--(13.02,-1.40)
    node[anchor=west,font=\scriptsize,text=black] {Time};

  \pgfmathsetmacro{\boundary}{\xzero+3*\sw}
  \pgfmathsetmacro{\finish}{\boundary+3*\kw}
  \node[note,anchor=south,font=\scriptsize,text=structural]
    at ({\xzero+1.5*\sw},.70) {Once per architecture};
  \node[note,anchor=south,font=\scriptsize,text=tlTeal]
    at ({\boundary+1.5*\kw},.70) {Each certification run};
  \draw[black,densely dashed,line width=.3pt]
    (\boundary,.63)--(\boundary,-1.24);
  \foreach \i in {0,1,2}{%
    \pgfmathsetmacro{\sx}{\xzero+\i*\sw}
    \pgfmathsetmacro{\gx}{\boundary+\i*\kw}
    \pgfmathtruncatemacro{\k}{\i+1}
    \draw[cpu] (\sx,0) rectangle ++(\sw,\lh);
    \node[note] at ({\sx+\sw/2},.22) {Structure $\k$};
    \draw[gpu] (\gx,-1.05) rectangle ++(\kw,\lh);
    \node[text=tlTeal] at ({\gx+\kw/2},-.83) {$\kappa_{\k}$};
  }
  \draw[dependency] (\boundary,0)--(\boundary,-.61);
  \node[note] at ({\xzero+1.5*\sw},-.83) {Numerical work deferred};
  \node[note] at ({\boundary+1.5*\kw},.22) {No repeated structural work};
\end{tikzpicture}}
        \caption{All structural computation precedes the numerical phase.}
        \Description{The same three structural computations are contiguous
            on the left of the CPU timeline. The three numerical operations
            begin to their right, after the structural phase ends.}
        \label{fig:gb-timeline-specialized}
    \end{subfigure}
    \caption[General and specialized execution]{Execution before and after
        specialization for a sequence of numerical calls
        $\kappa_i$. Subfigure (a) includes CPU decisions that consume GPU
        results. In (b), all structural work precedes numerical execution.
        }
    \Description{Two separately captioned CPU and GPU timelines show
        interleaved execution and execution with a complete structural
        phase before the numerical phase.}
    \label{fig:gb-timeline}
\end{figure}
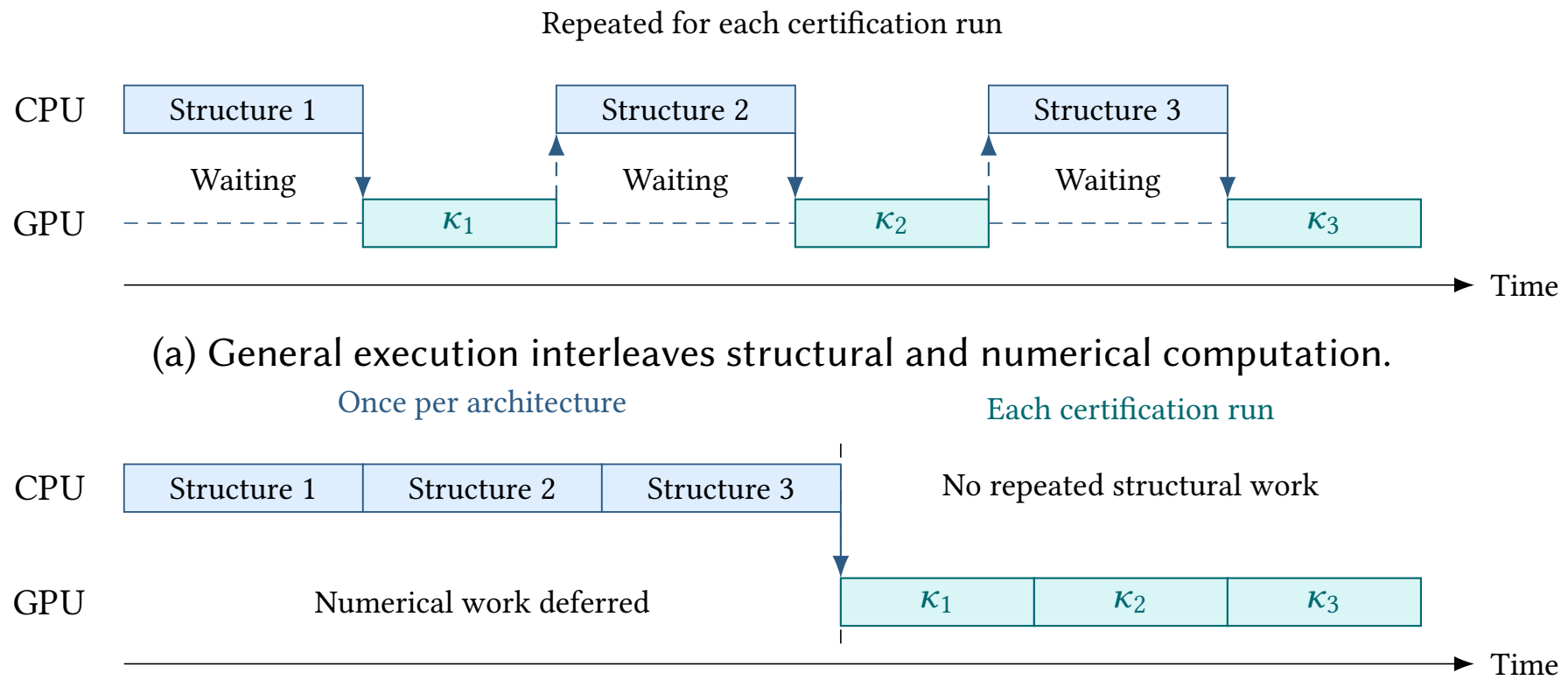

\sys formalizes this separation using novel structural semantics for DNN abstract interpreters. Each tensor
has a descriptor recording its structure and a payload containing its
numerical values. A structural rule computes the result descriptor from
the operand descriptors and records the numerical code to execute later.
Applying these rules successively determines intermediate descriptors
without evaluating their payloads. The recorded code preserves
dependencies between numerical operations.
Structural execution applies these rules to the interpreter and assembles
the recorded numerical code into an executable for the fixed
architecture. Fig.~\ref{fig:gb-timeline} illustrates this separation.
The resulting numerical code consists of tensor operations whose
dependencies are available before certification begins. A tensor
compiler can therefore apply kernel fusion and memory planning across
these operations. The same information also allows \sys to process
independent neuron computations in groups, retaining their final bounds
while releasing large intermediate tensors between groups. Structural
computation is performed once, and the executable can be reused with
different weights and input regions, including the subregions produced
by input splitting.

\begin{samepage}
\S\ref{sec:gb-format} defines structural semantics for tensor operations
and proves its correctness. 
\S\ref{sec:spec} uses these structural semantics to specialize DNN abstract interpreters and proves preservation of the computed bounds.
\S\ref{sec:spec-eval} evaluates \sys, and \S\ref{sec:spec-related}
discusses related work.
\par
\end{samepage}

\section{Structural Semantics of g-BCSR}\label{sec:gb-format}
We define structural semantics that determine the result descriptor
and numerical calls of a tensor operation from its input descriptors. We develop it for the \sparse
representation introduced by \neuton~\cite{neuton}. We first give
meaning to its descriptors, then define rules for tensor
operations and prove their correspondence with numerical execution.
The resulting semantics can be applied successively to a fixed sequence
of operations. \S\ref{sec:spec} extends it to the control of an abstract
interpreter.

\subsection{Descriptor and Payload of a g-BCSR Tensor}\label{sec:gb-defs}
A \sparse tensor $\mathcal T=(d,S,B,Z)$ stores background $d$, shape
$Z$, and blocks $B=\{B_k\}_{k=1}^m$ at positions $S=\{s_k\}_{k=1}^m$
(\S\ref{sec:background-gbcsr}). Each block has a structural support
$\mathcal S$ of positions that may be nonzero and an equivalence relation
$\approx$ specifying which positions share a stored value. A structural
type describes all arrays satisfying these constraints.

\begin{definition}[Structural model of a g-BCSR block]\label{def:gb-block}
Let $\EReal=\R\cup\{-\infty,+\infty\}$ be the extended reals.
A block is a tuple $B=(\Omega,\mathcal S,\approx,b)$, where $\Omega$ is
a finite rectangular index domain, $\mathcal S\subseteq\Omega$ is its
\emph{structural support}, $\approx$ is an equivalence relation on
$\mathcal S$, and the \emph{core} $b:\mathcal S/{\approx}\to\EReal$
assigns a value to each equivalence class. Write $[\iota]_\approx$ for
the class containing $\iota$. For $\iota\in\Omega$, the block denotes
$\sem B[\iota]=b[[\iota]_\approx]$ if $\iota\in\mathcal S$, and zero
otherwise. Its \emph{structural type} is the projection
$\tau=(\Omega,\mathcal S,\approx)$. The denotation $\sem\tau$ consists
of arrays $A:\Omega\to\EReal$ satisfying $A[\iota]=0$ for
$\iota\in\Omega\setminus\mathcal S$, and $A[\iota]=A[\iota']$ whenever
$\iota\approx\iota'$.
\end{definition}

\begin{definition}[Descriptor and payload of a g-BCSR tensor]\label{def:gb-descriptor}\label{sec:gb-split}
For a tensor $\mathcal T=(d,S,B,Z)$ with blocks
$B_k=(\Omega_k,\mathcal S_k,\approx_k,b_k)$ and types
$\tau_k=(\Omega_k,\mathcal S_k,\approx_k)$, its \emph{descriptor} and
\emph{payload} are, respectively,
\begin{equation}\label{eq:gb-descr}
  \descr(\mathcal T)=\bigl(d,Z,\{(s_k,\tau_k)\}_{k=1}^m\bigr),
  \qquad
  \payload(\mathcal T)=\{b_k\}_{k=1}^m.
\end{equation}
The descriptor records the background, tensor shape, block positions,
and block types, including their index domains. The payload contains
the numerical cores. The background is a fixed representation parameter.
Values that vary across certification runs belong to the payload.
Together the two projections determine $\mathcal T$ and extend
componentwise to tuples and environments of tensors.
\end{definition}

\subsection{Structural Rules for Tensor Operations}\label{sec:gb-algorithms}\label{sec:gb-separation}
We represent a block during structural evaluation by $\beta=(\tau,u)$.
Here $\tau$ is its structural type, and $u$ is a variable that will hold
its numerical core when the generated code runs. Structural evaluation
uses this variable without reading the core. 
An arrow above a symbol denotes a tuple. We first define operations
on two blocks over a common rectangular region $r$. The tensor operation below constructs these regions when
operand blocks occupy different positions.

\pheader{Operations on blocks}
An operand over $r$ is either a symbolic block $(\tau_\ell,u_\ell)$ or
$\underline{d_\ell}$, meaning the operand has its background
value throughout $r$. With operator $\odot$ and output background $d'$
fixed, the block judgment is
$\vec\beta\xRightarrow[r]{\odot}(\beta,\kappa)$, where
  $\vec\beta=(\beta_1,\beta_2).$
It produces a symbolic result block
$\beta$ and recorded numerical calls $\kappa$. The result $\beta=\bot$
means that no output block is needed because the region equals $d'$.
We write $\epsilon$ for an empty call sequence. Fig.~\ref{fig:gb-elementwise-rules}
defines the three cases.

\begin{figure}[t]
    \centering
    \begingroup\small
    \input{figures/gb-elementwise-rules.tex}
    \endgroup
    \caption[Structural rules for block operations]{Structural
        rules for two operands over a common region $r$. $\{i,j\}=\{1,2\}$. 
        }
    \Description{Three inference rules arranged horizontally reuse a
        block under an identity, omit an annihilated result, or record
        the numerical computation of a new block.}
    \label{fig:gb-elementwise-rules}
\end{figure}

\textsc{B-Identity} applies when one operand is a block and the other
supplies its background value $d_j$. If combining with this background leaves the
block unchanged, as tested by $\mathsf{id}_j(\odot,d_j)$ on operand side $j$,
the rule reuses the block's type and core and emits $\epsilon$.
\textsc{B-Annihilator} applies when the operation produces
$d_j$ at every position regardless of the block's entries, expressed by
$\mathsf{ann}(\odot,d_j)$. If $d_j=d'$, the output background already
represents this result, so the rule returns $\bot$ and emits $\epsilon$.

\textsc{B-Compute} handles the remaining cases. Its helper
$\operatorname{Lift}_r$ retains block
operands and replaces a background operand with a constant block,
recording its core construction in $\lambda$. The rule then selects
an admissible result type $\tau=\tau_1\mathbin{\odot^\tau}\tau_2$
and records the numerical conversions and arithmetic in the kernel
call $K_{\odot,\vec\tau,\tau}$. The new block is $(\tau,u)$, where
$u$ will hold that call's result. Fig.~\ref{fig:gb-result-types} gives the constraints
used to select the result type for addition and multiplication. It uses the union of supports for addition
and their intersection for multiplication. Sharing is retained where both
operands guarantee equality. The extension $\widehat{\approx}_\ell$
also relates positions outside $\mathcal S_\ell$, which are known zeros.
Restriction to $\mathcal S_\odot$ keeps only pairs in the result support.
Appendix~\ref{app:gb-block-rules} explains all three rules in detail,
and Appendix~\ref{sec:gb-result-types} gives the result-type selection
using the block-type order and tables.

\pheader{Operations on g-BCSR tensors}
We now define an operation $f$ on
complete tensors $\vec{\mathcal T}=(\mathcal T_1,\ldots,\mathcal T_q)$.
Given their descriptors
$\vec\pi=\descr(\vec{\mathcal T})$,
structural evaluation produces a result descriptor $\pi'$ and code
$\kappa$, written $\vec\pi\xRightarrow{f}(\pi',\kappa)$.
For an element-wise operation $f=\odot$, the two operands have common
shape $Z$ and backgrounds $d_1,d_2$. \textsc{T-Elementwise} in Fig.~\ref{fig:gb-tensor-rules}
first uses $\operatorname{Refine}$ to choose ordered, disjoint regions
covering their blocks. For each region $r$, $\operatorname{Align}$
describes both operands in $\vec\beta_r$ using the background value when no block is defined, and records any restriction or
materialization calls in $\alpha_r$. The block rules then produce $\beta_r$ and
$\kappa_r$ for each region. For each $r\in\mathcal G$ with $\beta_r=(\tau_r,u_r)$, the descriptor
retains $(s_r,\tau_r)$ and the return retains $u_r$. The output background
is $d'=d_1\odot d_2$ when defined, and zero otherwise. The recorded
program concatenates each region's alignment calls $\alpha_r$ and block
calls $\kappa_r$ in order. Alignment calls remain even when a block rule
emits $\epsilon$. These choices depend only on descriptors. The recorded code computes
the numerical values when executed on the input payloads. Appendix~\ref{app:gb-helpers}
defines the auxiliary functions and gives an example of the complete
operation.
\begin{figure}[t]
    \centering
    \begin{subfigure}[b]{.44\linewidth}
        \centering
        \vspace{0pt}
        \begingroup\small
        \input{figures/gb-result-type-rules.tex}
        \endgroup
        \caption{Block-type transfer.}
        \label{fig:gb-result-types}
    \end{subfigure}\hfill
    \begin{subfigure}[b]{.54\linewidth}
        \centering
        \vspace{0pt}
        \begingroup\small
        \input{figures/gb-tensor-rule.tex}
        \endgroup
        \caption{Operation on complete tensors.}
        \label{fig:gb-tensor-operation}
    \end{subfigure}
    \caption[(a) shows block types and tensor operations]{The type
        calculation used by \textsc{B-Compute} and (b) shows the tensor rule
        that applies the block rules across regions. 
        }
    \Description{Two panels show support and sharing calculations for
        block types and an inference rule combining regional results
        into a complete tensor descriptor and numerical program.}
    \label{fig:gb-tensor-rules}
\end{figure}

\pheader{Tensor multiplication}
For zero backgrounds, block pairs contribute when the left block's
column range overlaps the right block's row range. We partition their
output boxes into disjoint regions, determine
each product's support from its operand types, and accumulate
contributions using the addition rule. Nonzero backgrounds instead require
a dense fallback. Appendix~\ref{sec:gb-matmul} gives the full procedure.
Both constructions use fixed region, type-selection, and accumulation
policies. 

\subsection{Correctness and Composition}\label{sec:gb-staging}
The structural rules separate a tensor operation into descriptor
computation and numerical calls. Correctness requires that executing
these calls and combining their result with the computed descriptor
gives the same tensor values as direct dense evaluation and the existing
\sparse operation. This must hold for every numerical payload compatible
with the input descriptors. We write $\operatorname{Pack}(\pi,\rho)$
for the tensor reconstructed from descriptor $\pi$ and compatible payload
$\rho$. Thus
$\operatorname{Pack}(\descr(\mathcal T),\payload(\mathcal T))=\mathcal T$.
Both reconstruction and denotation extend componentwise to tuples and
environments of tensors.

We use $f$ for the operation defined by the structural rules and its dense
counterpart, distinguished by their operands, and $f_{\mathrm{gb}}$ for
the existing \sparse operation. 
We fix compatible shapes, operation
parameters, and structural policies, and use exact arithmetic. We define contracts for the semantics of procedures such as $\operatorname{Align}, \operatorname{Refine},$ and $\operatorname{Lift}_r$. Numerical kernels also must terminate and compute their specified operations correctly
whenever the corresponding dense computation is defined.
Appendix~\ref{app:gb-sep} states these contracts and proves the following theorem.

\begin{figure}[t]
    \centering
    \begin{subfigure}[t]{.33\linewidth}
        \centering
        \vspace{0pt}
        \fitwidth{
\begingroup
\definecolor{sepBlueWash}{RGB}{239,247,255}
\begin{tikzpicture}[x=1cm,y=1cm,>=Latex,font=\footnotesize,text=black,
    value/.style={minimum height=.34cm,inner sep=2pt},
    title/.style={font=\footnotesize,anchor=west},
    sarr/.style={->,draw=structural,line width=.4pt},
    rulearr/.style={draw=structural,double,double distance=.5pt,
      -{Implies},line width=.25pt}]
  \path[use as bounding box] (0,0) rectangle (3.8,3.10);
  \draw[structural,fill=sepBlueWash,dashed,line width=.3pt]
      (0,0) rectangle (3.8,1.00);
  \node[title] at (.12,2.86) {Tensor operation};
  \node[title,anchor=center] at (1.90,1.24) {Structural rules};
  \node[value] (inputs) at (.66,2.15) {$\vec{\mathcal T}$};
  \node[value] (result) at (3.12,2.15) {$\mathcal T'$};
  \node[value] (din) at (.66,.34) {$\descr(\vec{\mathcal T})$};
  \node[value] (dout) at (3.12,.34) {$\descr(\mathcal T')$};
  \draw[->,black,line width=.4pt] (inputs.east)--(result.west)
      node[midway,above=2pt] {$f$};
  \draw[sarr] (inputs.south)--(din.north)
      node[pos=.35,left=2pt] {$\descr$};
  \draw[sarr] (result.south)--(dout.north)
      node[pos=.35,right=2pt] {$\descr$};
  \draw[rulearr] (din.east)--(dout.west)
      node[midway,above=1pt] {$f$};
\end{tikzpicture}
\endgroup}
        \caption{One operation.}
        \label{fig:gb-square}
    \end{subfigure}\hfill
    \begin{subfigure}[t]{.65\linewidth}
        \centering
        \vspace{0pt}
        \fitwidth{
\begingroup
\definecolor{stageTeal}{RGB}{0,111,118}
\definecolor{stageBlueWash}{RGB}{239,247,255}
\definecolor{stageTealWash}{RGB}{232,249,247}
\begin{tikzpicture}[x=1cm,y=1cm,>=Latex,font=\footnotesize,text=black,
    value/.style={minimum height=.32cm,inner sep=2pt},
    title/.style={font=\footnotesize,anchor=west},
    note/.style={font=\scriptsize},
    sarr/.style={->,draw=structural,line width=.4pt},
    rulearr/.style={draw=structural,double,double distance=.5pt,
      -{Implies},line width=.25pt},
    narr/.style={->,draw=stageTeal,line width=.4pt}]
  \path[use as bounding box] (0,0) rectangle (7.8,3.10);
  \draw[structural,fill=stageBlueWash,dashed,line width=.3pt]
      (0,1.65) rectangle (7.8,3.10);
  \draw[stageTeal,fill=stageTealWash,dashed,line width=.3pt]
      (0,0) rectangle (7.8,1.00);
  \node[title] at (.12,2.86) {Compile time};
  \node[title] at (.12,.79) {Run time};
  \node[value] (d0) at (.58,2.15) {$\descr(E_0)$};
  \node[value] (d1) at (2.28,2.15) {$\descr(E_1)$};
  \node (ddots) at (3.90,2.15) {$\cdots$};
  \node[value] (dm) at (5.46,2.15) {$\descr(E_{N-1})$};
  \node[value] (dn) at (7.22,2.15) {$\descr(E_N)$};
  \draw[rulearr] (d0.east)--(d1.west)
      node[midway,above=2pt] {$f_1$};
  \draw[rulearr] (d1.east)--(ddots.west);
  \draw[rulearr] (ddots.east)--(dm.west);
  \draw[rulearr] (dm.east)--(dn.west)
      node[midway,above=2pt] {$f_N$};
  \node[value] (p0) at (.58,.34) {$\payload(E_0)$};
  \node[value] (k1) at (1.43,.34) {$\kappa_1$};
  \node (kdots) at (3.90,.34) {$\cdots$};
  \node[value] (kn) at (6.34,.34) {$\kappa_N$};
  \node[value] (pn) at (7.22,.34) {$\payload(E_N)$};
  \draw[narr] (p0.east)--(k1.west);
  \draw[narr] (k1.east)--(kdots.west);
  \draw[narr] (kdots.east)--(kn.west);
  \draw[narr] (kn.east)--(pn.west);
  \draw[structural,line width=.3pt]
      (d0.south)--(.58,1.80)--(2.28,1.80)--(d1.south);
  \draw[structural,line width=.3pt]
      (dm.south)--(5.46,1.80)--(7.22,1.80)--(dn.south);
  \draw[sarr] (1.43,1.80)--(k1.north);
  \draw[sarr] (6.34,1.80)--(kn.north);
  \node[note,anchor=west] at (1.55,1.32) {$\vec\pi_1,\pi_1$};
  \node[note] at (3.90,1.32) {Fixed descriptors};
  \node[note,anchor=west] at (6.46,1.32) {$\vec\pi_N,\pi_N$};
\end{tikzpicture}
\endgroup}
        \caption{A sequence of operations.}
        \label{fig:gb-stage}\label{fig:gb-stage-specialized}
    \end{subfigure}
    \caption[Structural evaluation of operations and sequences]{Structural
        evaluation of one operation (a) and a fixed sequence (b). (a) shows the descriptor component of
        $\vec\pi\xRightarrow{f}(\pi',\kappa)$. 
        (b) shows the intermediate descriptors
        and calls $\kappa_j$ generated for later numerical execution.}
    \Description{A commuting square shows agreement of result descriptors.
        A second panel separates the structural evaluation of a fixed
        operation sequence from execution of its recorded numerical calls.}
    \label{fig:gb-separation}
\end{figure}
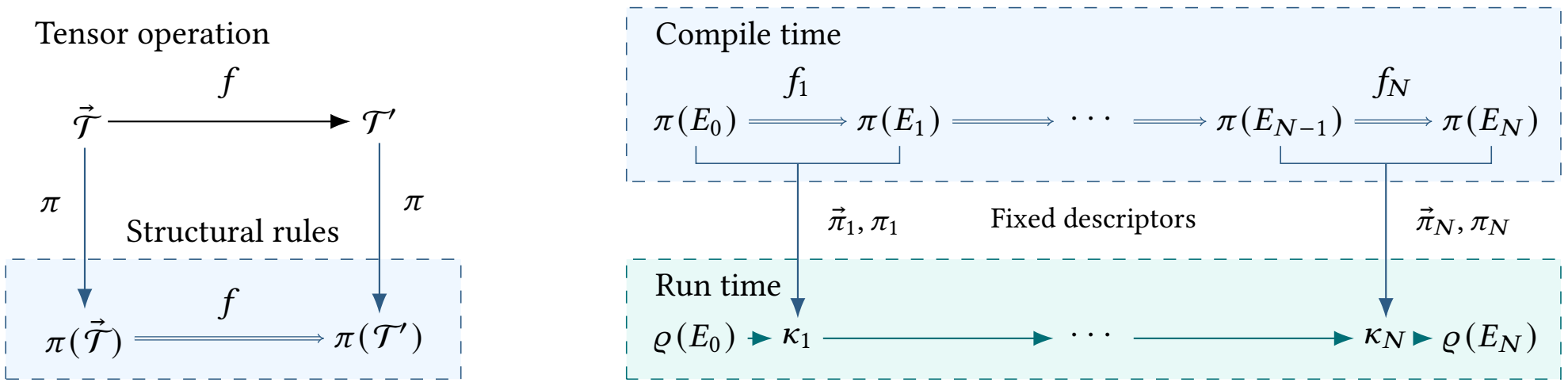

\begin{theorem}[Correctness of the structural rules]\label{thm:gb-separate}
Let $f$ be an operation defined by the rules of
Figs.~\ref{fig:gb-elementwise-rules} and~\ref{fig:gb-tensor-rules} or
the multiplication construction of Appendix~\ref{sec:gb-matmul}.
Under the contracts above, structural evaluation of compatible operand
descriptors $\vec\pi$ terminates with
$\vec\pi\xRightarrow{f}(\pi',\kappa)$, independently of payloads.
For every tensor tuple $\vec{\mathcal T}$ with
$\descr(\vec{\mathcal T})=\vec\pi$ whose dense computation is defined,
executing $\kappa(\payload(\vec{\mathcal T}))$ terminates with a payload
$\rho'$ compatible with $\pi'$ such that
\[
  \sem{\operatorname{Pack}(\pi',\rho')}
    =f(\sem{\vec{\mathcal T}})
    =\sem{f_{\mathrm{gb}}(\vec{\mathcal T})}.
\]
\end{theorem}
The equality concerns tensor values. The structural rules retain
redundant blocks that the existing operation may remove, so their result
descriptor requires no numerical tests. Fig.~\ref{fig:gb-square} shows
the descriptor computation for the rule-defined operation.

\pheader{Composition of tensor operations}
To extend the theorem to a sequence, we must determine each operation's
result descriptor and numerical calls before evaluating the operation on
its input payloads.
The following definition states these requirements using the judgment
defined in \S\ref{sec:gb-algorithms}.

\begin{definition}[Supported operations]\label{def:gb-supported}
An operation $f$ with fixed parameters and policies is \emph{supported}
if, for every compatible input descriptor tuple $\vec\pi$, structural
evaluation determines $\vec\pi\xRightarrow{f}(\pi',\kappa)$ in finitely
many steps without reading payloads, and the following hold for every
$\vec{\mathcal T}$ with $\descr(\vec{\mathcal T})=\vec\pi$ whose dense
computation is defined.
\begin{enumerate}[topsep=2pt,itemsep=1pt,label=(\roman*)]
  \item \emph{Dense equivalence:} $f(\vec{\mathcal T})$ terminates and
    $\sem{f(\vec{\mathcal T})}=f(\sem{\vec{\mathcal T}})$.
  \item \emph{Descriptor separation:}
    $\descr(f(\vec{\mathcal T}))=\pi'$.
  \item \emph{Fixed numerical calls:} $\kappa$ has a fixed finite sequence
    of numerical calls and fixed structural index arguments. On
    $\payload(\vec{\mathcal T})$, it terminates with
    $\payload(f(\vec{\mathcal T}))$.
\end{enumerate}
\end{definition}
Theorem~\ref{thm:gb-separate} establishes these properties
for the rule-defined operations. Fig.~\ref{fig:gb-stage} shows how they
allow structural computation to proceed through a sequence before any
numerical calls execute. Let $E_j$ contain the inputs and first $j$
results of a fixed sequence $f_1,\ldots,f_N$, with each operation reading
inputs or earlier results. Starting from $\descr(E_0)$, each structural
judgment adds the next result descriptor to the environment and records
its calls $\kappa_j$. Condition~(ii) permits later operations to use this
descriptor without computing the payload. Condition~(iii) supplies the
calls to execute once the payloads are available. \S\ref{sec:spec}
formalizes this composition for tensor programs, including branches and
loops, and proves correctness of the generated code.

\section{Structural Execution and Specialization}\label{sec:spec}\label{sec:spec-layer-schedule}
\S\ref{sec:gb-format} separates structural and numerical computation
when the sequence of tensor operations is fixed. To apply this result
to an abstract interpreter, we must determine which operations it
executes without computing numerical values. Let $P$ be a tensor program that \neuton generates from a \cf
specification.  Numerical inputs $\mathbf b$ contain
weights, biases, input region bounds and perturbation
radii. The architecture determines the layer sequence and the
transformer applied at each layer. Given a fixed DNN architecture, we unroll these calls and inline
them in topological order. A transformer may contain loops
that traverse the DNN and depend on computed coefficients. So, we first replace DNN
traversals that select neurons using coefficients with traversals that
select them using structural information. Then, we use the structural semantics on the transformed program to generate the numerical code.


\subsection{Specializing DNN Traversals}\label{sec:gb-relax}
DNN certifiers often refine a bound by traversing the network and
replacing neuron terms with bounds over other neurons. The traversal
repeats these substitutions until it reaches a stopping condition.
Backsubstitution in \deeppoly~\cite{deeppoly} and the backward bounding
pass used in \crownibp~\cite{crownibp} follow this pattern. We consider a traversal
that selects terms with nonzero coefficients and stops when none require
further substitution. Each selection therefore depends on coefficients
computed by earlier iterations. We replace these numerical tests with
selections based on tensor descriptors and DNN connectivity. 

\pheader{Numerical traversal}\label{sec:generated-certifiers}
Write $E=(E_i)_i$, where $E_i=d_i+\sum_j c_{i,j}n_j$ is a bound
expression for neuron $i$, and let $C=(c_{i,j})$ be the coefficient tensor.
These expressions describe the rows of a tensor updated
by \neuton. Each iteration applies the replacements to all rows together.
The neuron indices below specify the entries affected by this tensor
computation. They do not introduce a separate execution for each neuron.
The function $g(n,c)$ replaces the term $c n$. The priority
$\mathit{pri}$ orders substitutions, and $\mathit{stop}$ identifies
neurons requiring no further substitution. A traversal direction selects
which network edges to follow. In \cf, these four components specify
the \traverse construct. A term is active when its coefficient is nonzero
and its neuron has not reached the stopping condition, giving
\mbox{$A_i(E)=\{j:c_{i,j}\ne0\land\neg\mathit{stop}(n_j)\}$}.
The loop selects the highest priority among all active terms. For
nonempty $\bigcup_i A_i(E)$, let
$p(E)=\max_{j\in\bigcup_i A_i(E)}\mathit{pri}(n_j)$ and
$\Sigma_i(E)=\{j\in A_i(E):\mathit{pri}(n_j)=p(E)\}$.
One iteration computes
$E'_i=d_i+\sum_{j\notin\Sigma_i(E)}c_{i,j}n_j
          +\sum_{j\in\Sigma_i(E)}g(n_j,c_{i,j})$.
The loop ends when all active sets are empty. We require
$\mathit{pri}$ and $\mathit{stop}$ to depend only on structure. 
Any numerical choices within $g$, such as
selecting lower or upper bounds by coefficient sign, remain at runtime.
We also require the following replacement conditions.

\begin{definition}[Replacement conditions]\label{def:zero-preserving}\label{def:replacement-reach}
A replacement $g$ is \emph{zero-preserving} if $g(n,0)=0$ for every
neuron $n$. It \emph{follows the DNN} if every nonstopped neuron $n_j$
mentioned by $g(n,c)$ is reachable from $n$ in the traversal direction
by a nonempty path whose intermediate neurons are unstopped and have
priority strictly greater than $\mathit{pri}(n_j)$.
\end{definition}

\pheader{Structural traversal}
Assume these conditions and a finite acyclic network. The structural
traversal maintains a set $B$ of neuron positions to visit. It selects
the highest-priority positions and replaces them with their neighbors. Let
$\mathcal U=\{j:\neg\mathit{stop}(n_j)\}$ and let $\mathrm{nbr}(J)$
be the neighbors of $J$ in the traversal direction. Initialize
$B^0=\mathcal U\cap\mathrm{cols}(\descr(C))$, where $\mathrm{cols}$
contains all columns covered by stored blocks, or all columns when the
background is nonzero. For nonempty $B^m$, select and update
\begin{align}
  \Sigma^m&=\{j\in B^m:\mathit{pri}(n_j)
                     =\max_{k\in B^m}\mathit{pri}(n_k)\},
       \label{eq:structural-selection}\\
  B^{m+1}&=(B^m\setminus\Sigma^m)
       \cup(\mathrm{nbr}(\Sigma^m)\cap\mathcal U).
       \label{eq:structural-update}
\end{align}
We apply the numerical traversal's update with $\Sigma^m$ in place of
each $\Sigma_i(E)$ and stop when $B$ is empty. The traversal therefore
selects substitutions without computing coefficients. A stored bound
may skip layers, so an active term need not yet belong to $B$. However,
it remains reachable from $B$ through unstopped positions of higher
priority. This ordering ensures that whenever a structural iteration
selects a nonzero coefficient, it selects exactly the original loop's
maximum-priority active terms. The additional substitutions all have
coefficient zero and leave the expression unchanged.
Appendix~\ref{app:gb-relax} proves that the traversal covers the active
terms, preserves the bound expressions, and terminates.

Let $\sigma$ collect the fixed interpreter and architecture, input and
batch shapes, element type, and structural constants and policies.
Combining the fixed layer sequence with the structural traversals gives
$P_s=\operatorname{Structuralize}(P,\sigma)$. Its tensor operations retain
redundant blocks so that numerical tests do not change their descriptors.
These transformations preserve tensor values and bound expressions.
The original program $P$ observes tensors through their dense values,
so $\sem{P_s(\mathbf b)}=\sem{P(\mathbf b)}$ when both executions are
defined. The program $P_s$ still contains traversal loops, but their
selection and stopping decisions now depend on structural information.
The remaining code consists of finite sequences of tensor operations,
assignments, and conditionals. Structural execution will evaluate these
loops while generating the code for their numerical computations.

\subsection{Generating Specialized Code by Structural Execution}\label{sec:gb-dd}\label{sec:gb-closedform}
We now specialize the transformed program $P_s$ from \S\ref{sec:gb-relax}
to the fixed inputs $\sigma$. We first consider programs whose remaining
control depends only on structural information. Numerical decisions
that remain in program control are handled by specializing the regions
they invoke, as discussed below. The complete specialization is written
$(P,\sigma)\leadsto G_\sigma$, where $G_\sigma$ is executable tensor
code. Each certification run supplies numerical inputs $\mathbf b$ to
this code to compute its bounds. We define structural execution using
big-step rules for expressions and statements, following the tensor
execution semantics of \neuton~\cite{neuton}. This is partial
evaluation~\cite{pe}, with structural computations evaluated during
specialization and numerical computations recorded for later execution.

\pheader{Structural environment}\label{def:gb-sstate}
We distinguish tensor variables $x$ from structural variables $z$.
The environment $\Gamma$ maps each tensor variable to its descriptor and
each structural variable to its known value. These values include layer
indices and the sets maintained by structural traversals. The fixed
inputs determine the initial environment $\Gamma_0(\sigma)$. A structural
expression $a$ reads known values or descriptor fields from $\Gamma$.
Its value, written $\sem a_\Gamma$, must agree with the corresponding
computation in $P_s$ for every compatible numerical input. Such
expressions determine operation parameters, update traversal variables,
and evaluate guards. Numerical decisions within tensor operations,
such as selecting bounds by coefficient sign, remain in the generated
code.

\begin{definition}[Descriptor-determined program]\label{def:gb-dguard}
A program is \emph{descriptor-determined} if its structural environment
determines all guards, structural assignments, and tensor-operation
parameters at every reached statement, independently of numerical payloads.
\end{definition}

\pheader{Tensor expressions}\label{sec:gb-expressions}
We first define specialization of a tensor expression $i_e$. It reads
$\Gamma$ and produces a result descriptor $\pi$ and an expression
$i_e'$ that computes the result's payload, written
  $\Gamma\vdash i_e\leadsto(\pi,i_e')$.
During
specialization, a tensor variable denotes its descriptor in $\Gamma$.
In $i_e'$, the same variable denotes its payload. Expression
specialization leaves $\Gamma$ unchanged. The judgment extends to tuples.

\suppressfloats[t]
\begin{figure}[t]
    \centering
    \setlength{\abovecaptionskip}{3pt}
    \begingroup\small
    \input{figures/gb-expression-rules.tex}
    \endgroup
    \caption[Specialization of tensor expressions]{Tensor-expression
        rules. $f_\Gamma$ instantiates $f$'s structural parameters from $\Gamma$.}
    \Description{The variable rule obtains a descriptor and retains the
        variable for numerical execution. The operation rule specializes
        the operands and applies the structural tensor-operation rules.}
    \label{fig:gb-expression-rules}
\end{figure}

Figure~\ref{fig:gb-expression-rules} gives the variable and operation
rules. \textsc{E-Var} obtains a variable's descriptor and retains its
name in the generated expression. \textsc{E-Op} specializes the operand
expressions to obtain their descriptors $\vec\pi$ and generated
expressions $\vec i_e'$. It then applies the operation judgment of
\S\ref{sec:gb-algorithms}, with structural parameters instantiated from
$\Gamma$, to obtain $(\pi',\kappa)$. The generated expression
$\kappa(\vec i_e')$ evaluates its arguments once in order and executes
the recorded calls. Its structural parameters remain fixed even if
$\Gamma$ changes later. Here $f$ ranges over supported operations
(Definition~\ref{def:gb-supported}). Appendix~\ref{app:gb-expressions}
gives the constant rule and proves expression correctness using the
results of \S\ref{sec:gb-staging}.

\pheader{Statements}
A statement $i_s$ can assign variables, define sequences, or affect control flow. Specializing $i_s$ yeilds numerical code
$K$ and an updated structural environment $\Gamma'$.
  $\Gamma\vdash i_s\leadsto(\Gamma',K).$
We write $\epsilon$ for empty code and $K_1\cdot K_2$ for execution of
$K_1$ followed by $K_2$. Figure~\ref{fig:gb-program-rules} defines the
rules. In \textsc{S-Assign}, the expression judgment produces the
assigned tensor's descriptor and numerical expression $i_e'$. The rule
updates $\Gamma(x)$ and records the assignment $x\gets i_e'$.
\textsc{S-Static} instead evaluates an assignment to a structural
variable, updates its known value, and emits no code. \textsc{S-Seq}
uses the first statement's updated environment to specialize the second
and concatenates their generated code.

\begin{figure}[t]
    \centering
    \setlength{\abovecaptionskip}{3pt}
    \begingroup\small
    \input{figures/gb-program-rules.tex}
    \endgroup
    \caption[Specialization of tensor statements]{Statement
        rules. Guards and structural assignments are evaluated during specialization.}
    \Description{Rules specialize a tensor assignment using the expression
        semantics, evaluate a structural assignment, compose two statements,
        and handle both branches of a conditional and both loop cases.}
    \label{fig:gb-program-rules}
\end{figure}

\textsc{S-While-T} applies when the conditio n is true in $\Gamma$.
It specializes the body, then the remaining loop using the updated
environment. Its numerical code is the current iteration's code followed
by that of the remaining iterations. \textsc{S-While-F} ends this
process when the condition is false. Thus the generated code retains
the numerical computations in order while the structural loop control
is evaluated during specialization. Repeated assignments to a tensor
variable are permitted. Each generated assignment evaluates its entire
right-hand side before updating the payload variable.
Appendix~\ref{app:gb-dd} gives the empty-statement and ite rules. The traversal
termination result of \S\ref{sec:gb-relax} ensures that the loops of
$P_s$ have finite specialization derivations.

\pheader{Specialized program}
Write $P_s=(i_s,\vec o)$ for the transformed program's body and returned
tensor variables. Specializing the body from $\Gamma_0(\sigma)$ produces
its numerical code $K$ and final environment $\Gamma_f$. We append a
return that reconstructs the output tensors from their descriptors and
payloads.
\begin{equation}\label{eq:gb-specialize}
\frac{
 P_s=\operatorname{Structuralize}(P,\sigma)=(i_s,\vec o)
 \qquad
 \Gamma_0(\sigma)\vdash i_s\leadsto(\Gamma_f,K)
}{
 (P,\sigma)\leadsto
 K\cdot[\mathbf{return}\ \operatorname{Pack}(\Gamma_f(\vec o),\vec o)]
}.
\end{equation}
The final executable specialized program $G_\sigma$ takes  input variables $\mathbf b$, executes the numerical calls in $K$, then returns the
output tensors reconstructed from their computed payloads. The output descriptors $\Gamma_f(\vec o)$ and all
structural parameters are embedded as constants in the generated code. This exposes a sequence of purely numerical calls to \texttt{torch.compile}~\cite{pytorch2} for kernel fusion and elimination
of intermediate tensors. We can further optimize the execution by performing additional static analyses to reduce the memory footprint. Appendix~\ref{app:optimizations} gives these analyses
and their proofs.
%
%
If an interpreter uses numerical bounds to choose which analysis to
perform, we retain these decisions.
For example, \boxpoly is a \deeppoly variant
implemented in auto\_LiRPA and GPUPoly and supported by \neuton.
It keeps interval bounds for neurons whose ReLU is stable for every
input region and refines the others by backsubstitution. 
\sys preserves each region's output by preserving the numerical decisions.Appendix~\ref{app:spec-implementation} gives complete details.


\subsection{Correctness and Reuse}
We establish correctness by proving the traversal transformation and the specialization rules both preserve the bounds. The
following theorem establishes the result for full specialization.

\begin{theorem}[Specialization]\label{thm:gb-spec}
Fix specialization inputs $\sigma$. Let $P$ be a \neuton-generated tensor
program with the finite layer schedule and traversal conditions of
\S\ref{sec:generated-certifiers}, and set
$P_s=\operatorname{Structuralize}(P,\sigma)$. Suppose $P$'s tensor
operations compute their dense counterparts and $P_s$ is
descriptor-determined with supported operations. Then
$(P,\sigma)\leadsto G_\sigma$ has a finite derivation, and
\[
  \sem{G_\sigma(\mathbf b)}=\sem{P_s(\mathbf b)}=\sem{P(\mathbf b)}
\]
for every compatible input $\mathbf b$ on which $P$ and $P_s$ have
defined dense executions in exact arithmetic. The numerical calls and
output descriptors of $G_\sigma$ are fixed by $\sigma$.
\end{theorem}

Appendix~\ref{app:gb-relax} proves that the control transformation
preserves bound expressions and terminates. Lemma~\ref{lem:spec-rules}
proves by induction on the big-step derivation that executing the
generated code reconstructs the tensors computed by $P_s$. It uses
Lemma~\ref{lem:gb-expression-correct} for tensor expressions. Together,
these results give equivalence with \neuton's tensor execution.
\neuton's semantic-preservation result supplies the connection to the
original \cf specification.

\section{Evaluation}\label{sec:spec-eval}
We evaluate \scepter on eight abstract interpreters, including commonly used interval, zonotope, and polyhedral analyses and variants with different transformers and backsubstitution procedures. The evaluation uses 44 benchmark DNNs, ranging from small fully connected networks to convolutional and residual networks with millions of neurons. We first compare GPU runtime and scalability, followed by CPU runtime, memory use, and CPU--GPU interaction. We then examine how the runtime gains affect repeated verification through an input-splitting case study.

\pheader{Certifiers, Baselines, and Benchmarks}
The eight abstract interpreters are IBP, CROWN-IBP, \deepz, \deeppoly, BoxPoly, SkipPoly, ReuseCert, and ZID. They use different abstract domains and transformers, and include analyses that combine domains or reuse intermediate bounds. BoxPoly is the variant of \deeppoly that uses interval bounds to identify stable ReLU neurons and avoids backsubstitution for these neurons. Its decisions depend on numerical bounds, so its specialized program retains some CPU computation. All eight interpreters use the same specialization procedure. We compare \scepter with Neuton, auto\_LiRPA, IVAN, LUNA and GPUPoly on the GPU, and additionally include ERAN on the CPU. Table~\ref{tab:eval-baselines} lists the analyses and devices included in each comparison. The networks come from the ERAN~\cite{eran} and VNN-COMP benchmark suites~\cite{cifar100-tinyimagenet-resnets,vgg16-vnncomp} and the \texttt{robustness} library~\cite{robustness}, and include fully connected, convolutional, and residual architectures. We group them by neuron count using the four ranges in Table~\ref{tab:eval-size-groups}. The two very large networks are ResNet50 with $2.94$M neurons and VGG-16 with $13.6$M neurons. Table~\ref{tab:eval-networks} lists all networks and maps the plot indices to their original identifiers.

\begin{table}[t]
\centering\footnotesize
\caption{Abstract interpreters and devices evaluated for each system, and the network size groups. Dots mark evaluated configurations and dashes indicate no comparison.}
\label{tab:eval-baselines}\label{tab:eval-size-groups}
\setlength{\tabcolsep}{2pt}
\begin{tabular*}{\linewidth}{@{\extracolsep{\fill}}lcccccccccc@{}}
\toprule
& \multicolumn{8}{c}{Abstract interpreter} & \multicolumn{2}{c}{Device}\\
\cmidrule(lr){2-9}\cmidrule(l){10-11}
System & \deeppoly & \deepz & BoxPoly & CROWN-IBP & ZID & ReuseCert & SkipPoly & IBP & CPU & GPU\\
\midrule
\scepter & $\bullet$ & $\bullet$ & $\bullet$ & $\bullet$ & $\bullet$ & $\bullet$ & $\bullet$ & $\bullet$ & $\bullet$ & $\bullet$\\
Neuton & $\bullet$ & $\bullet$ & $\bullet$ & $\bullet$ & $\bullet$ & $\bullet$ & $\bullet$ & $\bullet$ & $\bullet$ & $\bullet$\\
auto\_LiRPA & $\bullet$ & -- & $\bullet$ & $\bullet$ & -- & -- & -- & $\bullet$ & $\bullet$ & $\bullet$\\
GPUPoly & $\bullet$ & -- & $\bullet$ & -- & -- & -- & -- & -- & -- & $\bullet$\\
IVAN & $\bullet$ & $\bullet$ & -- & -- & -- & -- & -- & $\bullet$ & $\bullet$ & $\bullet$\\
LUNA & -- & -- & $\bullet$ & $\bullet$ & -- & -- & -- & $\bullet$ & $\bullet$ & $\bullet$\\
ERAN & $\bullet$ & $\bullet$ & -- & -- & -- & -- & -- & -- & $\bullet$ & --\\
\bottomrule
\end{tabular*}
\par\smallskip
\begin{tabular*}{\linewidth}{@{\extracolsep{\fill}}lcccc@{}}
Group & Small & Medium & Large & Very large\\
Neurons & $<10\mathrm{K}$ & $10\mathrm{K}$--$100\mathrm{K}$ & $100\mathrm{K}$--$1\mathrm{M}$ & $>1\mathrm{M}$\\
Networks & 28 & 7 & 7 & 2\\
\bottomrule
\end{tabular*}
\end{table}

For each network, we compute bounds for $L_\infty$ perturbations of test inputs. We compute speedup as the baseline runtime divided by \scepter's runtime, and report geometric means using only networks on which both systems complete. We also report the largest networks completed.

\pheader{Experimental Setup}
The GPU experiments use an NVIDIA A100 with 80\,GB of device memory on a
server with two Intel Xeon Platinum 8358 processors and 256\,GB of host
memory. The CPU experiments use two 28-core Intel Xeon Gold 6348
processors and 1\,TB of RAM, with each run pinned to four physical cores
and a 120\,GB memory budget. 
Fig.~\ref{fig:scalability} lists the network configurations and batch
sizes, and Table~\ref{tab:ch8-runtime-configs} gives per-network runtimes. For each run, we set a timeout of $1000$ seconds.

\subsection{Runtime and Scalability}\label{sec:runtime-performance}\label{sec:scalability}
\makeatletter\@nobreakfalse\everypar{}\makeatother  

\begin{wrapfigure}{r}{0.42\linewidth}
\centering
\includegraphics[width=0.96\linewidth,alt={Horizontal paired bars for \deeppoly, BoxPoly, \deepz, ReuseCert and SkipPoly on a log scale of neurons: the largest network the best baseline completes and the largest network \scepter completes, each with its network name at its end (baselines ResNet34, ResNet50, Skip, ResNet34 and ConvSuper; \scepter VGG-16, VGG-16, ResNet18, ResNet50 and VGG-16), and the ratio printed in \scepter's bar: 14, 4.6, 7.8, 3.0 and 153 times.}]{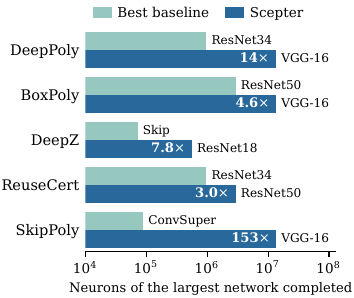}
\caption[Largest network completed by \scepter and by the best baseline]{Largest network completed per certifier. Multipliers indicate
the increase in network size over the best baseline.}
\label{fig:scalability}
\end{wrapfigure}

\pheader{Scalability}
Scepter enables precise abstract interpreters to process networks with up to $153\times$ as many neurons as the largest networks completed by their baselines. Fig.~\ref{fig:scalability} compares the largest network completed by Scepter with the largest completed by any baseline for each analysis. Scepter runs DeepPoly, BoxPoly, and SkipPoly on VGG-16 with $13.6$M neurons, increasing the largest completed network by $14\times$, $4.6\times$, and $153\times$, respectively. These gains extend across abstract domains and backsubstitution procedures. \deepz reaches ResNet18 with $558$K neurons, a $7.8\times$ increase over its best baseline, while ReuseCert reaches ResNet50, a $3.0\times$ increase. Across the suite, Scepter completes 20 certifier--network pairs that no corresponding baseline completes. In particular, it completes the widely used DeepPoly analysis on both very large networks, VGG-16 and ResNet50, whereas none of the evaluated DeepPoly baselines completes either.

We additionally compare Scepter with $\alpha,\!\beta$-CROWN~\cite{abcrown} to assess scalability on the two largest networks in our suite, ResNet50 and VGG-16. $\alpha,\!\beta$-CROWN is a state-of-the-art DNN verifier that goes beyond abstract interpretation. Even this verifier fails to complete either experiment. It exhausts GPU memory on ResNet50 and times out on VGG-16, while Scepter certifies both.

\begin{figure}[t]
\centering
\includegraphics[width=\linewidth]{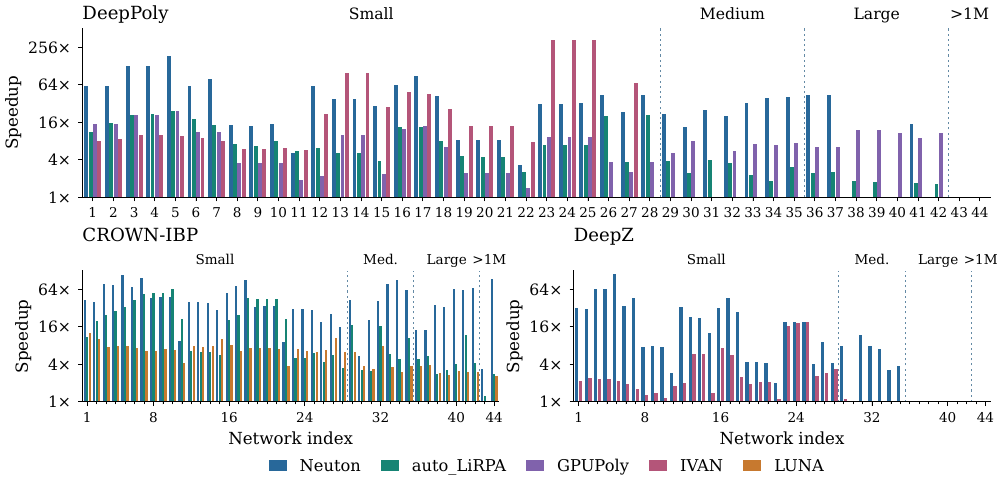}
\caption{GPU speedups for DeepPoly (top), CROWN-IBP (bottom left), and \deepz (bottom right). Networks are numbered as in Table~\ref{tab:eval-networks}. A missing bar means the baseline or Scepter does not complete that network.}
\Description{A full-width DeepPoly bar plot appears above side-by-side CROWN-IBP and \deepz plots. The lower panels use two bar positions per network for Neuton and auto LiRPA, and Neuton and IVAN, respectively. All measured speedups and network positions are retained.}
\label{fig:speedup-gpu}
\end{figure}

\pheader{GPU Runtime Speedup}
Program specialization achieves substantial speedups over GPU libraries with implementations tailored to individual analyses. For the widely used \deeppoly analysis, \scepter achieves geometric-mean speedups of $5.69\times$ over auto\_LiRPA, $6.57\times$ over GPUPoly, $22.3\times$ over IVAN, and $30.7\times$ over Neuton. Table~\ref{tab:ch8-main} reports the results for all eight analyses, with ResNet50 and VGG-16 in separate columns. Fig~\ref{fig:speedup-gpu} shows per-network results for \deeppoly, CROWN-IBP, and \deepz, representing polyhedral, hybrid, and zonotope analyses. The remaining five per-network plots are in the appendix\ref{app:spec-eval}.

\begin{table}[t]
\centering\footnotesize
\caption{GPU speedups (geometric means) on networks completed by both systems, with the number of such networks in parentheses. A cross (\sconly) marks networks that the baseline does not complete, and a dash (\neither) networks that neither system completes.}
\label{tab:ch8-main}
\setlength{\tabcolsep}{6pt}\setlength{\aboverulesep}{0pt}\setlength{\belowrulesep}{0pt}\renewcommand{\arraystretch}{1.15}
\begin{tabular}{ll?rrr?cc?r}
\toprule
& & \multicolumn{3}{c?}{Size groups} & \multicolumn{2}{c?}{Very large} & \multirow{2}{*}{Overall (44)}\\
\cmidrule{3-5}\cmidrule{6-7}
Certifier & Baseline & Small (28) & Medium (7) & Large (7) & ResNet50 & VGG-16 & \\
\midrule

\deeppoly & Neuton & $32.00$ (28) & $25.79$ (7) & $30.68$ (3) & \nopair & \nopairdiv & $30.65$ \\
 & auto\_LiRPA & $8.48$ (28) & $2.87$ (7) & $1.95$ (6) & \nopair & \nopairdiv & $5.69$ \\
 & GPUPoly & $6.02$ (28) & $6.61$ (6) & $9.32$ (7) & \nopair & \nopairdiv & $6.57$ \\
 & IVAN & $22.29$ (26) & \nopair & \nopairdiv & \nopair & \nopairdiv & $22.29$ \\
\grpline
BoxPoly & Neuton & $42.10$ (28) & $63.90$ (6) & $31.50$ (2) & \nopair & \nopairdiv & $44.41$ \\
 & auto\_LiRPA & $11.97$ (28) & $5.40$ (7) & $3.17$ (7) & \nopair & \nopairdiv & $8.40$ \\
 & GPUPoly & $4.45$ (28) & $1.75$ (6) & $2.36$ (7) & $6.71$ & \nopairdiv & $3.54$ \\
 & LUNA & $4.03$ (28) & $2.80$ (7) & $3.28$ (6) & \nopair & \nopairdiv & $3.68$ \\
\grpline
CROWN-IBP & Neuton & $39.48$ (28) & $36.72$ (7) & $35.26$ (7) & $93.11$ & $3.38$ & $36.96$ \\
 & auto\_LiRPA & $17.34$ (28) & $7.12$ (7) & $4.70$ (7) & $2.77$ & $1.26$ & $11.05$ \\
 & LUNA & $7.31$ (28) & $4.29$ (7) & $3.18$ (7) & $2.62$ & \nopairdiv & $5.71$ \\
\grpline
IBP & Neuton & $13.92$ (28) & $11.01$ (7) & $8.78$ (7) & $17.68$ & $1.48$ & $11.91$ \\
 & auto\_LiRPA & $25.65$ (28) & $8.72$ (7) & $5.01$ (7) & $3.02$ & $1.22$ & $14.81$ \\
 & IVAN & $1.59$ (28) & $1.44$ (3) & \nopairdiv & \nopair & \nopairdiv & $1.58$ \\
 & LUNA & $10.85$ (28) & $5.67$ (7) & $3.87$ (7) & $2.94$ & \nopairdiv & $8.01$ \\
\grpline
\deepz & Neuton & $15.34$ (28) & $6.35$ (6) & \nopairnotediv & \nosys & \nosysdiv & $13.1$ \\
 & IVAN & $2.92$ (28) & $1.10$ (1) & \nopairnotediv & \nosys & \nosysdiv & $2.83$ \\
\grpline
SkipPoly & Neuton & $16.67$ (28) & $26.39$ (7) & \nopairdiv & \nopair & \nopairdiv & $18.3$ \\
\grpline
ReuseCert & Neuton & $14.00$ (28) & $13.27$ (7) & $5.86$ (7) & \nopair & \nosysdiv & $12.00$ \\
\grpline
ZID & Neuton & $38.83$ (28) & $36.82$ (7) & $28.92$ (7) & $80.73$ & $3.59$ & $35.38$ \\
\bottomrule
\end{tabular}
\end{table}

\scepter improves both runtime and scalability over the baselines. On networks completed by auto\_LiRPA, the geometric-mean speedups over it are $5.69\times$ for \deeppoly, $8.40\times$ for BoxPoly, $11.0\times$ for CROWN-IBP, and $14.8\times$ for IBP, with improvements on every network for each analysis. Against GPUPoly, the speedups are $6.57\times$ for \deeppoly and $3.54\times$ for BoxPoly. Similarly, \scepter achieves speedups over IVAN of $22.3\times$ for \deeppoly, $2.83\times$ for \deepz, and $1.58\times$ for IBP. It is faster on every network for \deeppoly and \deepz. Against LUNA, \scepter is $8.01\times$ faster on \ibp, $5.71\times$ on \crownibp, and $3.68\times$ on \boxpoly. Finally, against \neuton, \scepter is faster on every network across all eight analyses, with geometric-mean speedups ranging from $11.9\times$ to $44.4\times$. In this evaluation, we used \texttt{torch.compile} to optimize \scepter's specialized code. In Appendix~\ref{app:specialization-ablation}, we show that i) specialization alone accounts for most of the speedup, ii) even without \texttt{torch.compile}, \scepter remains faster than the baselines, and iii) \texttt{torch.compile} without specialization is not effective.

\pheader{CPU Runtime}
We evaluate CPU runtime on the the small, medium, and large networks. GPUPoly targets GPUs and has no CPU implementation, so we compare with Neuton, auto\_LiRPA, IVAN, and LUNA and with ERAN's CPU implementations of \deeppoly and \deepz. Across these 42 networks, \scepter achieves overall geometric-mean speedups of $5.27\times$ over Neuton, $2.30\times$ over auto\_LiRPA, $4.49\times$ over IVAN, $3.48\times$ over LUNA and $182\times$ over ERAN. These comparisons use the certifier--network pairs completed by both systems. Table~\ref{tab:ch8-cpu} in the appendix reports the results for each analysis and baseline.

The largest gains are against ERAN, with geometric-mean speedups of $87.4\times$ for \deeppoly and $522\times$ for \deepz. For \deeppoly, \scepter is also $5.58\times$ faster than Neuton, $2.04\times$ faster than auto\_LiRPA, and $19.7\times$ faster than IVAN. For \deepz, the corresponding speedups over Neuton and IVAN are $4.02\times$ and $4.66\times$. The gains extend to interval and hybrid analyses, with IBP and CROWN-IBP speedups of $5.30\times$ and $7.74\times$ over Neuton, respectively. Against LUNA, \scepter is $2.90\times$ faster for BoxPoly and CROWN-IBP and $5.01\times$ faster for IBP.

\subsection{Memory Use and CPU--GPU Interaction}\label{sec:communication-overhead}
\pheader{Memory Use} \scepter reduces peak GPU memory by geometric-mean factors of $2.60\times$ relative to Neuton, $1.32\times$ relative to auto\_LiRPA, $4.51\times$ relative to IVAN, and $1.74\times$ relative to LUNA, over configurations completed by both systems (Table~\ref{tab:app-profiles}). For \deeppoly alone, the reductions are $3.14\times$, $1.40\times$, and $11.2\times$, respectively (Table~\ref{tab:eval-profiles}). \scepter also completes networks on which these baselines exhaust device memory. We omit GPUPoly from the memory comparison because it processes inputs individually, while the other systems process input batches, so their peak memory measurements do not represent the same workload.

\begin{wraptable}{r}{0.55\linewidth}
\centering\footnotesize
\caption{\deeppoly memory use and execution profiles. Ratios compare the baseline with \scepter. Copies are median D2H and H2D copies per batch (baseline / \scepter), across networks profiled both systems.}
\label{tab:eval-profiles}
\setlength{\tabcolsep}{3pt}
\begin{tabular}{lrrr@{\hspace{8pt}}rr}
\toprule
& \multicolumn{3}{c}{Reduction} & \multicolumn{2}{c}{Copies (median)}\\
\cmidrule(lr){2-4}\cmidrule(l){5-6}
Baseline & Memory & Idle time & Transfer & D2H & H2D\\
\midrule
Neuton & $3.14\times$ & $128\times$ & $2.34\times$ & $127$ / $0$ & $35$ / $16$\\
auto\_LiRPA & $1.40\times$ & $14.2\times$ & $11.3\times$ & $281$ / $0$ & $77$ / $16$\\
IVAN & $11.2\times$ & $2.74\times$ & $2.83\times$ & $6$ / $0$ & $41$ / $14$\\
\bottomrule
\end{tabular}
\end{wraptable}
\pheader{CPU--GPU Interaction} We profile execution separately from the runtime experiment and measure GPU idle time as the part of the profiled run where no kernel, copy, or memory-set operation occurs. Across all analyses, \scepter reduces GPU idle time by geometric-mean factors of $52.7\times$ relative to Neuton, $11.4\times$ relative to auto\_LiRPA, $3.02\times$ relative to GPUPoly, $1.66\times$ relative to IVAN, and $4.61\times$ relative to LUNA, and transfer time by $3.74\times$, $11.3\times$, $4.06\times$, $2.04\times$, and $3.56\times$ respectively (Table~\ref{tab:app-profiles}).  
For \deeppoly, \scepter reduces GPU idle time by geometric-mean factors of $128\times$ relative to Neuton, $14.2\times$ relative to auto\_LiRPA, and $2.74\times$ relative to IVAN, over 38, 41, and 26 networks profiled with both systems, respectively. The corresponding reductions in transfer time are $2.34\times$, $11.3\times$, and $2.83\times$ (Table~\ref{tab:eval-profiles}). The separate profiling comparisons with GPUPoly give reductions of $3.52\times$ in GPU idle time and $2.69\times$ in transfer time, respectively. For Neuton, the $128\times$ reduction in idle time is much larger than the $2.34\times$ reduction in transfer time. This is consistent with reducing the repeated CPU work between GPU operations, whose cost is not reflected in transfer time. The transfer counts further show the reduction in CPU--GPU communication. For \deeppoly, \scepter has zero median D2H copies (versus $6$--$281$ for baselines) and $14$--$16$ median H2D copies (versus $35$--$77$).

\subsection{Input Splitting Case Study}\label{sec:spec-eval-case-studies}
\makeatletter\@nobreakfalse\everypar{}\makeatother
\begin{wrapfigure}{r}{0.45\linewidth}
\centering
\includegraphics[width=\linewidth]{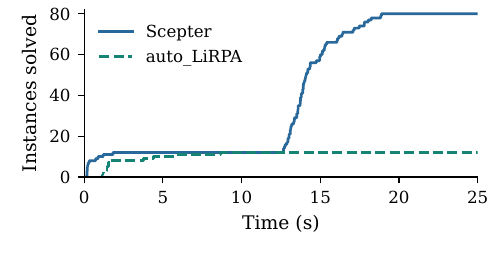}
\caption{Instances of lsnc\_relu solved by input splitting with \deeppoly.}
\Description{Over 25 seconds, \scepter solves all 80 instances, including 68 proofs and 12 counterexamples. Auto LiRPA finds the same 12 counterexamples; the other 68 instances time out.}
\label{fig:input-split}
\end{wrapfigure}
We now show a practical benefit of specializing the code of an abstract interpreter. Input splitting~\cite{reluval} is a common technique in formal verification that divides a problem into smaller subproblems and verifies them individually. In DNN certification, it partitions the input region into subregions and computes bounds on each. A property that cannot be established over the original region may become provable over these smaller regions. General implementations of DNN abstract interpreters repeat architecture-dependent computation for each call, although the DNN remains fixed throughout the splitting procedure. \scepter specializes the interpreter once for that DNN and reuses the resulting code across subregions. It can therefore devote more of the time budget to computing bounds, allowing more subregions to be processed and more properties to be established.

We use two benchmark families from VNN-COMP~\cite{vnncomp-benchmarks}. The lsnc\_relu workload contains 80 Lyapunov-stability properties of a neural quadrotor controller. The Cersyve workload checks safety certificates for nine control systems. We evaluate its 36 network--property pairs, including the 12 official competition instances. We use time budgets of 25\,s per property for lsnc\_relu and 100\,s per instance for Cersyve. An instance is solved by a proof or a counterexample.
We compare input splitting using \scepter and auto\_LiRPA as the bound-computation backends. We show \deeppoly, a widely used abstract interpreter for DNN certification, with the same splitting procedure and search settings for both backends. \scepter reuses the specialized code for all bound computations on each network. Runtime speedups are geometric means over instances solved by both backends. All reported counterexamples are validated.

Fig.~\ref{fig:input-split} shows that \scepter solves all 80 lsnc\_relu instances within 19\,s, whereas auto\_LiRPA solves only 12 within 25\,s. \scepter additionally proves the other 68 properties, all of which time out with auto\_LiRPA. \scepter processes a median of 6.27 million subregions per second, with a geometric-mean throughput improvement of $5.42\times$ over auto\_LiRPA. This higher throughput allows the splitting procedure to establish properties that remain unresolved with the baseline. On the 12 instances solved by both backends, \scepter is $5.94\times$ faster.

On Cersyve, \scepter solves 33 of 36 instances, compared with 32 for auto\_LiRPA. Both find counterexamples for 18 instances, while \scepter proves 15 properties and auto\_LiRPA proves 14. The remaining three \scepter instances and four auto\_LiRPA instances time out. Both solve all 12 official competition instances. 
Although the final solved counts are close, \scepter is $3.37\times$ faster on the 32 instances solved by both backends. The same splitting procedure therefore benefits both from proving additional properties and from completing shared instances sooner.

\section{Related Work}\label{sec:spec-related}

\pheader{Scaling DNN certification}
Existing DNN certifiers reduce the cost of precise analysis by exploiting
structure in coefficient tensors. GPUPoly restricts computation to neuron
dependence sets and uses parallel GPU algorithms~\cite{gpupoly}, while
auto\_LiRPA stores convolutional coefficients as patches~\cite{lirpagithub}.
\neuton generates tensor programs from \cf~\cite{constraintflowsas}
specifications and introduces \mbox{\gb} to exploit sparse and repeated
coefficients~\cite{neuton}. These systems still execute general programs
that determine how to represent intermediate results and which operations
to perform for the supplied DNN. 
\sys builds on
\neuton's tensor programs and representation to perform this reusable
work before certification, preserving the chosen interpreter's bounds.

\pheader{Scaling abstract interpretation}
Exploiting structure also improves the scalability of abstract
interpretation for conventional programs. Fast Polyhedra~\cite{fastpoly}
decomposes polyhedra into independent groups of variables and adapts
domain operators to this decomposition without losing precision. The
groups are maintained online as constraints change. \sys also preserves
precision, but identifies structural patterns whose evolution can be
determined from the DNN architecture before numerical execution.
EigenCFA~\cite{eigencfa} accelerates control-flow analysis by expressing
it as sparse matrix operations on GPUs. In our setting, \neuton already
supplies tensor execution. \sys removes structural computation that
remains interleaved with these numerical operations.

\pheader{Partial evaluation}
Partial evaluation~\cite{pe} evaluates computations determined by known
inputs and generates code for the remainder. Wei et al.~\cite{stagedai}
specialize abstract interpreters to a fixed program, retaining computations
that depend on dynamic analysis state. Their work studies control-flow
analyses executed on CPUs. Scaling precise DNN certification involves computations over large tensors and has CPU-GPU dependencies.
\sys uses structural semantics to derive intermediate tensor descriptors
without numerical coefficients and transforms eligible coefficient-dependent
traversals into structural ones. This allows the corresponding CPU
computation to be performed before GPU execution, producing a sequence
of numerical calls available to the tensor compiler. 

\pheader{Sparse representations and compilation}
TACO~\cite{taco} and Finch~\cite{finch} generate sparse iteration code
for given tensor computations using properties of their storage
formats. \sys instead executes descriptor-dependent analysis control
ahead of certification, deriving a sequence of numerical tensor calls
that remains valid across compatible numerical inputs. Sympiler~\cite{sympiler} predicts intermediate
sparsity using symbolic inspectors designed for particular sparse matrix
algorithms. Our structural semantics instead composes rules for tensor
operations, providing one
specialization procedure across arbitrary certifiers without a separate symbolic
inspector for each analysis.

\section{Conclusion}\label{sec:spec-discussion}\label{sec:conclusion}
We presented \sys, a framework that uses structural semantics to scale
precise DNN abstract interpreters. \sys separates structural computation
from numerical evaluation and generates executable code specialized to
a fixed architecture. The code is reusable across certification runs
and enables tensor compilation and memory planning. We prove that
specialization preserves the interpreter's bounds in exact arithmetic.
Across eight abstract
interpreters and 44 DNNs, \sys achieves runtime speedups of one to two
orders of magnitude and scales \deeppoly to ResNet50 and VGG-16,
which none of the evaluated \deeppoly baselines
completes. These runtime gains also allow input splitting to prove
additional properties within the same time budget.

\label{main-text-end}
\clearpage
\bibliographystyle{ACM-Reference-Format}
\bibliography{main}
\clearpage
\appendix
\section{Structural Semantics of g-BCSR}\label{app:representation}\label{app:gbcsr}
This appendix gives the definitions and proofs used in
\S\ref{sec:gb-format}. We first describe the block types and relate the structural block model
to \neuton's representation. We then define
the auxiliary functions used by the structural rules and explain the
result-type calculations. Finally, we prove operation correctness and descriptor propagation
through sequences. The reference procedures retain the operational
design of \neuton~\cite{neuton}. Our structural rules separate their
descriptor computation from the numerical calls.

\subsection{Tensor Representation and Block Types}\label{sec:gb-types}
For a fixed index domain $\Omega$, we abbreviate the structural type
$(\Omega,\mathcal S,\approx)$ by $(\mathcal S,\approx)$. Type comparisons
use this common domain; descriptors retain the domain of each block as
in Definition~\ref{def:gb-descriptor}. The implementation stores a core
as a tensor and represents the quotient map
$\iota\mapsto[\iota]_\approx$ by an index map $\psi$ into that core.
An index in the block's support reads its mapped core entry; other
indices in the block hold zero. Positions outside all block boxes take
the tensor background. Some physical core entries may be unused, so
the allocated core can contain more entries than the number of
sharing classes.

\subsubsection{Named Block Types}

The named block types describe the sparsity and sharing patterns of
DNN coefficient tensors. A block is indexed by a row space $\mathcal{R}$
and a column space $\mathcal{C}$, the neurons of two layers, and we index
it by a pair $\iota = (\rho, \gamma)$ with $\rho \in \mathcal{R}$ and
$\gamma \in \mathcal{C}$. Four of the six types below are defined over
arbitrary index spaces. The remaining two arise from convolutional layers
and require the spatial structure of such a layer, for which we fix the
following conventions. Write $[m]=\{0,\ldots,m-1\}$. The column space
factors as $c$ channels and $i$
spatial positions and the row space as $K$ channels and $o$ positions, so
that a position is a pair of pairs $\bigl((\kappa, a), (\gamma, y)\bigr)$
with $\kappa \in [K]$, $a \in [o]$, $\gamma \in [c]$ and $y \in [i]$. All
spatial reasoning here is one-dimensional; the two-dimensional case is the
product of two independent copies and every statement extends
componentwise.

A convolution carries three further parameters, $\theta = (k, s, p)$ for
kernel extent, stride and padding. Output position $a$ reads the $k$
consecutive input positions beginning at $as - p$, of which only those
inside $[i]$ exist, so the \emph{window} of $a$ is
\[
  \mathcal{W}_\theta(a) \;=\; \{\, as - p + d \;:\; d \in [k] \,\} \cap [i].
\]
For the first convolution of the running example, with $k = 3$, $s = 1$,
$p = 0$ and $i = 7$, this gives $\mathcal{W}(0) = \{0,1,2\}$,
$\mathcal{W}(1) = \{1,2,3\}$ and so on up to
$\mathcal{W}(4) = \{4,5,6\}$, which are the rows of the block $L_2^1$ in
Fig.~\ref{fig:gb-running}. The \emph{window pattern} $\mathcal{S}_\theta$
is the set of positions whose column lies in the window of its row,
$\{\,((\kappa,a),(\gamma,y)) : y \in \mathcal{W}_\theta(a)\,\}$, over all
channel pairs. A window is \emph{clipped} when $as - p < 0$ or
$as - p + k - 1 \geq i$, in which case
$\lvert \mathcal{W}_\theta(a) \rvert < k$, and we write
$\chi = ko - \sum_{a \in [o]} \lvert \mathcal{W}_\theta(a) \rvert$ for the
number of spatial positions removed by clipping per channel pair.
For the coefficient block of a convolutional layer,
\S\ref{app:gb-types} derives the output extent
$o = \lfloor (i + 2p - k)/s \rfloor + 1$ from the same reading. The types
below do not require it: $o$ and $i$ are the extents of the block, and $p$
may be any integer.

\begin{definition}[The \sparse block types]
\label{def:gb-named}
Write $\top$ for the set of all positions of a block's box and $=$ for the
identity relation, which relates no two distinct positions. Four types are
defined over arbitrary index spaces:
\begin{enumerate}[topsep=2pt,itemsep=3pt,label=(\alph*)]
\item $\dblock = (\top, =)$. Its support is the whole domain, and no two distinct positions
  are related. Any block whose structure matches none of the types below
  falls back to $\dblock$, whose core is the block itself and whose index
  map is the identity. It stores
  $\lvert \mathcal{R} \rvert \cdot \lvert \mathcal{C} \rvert$ values.
\item $\diagblock = \bigl(\{\,(\rho, \rho) : \rho \in \mathcal{R}\,\},
  =\bigr)$, defined when $\mathcal{R} = \mathcal{C}$. All the non-zero
  entries lie on a single diagonal, which arises when the expression of a
  neuron references exactly one other neuron, e.g., when computing the
  bounds of a \relu layer. The core stores the diagonal entries, so the
  block stores $\lvert \mathcal{R} \rvert$ values.
\item $\rblock_j = (\top, \approx_j)$, where $\approx_j$ relates two
  positions that agree on every axis but the $j$th. The block has full
  support and one slice repeated along axis $j$, which arises when an
  operand is broadcast against another of higher rank. The core stores the
  slice. We drop the subscript where the axis is clear.
\item $\cblock = (\top, \top \times \top)$. The block has full support with
  every position in a single class, which arises from a scalar literal in a
  specification. The core stores one value.
\end{enumerate}
The remaining two types are those of a convolutional layer, and use the
spatial conventions fixed above:
\begin{enumerate}[topsep=2pt,itemsep=3pt,label=(\alph*),start=5]
\item $\stblock_\theta = (\mathcal{S}_\theta, \approx_\theta)$, where
  $\bigl((\kappa,a),(\gamma,y)\bigr) \approx_\theta
   \bigl((\kappa',a'),(\gamma',y')\bigr)$ exactly when $\kappa = \kappa'$,
  $\gamma = \gamma'$ and $y - as = y' - a's$. In a convolutional layer each
  output neuron is an affine combination of a fixed-size window of the
  input, with weights shared across all spatial positions, so the
  coefficient block of such a layer has a \emph{shifted-repeat} structure:
  before clipping, each row has $k$ kernel positions per input channel,
  shifted by $as$ relative to spatial position $0$. The core stores the kernel values once and $\theta$
  encodes the shift pattern, so the core holds $Kck$ values.
\item $\wblock_\theta = (\mathcal{S}_\theta, =)$. The support of a
  convolution with no sharing. This arises for certifiers that maintain
  per-neuron polyhedral expressions over a previous layer, e.g., after
  backsubstituting through a convolutional layer: the non-zero entries of
  each row occupy the window of that row, and different rows carry
  unrelated values. The core stores the entries of every window and
  $\theta$ encodes where each window begins and ends. The block has
  $Kc(ko - \chi)$ values, which the implementation stores in a core of
  $Kcko$ entries, one window of extent $k$ per output position.
\end{enumerate}
\end{definition}

The list separates the two sources of compression. $\diagblock$ constrains
only the support, and $\cblock$ and $\rblock$ constrain only the sharing.
$\stblock_\theta$ and $\wblock_\theta$ have the same support and differ
only in $\approx$. The cost of the last pair differs by the number of
output positions: ignoring clipping, $\wblock_\theta$ stores $Kcko$ values
where $\stblock_\theta$ stores $Kck$, a ratio of $o$, which is $1024$ for a
$32 \times 32$ feature map.

\subsubsection{Index Maps and the Structural Model}\label{app:gb-types}

The named types admit compact index maps. The map sends a position in a
block's support to the core entry that stores its value. The following
maps use the convolution coordinates fixed above. For the coefficient block of a convolutional layer, the output positions
are those whose window lies within the padded input $[-p, i + p)$. The
last of them, $a = o - 1$, is the largest with $as - p + k \leq i + p$,
so $(o-1)s \leq i + 2p - k$ and
\[
  o = \lfloor (i + 2p - k)/s \rfloor + 1 .
\]
The block types do not require this equation, and
\S\ref{app:gb-restrict} uses blocks that do not satisfy it.

The index maps of the named types are then as follows.

\begin{center}\small
\begin{tabular}{@{}ll@{}}
\toprule
Type & $\psi_\Theta\bigl((\kappa,a),(\gamma,y)\bigr)$ \\
\midrule
$\dblock$        & the position itself \\
$\wblock_\theta$ & $(\kappa, a, \gamma, y - as + p)$, defined when $y \in \mathcal{W}_\theta(a)$ \\
$\stblock_\theta$ & $(\kappa, \gamma, y - as + p)$, defined when $y \in \mathcal{W}_\theta(a)$ \\
$\diagblock$     & the shared index, defined when the two agree \\
$\rblock$        & the position with the repeated axis dropped \\
$\cblock$        & the unique core index \\
\bottomrule
\end{tabular}
\end{center}

$\wblock$ and $\stblock$ differ only in whether the output index $a$ survives
into the core index. Dropping $a$ makes the same kernel entry serve every output position.

\begin{lemma}[Correspondence of the block models]\label{lem:gb-model-correspondence}
Let an index-map block on $\Omega$ have a finite core index set $Q$,
a core $c:Q\to\EReal$, and a partial map
$\psi:\Omega\rightharpoonup Q$. Its lookup returns $c[\psi(\iota)]$
when $\psi(\iota)$ is defined and zero otherwise. Set
\[
\begin{gathered}
  \mathcal S=\operatorname{dom}\psi,\qquad
  \iota\approx\iota'\ \Longleftrightarrow
     \psi(\iota)=\psi(\iota')\quad(\iota,\iota'\in\mathcal S),\\
  b[[\iota]_\approx]=c[\psi(\iota)].
\end{gathered}
\]
Then $B=(\Omega,\mathcal S,\approx,b)$ has the same array denotation
as the index-map block. Applying this construction to each block while
retaining the tensor's background, positions, and shape preserves the
tensor denotation.
\end{lemma}
\begin{proof}
The fibers of $\psi$ form the equivalence classes of $\approx$, so
$b$ is well defined. At an index in $\mathcal S$, both lookups return
$c[\psi(\iota)]$; outside $\mathcal S$, both return zero. Tensor lookup
selects the same block at each covered position and the same background
elsewhere. Physical core entries outside $\operatorname{im}\psi$ are
unobserved and need not be retained by this construction. The conclusion
therefore concerns dense values, without identifying physical core
arrays with quotient-indexed cores.
\end{proof}

\subsubsection{Type Ordering}\label{sec:gb-type-order}
\begin{lemma}
\label{lem:gb-subspace}
For a block type $\tau$, the real tensors in $\sem{\tau}$ form a linear
subspace whose dimension is the number of $\approx$-classes.
\end{lemma}

\begin{proof}
Addition and scalar multiplication preserve both constraints. Positions
outside the support remain zero, and positions in one sharing class
remain equal. The zero tensor satisfies every type. Each admitted
tensor is determined by one freely chosen real value per sharing class,
so this subspace is isomorphic to $\mathbb{R}^{N}$ for $N$ classes.
\end{proof}

The number of classes is the number of values a block of type $\tau$
holds, which can differ from the storage the implementation allocates for
it. A window is \emph{clipped} when $as - p < 0$ or
$as - p + k - 1 \geq i$, in which case
$\lvert \mathcal{W}_\theta(a) \rvert < k$, and we write
$\chi = ko - \sum_{a \in [o]} \lvert \mathcal{W}_\theta(a) \rvert$ for the
number of spatial positions removed by clipping per channel pair. For
$\wblock_\theta$ there are $Kc(ko - \chi)$ classes, and the implementation
stores a core of $Kcko$ entries, one window of extent $k$ per output
position. The core of a type can therefore have entries that no position
maps to, and widening a block need not increase the storage that the
implementation allocates. For $\stblock_\theta$ there are $Kc$ classes for each
kernel offset $d \in [k]$ that some output position reads, and the
implementation stores the whole kernel, $Kck$ entries. The two counts
differ when the input is too small for some offset to be read: with
$i = o = 1$ and $\theta = (3, 1, 1)$, only the middle offset is read, and
there are $Kc$ classes.

Block types are ordered by containment of the tensors they admit. We write
$\tau \sqsubseteq \tau'$ when $\sem{\tau} \subseteq \sem{\tau'}$, so that

\[
\begin{gathered}
  \stblock_\theta\sqsubseteq\wblock_\theta\sqsubseteq\dblock,\\
  \cblock\sqsubseteq\rblock_j\sqsubseteq\dblock,\\
  \diagblock\sqsubseteq\dblock.
\end{gathered}
\]

\begin{figure}[htbp]
    \centering
    \fitwidth{
\begin{tikzpicture}[>=stealth, font=\small,line width=.3pt,x=.9cm,y=.85cm,
    alt={A Hasse diagram of block types. Zero is at the bottom. Above it
      sit StencilBlock, DiagonalBlock and UniformBlock; StencilBlock is
      below WindowBlock and UniformBlock is below BroadcastBlock. Window,
      Diagonal and Broadcast all sit below DenseBlock at the top.},
    t/.style={draw, rounded corners=2pt, inner sep=3pt, font=\footnotesize},
    c/.style={font=\scriptsize, text=black, inner sep=1pt}]

  \node[t] (dense) at (0,3.2) {\dblock};
  \node[c, above=1pt of dense] {$|\mathcal{R}|\,|\mathcal{C}|$};

  \node[t] (win)  at (-3.6,1.9) {$\wblock_\theta$};
  \node[t] (diag) at ( 0,  1.25) {\diagblock};
  \node[t] (bc)   at ( 3.6,1.9) {\rblock};
  \node[c, left=2pt of win]   {$Kc(ko-\chi)$};
  \node[c, left=2pt of diag] {$n$};
  \node[c, right=2pt of bc]   {one slice};

  \node[t] (sten) at (-3.6,0.6) {$\stblock_\theta$};
  \node[t] (unif) at ( 3.6,0.6) {\cblock};
  \node[c, left=2pt of sten]  {$Kck$};
  \node[c, right=2pt of unif] {$1$};

  \node[t] (zero) at (0,-0.6) {$\mathsf{Zero}$};

  \draw (win) -- (dense); \draw (diag) -- (dense); \draw (bc) -- (dense);
  \draw (sten) -- (win);  \draw (unif) -- (bc);
  \draw (zero) -- (sten); \draw (zero) -- (diag); \draw (zero) -- (unif);
\end{tikzpicture}}
    \caption{Order on the named block types. Higher types admit more tensors;
        the diagram shows relations valid for all parameters, not the full
        lattice. Labels count stored value classes: $K,c$ are channel
        counts, $o$ the output extent, $k$ the kernel extent, and $\chi$ the
        number of clipped window positions; $\mathcal{R}$ and $\mathcal{C}$ are
        the row and column spaces. The stencil count assumes every
        kernel offset is read; $n$ is the diagonal extent.}
    \label{fig:gb-lattice}
\end{figure}
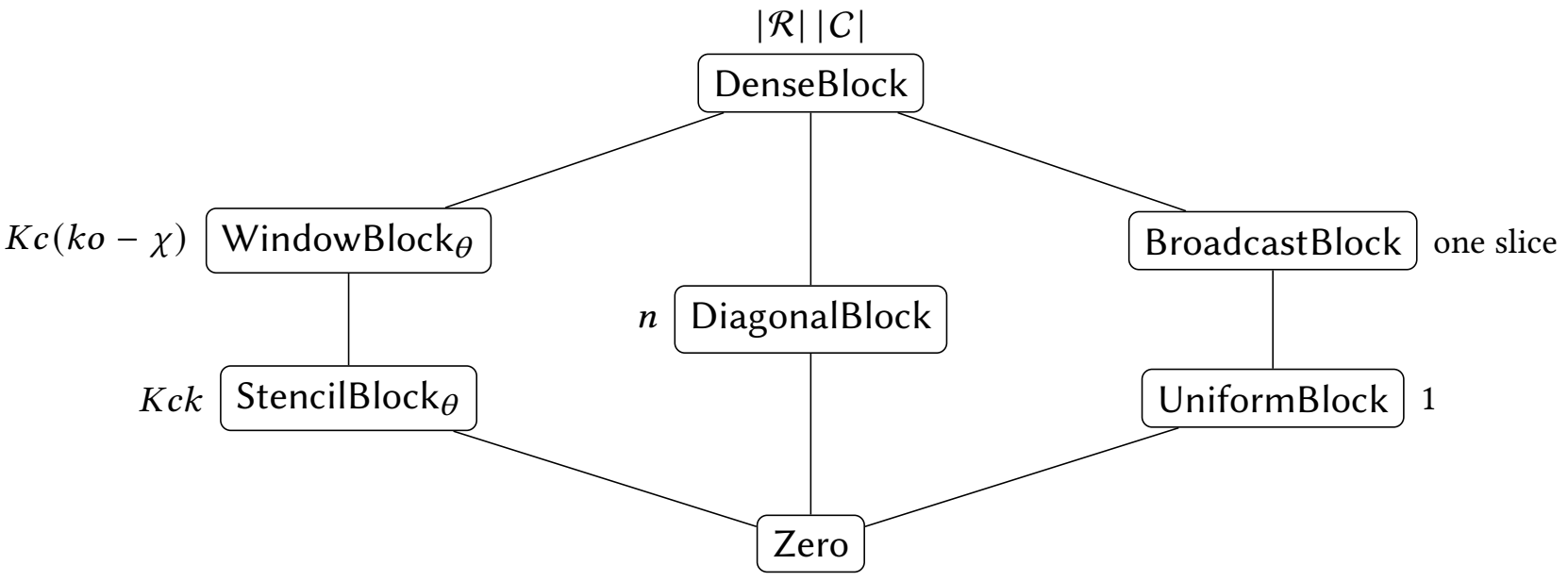

By Lemma~\ref{lem:gb-subspace}, $\tau \sqsubseteq \tau'$ implies that
$\tau$ has no more $\approx$-classes than $\tau'$;
the stencil, window, and dense types
have at most $Kck$, exactly $Kc(ko - \chi)$, and
$\lvert \mathcal{R} \rvert \lvert \mathcal{C} \rvert$ classes. Some edges
of the order depend on the parameters. Over three positions,
$\stblock_{(1,1,0)} \sqsubseteq \stblock_{(3,1,1)}$, since the kernel $[a]$
is the kernel $[0, a, 0]$ with one position of padding, and
$\diagblock \sqsubseteq \wblock_\theta$ whenever the diagonal lies in
$\mathcal{S}_\theta$. Fig.~\ref{fig:gb-lattice} shows the edges that hold
for all parameters. A block of type $\tau$ is also a block of every
$\tau' \sqsupseteq \tau$, so widening is always available. It preserves the
tensor that the block denotes, and since the wider type has at least as
many $\approx$-classes, it can lose compression and increase the arithmetic
of the kernels. The storage that the implementation allocates need not
grow, since the core of a type can have entries that no position maps to
(as described at the start of this appendix). Widening gives a fallback: where
no rule below applies, we widen to $\dblock$.

The full family of support-and-sharing types on a finite box forms a
lattice (Lemma~\ref{lem:gb-type-lattice}), although the six named types
need not contain its joins or meets. Specialization requires a sound,
deterministic result type; it does not require a least result type or
lattice meets.

Throughout this subsection the two operands' types are read on a common
ambient index space, with every position outside a support treated as a
known zero. The positions outside a support are related to \emph{one
another} as well as to themselves, since all of them hold the same value.
The convention preserves the equalities between known zeros, which the
intersection of two sharing relations would otherwise forget. Relating each
off-support position only to itself would give
$\mathtt{Zero} \sqcup \cblock = \dblock$, since the identity relation of
the first operand meets the all-relation of the second in the identity,
whereas every matrix of either type is constant and $\cblock$ admits them
all. Relating the off-support positions to one another gives $\cblock$.

With this convention, the union-of-supports and intersection-of-relations
construction is
\begin{equation}\label{eq:gb-type-join}
  \tau_1\sqcup\tau_2=
  \bigl(\mathcal S_1\cup\mathcal S_2,
  (\widehat{\approx}_1\cap\widehat{\approx}_2)
       |_{\mathcal S_1\cup\mathcal S_2}\bigr),
\end{equation}
where $\widehat{\approx}_j$ extends $\approx_j$ by relating all positions
outside $\mathcal S_j$ to one another. Element-wise multiplication uses
the intersection of supports with the same intersection of extended
sharing relations, restricted to that support. This operation is not the
lattice meet: an equality guaranteed by only one operand need not survive
multiplication.

\begin{lemma}[Type lattice]\label{lem:gb-type-lattice}
On a fixed finite box, the support-and-sharing types ordered by semantic
inclusion form a lattice. Their join is the union-of-supports and
intersection-of-extended-relations construction of
Equation~\eqref{eq:gb-type-join}. Their meet denotes the intersection of their
sets of tensors.
\end{lemma}

\begin{proof}
Represent a type by equations $x_\iota=0$ outside its support and
$x_\iota=x_{\iota'}$ for each sharing class. Conjoining the equations for
two types represents their intersection: take the equivalence closure of
all equality constraints, mark a class zero if it contains a position
forced to zero by either type, remove these classes from the support, and
retain the remaining classes as the sharing relation. This gives a type
and is the greatest lower bound by its denotation.

For the join, every position possible in either operand must be in the
support of an upper bound: it can be assigned a nonzero value in that
operand. A pair of positions must be equal in every tensor of one type
exactly when they lie in the same sharing class or are both outside its
support. Thus any equality imposed by an upper bound must belong to both
extended relations. The construction keeps exactly these equalities over
the union of the supports. It admits both operand types, and every upper
bound admits it, so it is their least upper bound. The empty support and
full support with independent entries give bottom and top. Since the box
is finite, so is the family of types; these operations also yield joins
and meets of arbitrary families.
\end{proof}

\subsection{Structural Rules for Tensor Operations}\label{app:gb-operations}
The rules of \S\ref{sec:gb-algorithms} first operate on two blocks over
a common region, then combine their results into a tensor. We follow
the same order here. The regional rules use the block types to choose
the output representation and numerical calls. For complete tensors,
refinement and alignment identify the regional operands. These choices
use descriptors; the recorded calls compute the numerical cores.

\subsubsection{Operations on Blocks}\label{app:gb-block-rules}
The three rules operate on aligned operands over one local index domain.
An operand is a symbolic pair $(\tau,u)$ or a background marker
$\underline d$. The background $d'$ of the eventual result tensor is
fixed when applying the rules.

\textsc{B-Identity} handles a marker that is an identity on its operand
side. It returns the other block's type and core name, with no new
numerical call. \textsc{B-Annihilator} handles an annihilator equal to
$d'$ and returns no block. Its result is represented by the tensor
background. Both decisions inspect the marker's descriptor value;
neither tests a stored core for equality with an identity or annihilator.

\textsc{B-Compute} covers the remaining cases. Constant lifting first
gives two block types on the same domain. The support and sharing
construction of \S\ref{sec:gb-result-types} yields a sound type in the
full type lattice; the fixed library rule selects an admissible named
type above it. The result need not be a least named type, and the
element-wise multiplication rule is not the lattice meet. The rule then
records the lifting calls followed by the numerical kernel, including
any conversions needed for that selected type. Its result core name is
fresh. Lemma~\ref{lem:gb-regional-rules} proves these three cases;
\textsc{T-Elementwise} composes their calls and places their outputs in
the tensor. We define constant lifting below, then give the result-type
calculation in \S\ref{sec:gb-result-types}.

Both tests depend on the operator, and the identity test also depends
on the operand side. We use the following elements of a binary operator
$\odot$.
\begin{itemize}[topsep=2pt,itemsep=0pt]
\item A \emph{right identity} $\mathbf{1}^{\mathrm R}_\odot$ is a value such
  that $x \odot \mathbf{1}^{\mathrm R}_\odot = x$ for all $x$, and a
  \emph{left identity} $\mathbf{1}^{\mathrm L}_\odot$ is a value such that
  $\mathbf{1}^{\mathrm L}_\odot \odot x = x$ for all $x$.
\item The \emph{annihilator element} $\mathbf{0}_\odot$ is the value such
  that $x \odot \mathbf{0}_\odot = \mathbf{0}_\odot \odot x =
  \mathbf{0}_\odot$ for all $x$ at which these are defined.
\end{itemize}
Table~\ref{tab:gb-units} lists them for the operators the library provides.
Subtraction and division have a right identity and no left identity:
$0 - x$ and $1 \div x$ differ from $x$ in general. We say that a background is an
identity of $\odot$ \emph{on its side} if it is the background $d_2$ of
$\mathcal{T}_2$ and equals $\mathbf{1}^{\mathrm R}_\odot$, or the
background $d_1$ of $\mathcal{T}_1$ and equals
$\mathbf{1}^{\mathrm L}_\odot$.

\begin{table}[htbp]
\centering\small
\caption[Identity and annihilator elements]{Identity and annihilator
    elements of the element-wise operators. A dash marks an operator with
    no such element.}
\label{tab:gb-units}
\begin{tabular}{@{}lcccccc@{}}
\toprule
& $+$ & $-$ & $\times$ & $\div$ & $\max$ & $\min$ \\
\midrule
$\mathbf{1}^{\mathrm R}_\odot$ & $0$ & $0$ & $1$ & $1$ & $-\infty$ & $+\infty$ \\
$\mathbf{1}^{\mathrm L}_\odot$ & $0$ & --- & $1$ & --- & $-\infty$ & $+\infty$ \\
$\mathbf{0}_\odot$ & --- & --- & $0$ & --- & $+\infty$ & $-\infty$ \\
\bottomrule
\end{tabular}
\end{table}
The predicate $\mathsf{id}_j(\odot,d)$ holds when $d$ is an identity
of $\odot$ on side $j$, with side $1$ on the left and side $2$ on the
right. The predicate $\mathsf{ann}(\odot,d)$ holds when $d$ is a
two-sided annihilator of $\odot$. The condition
$d=d'$ in \textsc{B-Annihilator} ensures that omitting the output block
represents the regional result correctly.

\paragraph{Constant lifting.}
For a region $r$ with origin $s_r$, let
$\Omega_r=\{\iota-s_r:\iota\in r\}$ be its local index domain. Set
$\tau_{\mathrm c,r}=(\Omega_r,\Omega_r,\Omega_r\times\Omega_r)$ and define
\[
\begin{aligned}
  L_r((\tau,u))&=(\tau,u,\epsilon),\\
  L_r(\underline d)&=(\tau_{\mathrm c,r},c,[c\gets(d)]),
                      \qquad c\text{ fresh}.
\end{aligned}
\]
The recorded assignment constructs the one-entry core $(d)$.
For regional operands, $\operatorname{Lift}_r$ is defined exactly when
neither \textsc{B-Identity} nor \textsc{B-Annihilator} applies, and then
\[
\begin{gathered}
  (\tau_\ell,u_\ell,\lambda_\ell)=L_r(\beta_\ell)
       \quad(\ell=1,2),\\
  \operatorname{Lift}_r(\vec\beta)
       =((\tau_1,\tau_2),(u_1,u_2),\lambda_1;\lambda_2).
\end{gathered}
\]
Lifting inspects the type or background marker and records numerical
construction; it does not read any numerical core.

\subsubsection{Result Types}\label{sec:gb-result-types}\label{app:gb-lattice}
Take two block types over the same index domain. Every position outside
an operand's support holds $0$ by Definition~\ref{def:gb-block},
independently of its tensor's background, so the structural result type
depends on how $\odot$ treats $0$. These are the type computations used
by \textsc{B-Compute} after any constant lifting.

\begin{definition}[Disjunctive and conjunctive operators]
\label{def:gb-disj}
A binary operator $\odot$ is \emph{conjunctive at $0$} if $0$ is its
annihilator, and \emph{disjunctive at $0$} if $0 \odot 0 = 0$ and $0$ is
not its annihilator.
\end{definition}

Addition, subtraction, maximum, minimum and disjunction are disjunctive at
$0$, as are the
strict comparisons and $\neq$, since $0 < 0$ and $0 \neq 0$ are false and
false is the zero of the Boolean semiring. Multiplication and conjunction
are conjunctive at $0$. Four operators are neither: $\leq$, $\geq$ and $=$
return true at $(0,0)$, and $0 \div 0$ is undefined, so for these we
compute the region and assign the result the type $\dblock$.

In the conjunctive case a position at which either operand lies outside its
support contributes $0$, so the support of the result is
contained in the intersection of the two supports. In the disjunctive case
only a position at which both operands lie outside their supports is
guaranteed to contribute $0$, so the support of the result is contained in
their union:
\begin{equation}
\label{eq:gb-supp}
  \supp(A \odot B) \subseteq
  \begin{cases}
    \supp(A) \cup \supp(B) & \text{disjunctive},\\
    \supp(A) \cap \supp(B) & \text{conjunctive}.
  \end{cases}
\end{equation}
A diagonal tensor added to a dense tensor is therefore dense, while a
diagonal tensor multiplied element-wise by a dense tensor is diagonal.

The sharing relations are intersected in either case, since two positions
of the result are guaranteed to be equal whenever both operands constrain
them to be equal. Here each sharing relation is read over the whole box,
with the positions outside its support related to one another, since all of
them hold $0$. The intersection of two relations read this way is an
equivalence relation on the box, and we restrict it to the support of the
result. Define
\[
\begin{aligned}
  \tau_1 \sqcup \tau_2
    &=\bigl(\mathcal{S}_1\cup\mathcal{S}_2,
       (\widehat{\approx}_1\cap\widehat{\approx}_2)
          |_{\mathcal{S}_1\cup\mathcal{S}_2}\bigr),\\
  \tau_1 \curlywedge \tau_2
    &=\bigl(\mathcal{S}_1\cap\mathcal{S}_2,
       (\widehat{\approx}_1\cap\widehat{\approx}_2)
          |_{\mathcal{S}_1\cap\mathcal{S}_2}\bigr),
\end{aligned}
\]
where the hats denote the extensions of the equation in Fig.~\ref{fig:gb-result-types}. These constructions give
sound structural result types for disjunctive and conjunctive operators,
respectively. The operation $\sqcup$ is the join of the full type order.
The operation $\curlywedge$ is not its meet: a meet combines constraints
from both operands, whereas multiplication preserves only equalities
guaranteed by both. The constructions apply to all six named families,
including the full-support types $\cblock$ and $\rblock$.

\paragraph{Selecting a named result type.}
A structural result need not belong to a named family. The fixed library
rule $\odot^\tau$ therefore selects an admissible named type above it,
using $\dblock$ when necessary. Selection can enlarge support or forget
sharing, but every defined dense result remains admitted. It depends on
types and operator parameters, not the core values.
Table~\ref{tab:gb-lattice} records the choices for the four displayed
families, with matching window parameters. It does not specify every
pair among the six families; other cases use the same support/sharing
construction and the fixed library selection policy, with the dense
fallback. We derive the displayed entries below after establishing
support and sharing propagation.

\begin{table}[htbp]
    \centering\small
    \caption[Result types for element-wise operations]{Result types for
        element-wise operations over a region that both operands cover.
        Top: operators disjunctive at $0$ ($+$, $-$, $\max$, $\min$, the strict
        comparisons, $\neq$, $\vee$). Bottom: operators conjunctive at $0$
        ($\times$, $\wedge$). The two differ only in whether supports are
        united or intersected. An asterisk marks an entry that can be
        strictly above the result of $\sqcup$ or $\curlywedge$.
        The accompanying text derives each entry.}
    \label{tab:gb-lattice}
    \begin{tabular}{@{}l|cccc@{}}
    \toprule
    $\sqcup$ & $\stblock_\theta$ & $\wblock_\theta$ & $\diagblock$ & $\dblock$ \\
    \midrule
    $\stblock_\theta$ & $\stblock_\theta$ & $\wblock_\theta$ & $\dblock{}^{*}$ & $\dblock$ \\
    $\wblock_\theta$ & $\wblock_\theta$ & $\wblock_\theta$ & $\dblock{}^{*}$ & $\dblock$ \\
    $\diagblock$     & $\dblock{}^{*}$ & $\dblock{}^{*}$ & $\diagblock$ & $\dblock$ \\
    $\dblock$        & $\dblock$ & $\dblock$ & $\dblock$ & $\dblock$ \\
    \bottomrule
    \end{tabular}
    \hfill
    \begin{tabular}{@{}l|cccc@{}}
    \toprule
    $\curlywedge$ & $\stblock_\theta$ & $\wblock_\theta$ & $\diagblock$ & $\dblock$ \\
    \midrule
    $\stblock_\theta$ & $\stblock_\theta$ & $\wblock_\theta$ & $\diagblock{}^{*}$ & $\wblock_\theta$ \\
    $\wblock_\theta$ & $\wblock_\theta$ & $\wblock_\theta$ & $\diagblock{}^{*}$ & $\wblock_\theta$ \\
    $\diagblock$     & $\diagblock{}^{*}$ & $\diagblock{}^{*}$ & $\diagblock$ & $\diagblock$ \\
    $\dblock$        & $\wblock_\theta$ & $\wblock_\theta$ & $\diagblock$ & $\dblock$ \\
    \bottomrule
    \end{tabular}
\end{table}

\begin{lemma}[Support propagation]
\label{lem:gb-supp}
Let $A \odot B$ be defined. If $\odot$ is disjunctive at $0$
(Definition~\ref{def:gb-disj}) then
$\supp(A \odot B) \subseteq \supp(A) \cup \supp(B)$; if it is conjunctive
at $0$ then $\supp(A \odot B) \subseteq \supp(A) \cap \supp(B)$.
\end{lemma}

\begin{proof}
Entrywise. If $\iota$ lies outside both supports then
$A[\iota] = B[\iota] = 0$ and $0 \odot 0 = 0$, which the disjunctive case
assumes and the conjunctive case gives by annihilation. If $\iota$ lies
outside $\supp(A)$ alone then $A[\iota] = 0$ and the conjunctive case
gives $0$ by annihilation, and symmetrically for $B$. \end{proof}

Both inclusions are conservative and neither is attained in general. The
operator $x \odot y = y$ is disjunctive at $0$, yet
$\supp(A \odot B) = \supp(B)$, which is a proper subset of the union
whenever $A$ has support outside $B$'s. We use the inclusions as transfer rules and
claim no tightness for them; a rule that over-approximates the support
assigns a wider type, which the fallback property of
\S\ref{sec:gb-types} makes sound.

\begin{lemma}[Sharing propagation]
\label{lem:gb-share}
Let $A\in\sem{\tau_1}$ and $B\in\sem{\tau_2}$, and suppose $A\odot B$
is defined. If $\iota\widehat{\approx}_1\iota'$ and
$\iota\widehat{\approx}_2\iota'$, then
$(A\odot B)[\iota]=(A\odot B)[\iota']$.
\end{lemma}

\begin{proof}
For each operand, the two positions either share a core entry or both
lie outside its support and hold zero. Thus both operands take equal
values at the two positions, and applying the same scalar operation
gives equal result values.
\end{proof}

The converse fails, so $\curlywedge$ is a sound rule and not a complete
one. Taking $\odot$ to be multiplication and $\tau_1$ the type whose every
matrix is zero, the result is zero at both positions however
$\approx_2$ relates them, so the shared-equality hypothesis is not
necessary. A rule that forgets an equality assigns a wider type, which is
again sound by the fallback property.

These lemmas justify the support and sharing calculations in the equation in Fig.~\ref{fig:gb-result-types}. The operation
$\sqcup$ is the join of the type order: it is the least type admitting every
matrix of either operand type. The operation $\curlywedge$ retains only equalities guaranteed by both
operands, as required by Lemma~\ref{lem:gb-share}. In contrast, the
lattice meet conjoins the two sets of constraints and can impose
additional equalities that multiplication does not preserve.

The entries of Table~\ref{tab:gb-lattice} follow by computing $\sqcup$ and
$\curlywedge$ on the named types. Write $\Delta$ for the diagonal.
\begin{itemize}[topsep=2pt,itemsep=1pt]
\item Two window types with the same parameters share the support
  $\mathcal{S}_\theta$. Intersecting $\approx_\theta$ with itself gives
  $\approx_\theta$, and intersecting it with the identity relation gives
  the identity, so the results are $\stblock_\theta$ for two
  $\stblock_\theta$ operands and $\wblock_\theta$ otherwise, under either
  operation.
\item Under $\sqcup$, a type combined with $\dblock$ has full support and the
  identity relation, which is $\dblock$. Under $\curlywedge$ it keeps its
  own support and has the identity relation, which is $\wblock_\theta$ for
  both window types and $\diagblock$ for $\diagblock$.
\item Two $\diagblock$ operands give $\diagblock$ under either operation.
\item A window type combined with $\diagblock$ under $\curlywedge$ has
  support $\mathcal{S}_\theta \cap \Delta$ and the identity relation,
  which $\diagblock$ admits. It equals $\diagblock$ when
  $\Delta \subseteq \mathcal{S}_\theta$, and lies strictly below it
  otherwise.
\item A window type combined with $\diagblock$ under $\sqcup$ has support
  $\mathcal{S}_\theta \cup \Delta$, and the table assigns $\dblock$. When
  $\Delta \subseteq \mathcal{S}_\theta$, $\wblock_\theta$ also admits the
  result.
\end{itemize}
The last two items are the entries Table~\ref{tab:gb-lattice} marks with an
asterisk.

\subsubsection{Refinement and Alignment}\label{app:gb-refine}

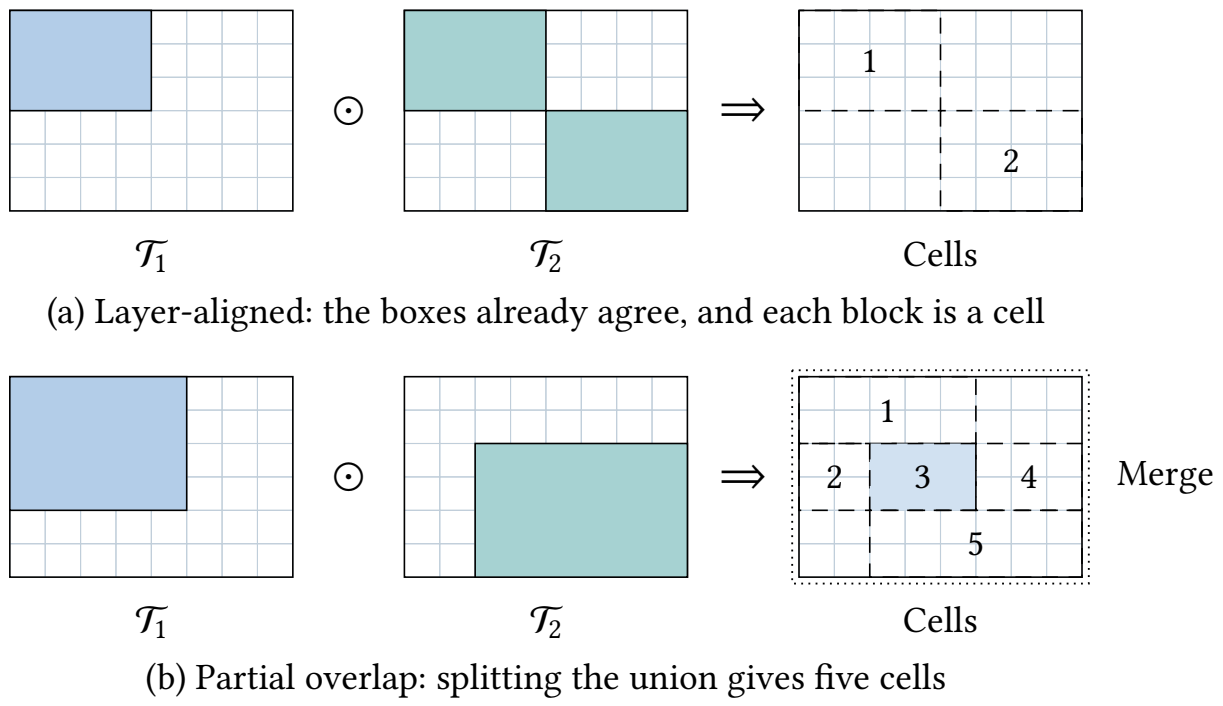
\begin{figure}[htbp]
    \centering
    \fitwidth{
\begin{tikzpicture}[>=stealth, font=\small,x=.9cm,y=.85cm,
    alt={Two panels, each showing three two-dimensional tensors drawn as
      grids of cells. In the first panel the block of the left operand and
      a block of the right operand occupy the same rectangle, a second
      block of the right operand occupies a disjoint rectangle, and the
      refinement on the right has one cell per block. In the second panel
      the two blocks overlap partially, and the refinement on the right
      cuts their union into five rectangular cells; a dashed outline marks
      the bounding box that merging would use instead.},
    cellg/.style={draw=structural!30, line width=0.2pt},
    blk/.style={line width=0.3pt},
    cel/.style={draw=black, densely dashed, line width=0.3pt},
    lab/.style={font=\scriptsize, anchor=north},
    cnum/.style={font=\scriptsize, text=black}]

  \def\cs{0.26}
  \def\nc{8}
  \def\nr{6}
  \colorlet{opA}{diagramcolor!30}
  \colorlet{opB}{teal!35}

  \newcommand{\gbgrid}[2]{%
    \foreach \r in {0,...,\the\numexpr\nr-1\relax}{%
      \foreach \c in {0,...,\the\numexpr\nc-1\relax}{%
        \draw[cellg] ({#1+\c*\cs},{#2-\r*\cs}) rectangle ++(\cs,-\cs);}}%
    \draw[black, line width=0.3pt]
      ({#1},{#2}) rectangle ({#1+\nc*\cs},{#2-\nr*\cs});}
  \newcommand{\gbblk}[7]{%
    \draw[blk,#7] ({#1+#5*\cs},{#2-#3*\cs})
      rectangle ({#1+(#6+1)*\cs},{#2-(#4+1)*\cs});}

  \def\xa{0}
  \def\xb{2.90}
  \def\xc{5.80}

  \def\ya{0}
  \gbgrid{\xa}{\ya} \gbblk{\xa}{\ya}{0}{2}{0}{3}{fill=opA, draw=black}
  \gbgrid{\xb}{\ya}
  \gbblk{\xb}{\ya}{0}{2}{0}{3}{fill=opB, draw=black}
  \gbblk{\xb}{\ya}{3}{5}{4}{7}{fill=opB, draw=black}
  \gbgrid{\xc}{\ya}
  \gbblk{\xc}{\ya}{0}{2}{0}{3}{cel}
  \gbblk{\xc}{\ya}{3}{5}{4}{7}{cel}
  \node[cnum] at ({\xc+2*\cs},{\ya-1.5*\cs}) {1};
  \node[cnum] at ({\xc+6*\cs},{\ya-4.5*\cs}) {2};
  \node at ({(\xa+\nc*\cs+\xb)/2},{\ya-\nr*\cs/2}) {$\odot$};
  \node at ({(\xb+\nc*\cs+\xc)/2},{\ya-\nr*\cs/2}) {$\Rightarrow$};
  \node[lab] at ({\xa+\nc*\cs/2},{\ya-\nr*\cs-0.08}) {$\mathcal{T}_1$};
  \node[lab] at ({\xb+\nc*\cs/2},{\ya-\nr*\cs-0.08}) {$\mathcal{T}_2$};
  \node[lab] at ({\xc+\nc*\cs/2},{\ya-\nr*\cs-0.08}) {Cells};
  \node[font=\scriptsize, anchor=north] at ({\xb+\nc*\cs/2},{\ya-\nr*\cs-0.52})
    {(a) Layer-aligned: the boxes already agree, and each block is a cell};

  \def\yb{-2.85}
  \gbgrid{\xa}{\yb} \gbblk{\xa}{\yb}{0}{3}{0}{4}{fill=opA, draw=black}
  \gbgrid{\xb}{\yb} \gbblk{\xb}{\yb}{2}{5}{2}{7}{fill=opB, draw=black}
  \gbgrid{\xc}{\yb}
  \gbblk{\xc}{\yb}{0}{1}{0}{4}{cel}
  \gbblk{\xc}{\yb}{2}{3}{0}{1}{cel}
  \gbblk{\xc}{\yb}{2}{3}{2}{4}{cel, fill=opA!60}
  \gbblk{\xc}{\yb}{2}{3}{5}{7}{cel}
  \gbblk{\xc}{\yb}{4}{5}{2}{7}{cel}
  \node[cnum] at ({\xc+2.5*\cs},{\yb-1.0*\cs}) {1};
  \node[cnum] at ({\xc+1.0*\cs},{\yb-3.0*\cs}) {2};
  \node[cnum] at ({\xc+3.5*\cs},{\yb-3.0*\cs}) {3};
  \node[cnum] at ({\xc+6.5*\cs},{\yb-3.0*\cs}) {4};
  \node[cnum] at ({\xc+5.0*\cs},{\yb-5.0*\cs}) {5};
  \node at ({(\xa+\nc*\cs+\xb)/2},{\yb-\nr*\cs/2}) {$\odot$};
  \node at ({(\xb+\nc*\cs+\xc)/2},{\yb-\nr*\cs/2}) {$\Rightarrow$};
  \node[lab] at ({\xa+\nc*\cs/2},{\yb-\nr*\cs-0.08}) {$\mathcal{T}_1$};
  \node[lab] at ({\xb+\nc*\cs/2},{\yb-\nr*\cs-0.08}) {$\mathcal{T}_2$};
  \node[lab] at ({\xc+\nc*\cs/2},{\yb-\nr*\cs-0.08}) {Cells};
  \draw[black, densely dotted, line width=0.35pt]
    ({\xc-0.05},{\yb+0.05}) rectangle ({\xc+\nc*\cs+0.05},{\yb-\nr*\cs-0.05});
  \node[font=\scriptsize, anchor=west, text=black]
    at ({\xc+\nc*\cs+0.12},{\yb-\nr*\cs/2}) {Merge};
  \node[font=\scriptsize, anchor=north] at ({\xb+\nc*\cs/2},{\yb-\nr*\cs-0.52})
    {(b) Partial overlap: splitting the union gives five cells};
\end{tikzpicture}}
    \caption[A common refinement]{A common refinement of two block lists.
        (a) Aligned operands: the boxes coincide or are disjoint, and each
        block is a cell. (b) Partially overlapping operands: splitting the
        union into rectangles gives five cells; the dotted outline is the
        bounding box that merging would use instead.}
    \label{fig:gb-refinement}
\end{figure}

The block rules require both operands to be described over the same
region. The tensor rule obtains these regions by refining the operand
block boxes. For layer-aligned operands, the boxes coincide or are
disjoint, as in Fig.~\ref{fig:gb-refinement}(a). Partially overlapping
boxes require splitting or materialization, as in
Fig.~\ref{fig:gb-refinement}(b).

Write $\boxof(B)$ for the box a block occupies. Two blocks \emph{overlap}
if their boxes intersect.

\begin{definition}[Common refinement]
\label{def:gb-refine}
A \emph{common refinement} of two block lists is a finite set of pairwise
disjoint boxes $r_t$ with labels $L_t = (L^1_t, L^2_t)$ such that the $r_t$
cover the union of all the boxes, and for each $t$ and each operand $l$, the
set $L^l_t$ collects the blocks of operand $l$ whose box contains $r_t$,
with no block of operand $l$ partially overlapping $r_t$.
\end{definition}

The construction first groups the blocks of both operands by overlap.
Grouping is transitive: if a block $B_1$ of the first operand overlaps a
block $B_2$ of the second, and $B_2$ overlaps another block $B_3$ of the
first, then $B_1$ and $B_3$ belong to the same group even though they do
not overlap, since both determine values in the region of $B_2$. The
groups are therefore the connected components of the graph whose vertices
are the blocks of both operands and whose edges join overlapping blocks.
Our implementation finds them by depth-first search; a union-find
structure computes the same components. Distinct components occupy
disjoint regions, so the construction resolves each independently.

A component is realized either by splitting it into smaller disjoint
cells or by merging it into one cell. Splitting keeps the blocks small at
the cost of extra bookkeeping: two overlapping boxes that differ along at
most one axis split into three pieces along that axis, and the number of
pieces grows once the boxes are offset along several axes, as in
Fig.~\ref{fig:gb-refinement}(b). Merging is always possible, at the cost
of the structure of the blocks. The construction splits a component when its boxes differ
along at most one axis, pairwise into at most three pieces along that
axis. Otherwise the component is \emph{merged}: each
operand's blocks within the component are first materialized into a single
block covering the component's bounding box, padded with that operand's
background, and the bounding box becomes one cell. Materialization is what
makes the merge policy satisfy the second condition of
Definition~\ref{def:gb-refine}, since afterwards each operand contributes
exactly one block to the cell. The output is therefore a common refinement
of the block lists after materialization, in which the blocks of each
merged component are replaced by the materialized ones, and uniformity, as
well as the case analysis of the proof of Theorem~\ref{thm:gb-correct},
refers to these blocks.

Disjointness needs one step more. A component's bounding box may be strictly
larger than the union of its blocks and so may overlap other components,
in which case emitting both cells would violate the first
condition. Concretely, with $A = [0,4) \times [0,1)$ and
$C = [2,3) \times [2,3)$ in the first operand and $B = [0,1) \times [0,4)$
in the second, the blocks $A$ and $B$ overlap and their bounding box
$[0,4) \times [0,4)$ contains $C$, which overlaps neither. The construction
therefore recomputes overlap components using the enlarged boxes and
repeats whenever a new overlap joins distinct components. Each such
nonterminal repetition merges at least two components, so the process
terminates. At termination, no cells from distinct components overlap.

\begin{lemma}[Uniform cells]
\label{lem:gb-cells}
In the common refinement that the construction produces, each cell is
contained in at most one block of each operand, after materialization, and
the cells together cover every index that either operand covers.
\end{lemma}

\begin{proof}
The cells cover the union of the boxes by the first condition of
Definition~\ref{def:gb-refine}, and they are pairwise disjoint by the
argument of the previous paragraph. A cell produced by splitting lies in at
most one block of each operand, since the blocks of one operand are
disjoint, and a cell produced by merging lies in exactly one materialized
block of each operand.
\end{proof}

\begin{lemma}[Support-determinacy]
\label{lem:gb-support-det}
The refinement that the construction above produces, including its choice
between splitting and merging, is a function of the block boxes and the
tensor shape alone. It does not depend on the core tensors, on the
backgrounds, or on the operator.
\end{lemma}

\begin{proof}
Every choice of the construction is made on boxes. The overlap relation and
its connected components are defined by box intersection; the choice
between splitting and merging a component tests whether its boxes differ
along at most one axis; the pieces of a split and the bounding box of a
merge are computed from the boxes; and the recomputation of the components
after an enlargement uses box intersection and containment again. None of
these steps reads a core, a background or the operator. Materialization
reads the background to fill the materialized block, but it does not
change the box of that block, so it does not affect the cells.
\end{proof}

Theorem~\ref{thm:gb-separate} uses Lemma~\ref{lem:gb-support-det} for the
cells of an element-wise result.

\paragraph{Restriction and extension.}\label{app:gb-restrict}

The refinement produces cells that may be smaller or larger than the operand
blocks, so each operand must be restricted or extended to the cell before
the type rules apply. The types assigned are as follows. A restriction of
$\diagblock$ keeps its type when the row and column ranges of the sub-box
coincide, and a restriction of a window type keeps its type, with
parameters $\theta'$, when the sub-box is a product in the sense below.
The table assigns $\dblock$ otherwise.

\begin{center}\footnotesize
\setlength{\tabcolsep}{4pt}
\begin{tabular}{@{}lll@{}}
\toprule
Type & Restrict & Extend \\
\midrule
$\dblock$        & $\dblock$ & $\dblock$ \\
$\cblock$        & $\cblock$ & $\dblock$ \\
$\rblock$        & $\rblock$ & $\dblock$ \\
$\diagblock$     & $\diagblock$ or $\dblock$ & $\dblock$ \\
$\wblock_\theta$ & $\wblock_{\theta'}$ or $\dblock$ & $\dblock$ \\
$\stblock_\theta$ & $\stblock_{\theta'}$ or $\dblock$ & $\dblock$ \\
\bottomrule
\end{tabular}
\end{center}

For the two window types, call a sub-box a \emph{product} if, in the
coordinates of Definition~\ref{def:gb-named}, it is a range of output
channels and positions beginning at $(\kappa_0, a_0)$ together with a range
of input channels and positions beginning at $(\gamma_0, y_0)$. Position
$(a', y')$ of such a sub-box is position $(a' + a_0, y' + y_0)$ of the
block, and
\[
  (y' + y_0) - (a' + a_0)s + p \;=\; y' - a's + p',
  \qquad p' = p - a_0 s + y_0 ,
\]
so it lies in the window of its row exactly when it does under
$\theta' = (k, s, p')$, and it reads the same kernel offset. The
restriction of a $\stblock_\theta$ block is therefore the
$\stblock_{\theta'}$ block whose core is the kernel restricted to the
selected channels, and the restriction of a $\wblock_\theta$ block is the
$\wblock_{\theta'}$ block whose core holds the windows of the selected
rows. The padding $p'$ can be negative or exceed $k$, which the window
definition allows. A sub-box of the flattened neurons of a layer need not be
a product: a range of columns that ends inside a row of the image selects part
of that row.

\begin{lemma}[Widening]
\label{lem:gb-widen}
Let $\beta$ be a block of type $\tau$ on a box $r$ and let $r'$ be a
sub-box of $r$. Write $\tau|_{r'}$ for the restriction of $\tau$ to $r'$:
the type whose support is $\supp(\tau) \cap r'$ and whose sharing relation
is that of $\tau$ restricted to $r'$, both read in the index space of
$r'$. Then the restriction of $\beta$ to $r'$ has type $\tau|_{r'}$, and
the type assigned to it above lies above $\tau|_{r'}$ in the order on
types over $r'$. Extending $\beta$ to a box $r'' \supseteq r$ with a
background of $0$ gives a block of type $\tau$ read in the index space of
$r''$, and extending it with any background gives a block that $\dblock$,
the type assigned to it, admits.
\end{lemma}

Both comparisons in Lemma~\ref{lem:gb-widen} are made within one index
space, since a type over $r$ and a type over $r'$ are not comparable. The
type $\tau$ read in the index space of $r''$ embeds the support and the
sharing relation of $\tau$, and leaves the added positions outside the
support. It is in general not the named type of $\tau$ over $r''$:
extending the $\cblock$ $[5]$ by one position with background $0$ gives
$[5, 0]$, which is not constant. The table therefore assigns $\dblock$ to
every extension.

\begin{proof}[Proof of Lemma~\ref{lem:gb-widen}]
Restriction removes positions and changes no value. The support can only
shrink, and two surviving positions that were equal are still equal, so the
restricted block has type $\tau|_{r'}$. Each type the table assigns to a
restriction lies above $\tau|_{r'}$: $\dblock$ lies above every type, the
restrictions of $\cblock$ and $\rblock$ remain constant and repeated along
the same axis, a sub-box whose row and column ranges coincide contains the
diagonal of its own index space and no other position of the support, and
the paragraph above shows that the restriction of a window type to a
product has the window type with parameters $\theta'$. Extension with a
background of $0$ adds positions outside the support and related to
nothing, so the extended block has type $\tau$ read in the index space of
$r''$. Every extension is assigned $\dblock$, which admits it whatever the
background.
\end{proof}

\paragraph{The structural helpers.}\label{app:gb-helpers}
Assign each input block a symbolic core name. Write $Q_{\ell k}$ for
its tensor-coordinate box and $\Omega_Z$ for the tensor index domain.
The local domain $\Omega_r$ is defined in \S\ref{app:gb-block-rules}.
The fixed split/merge construction of \S\ref{app:gb-refine}
defines $\operatorname{Refine}(\vec\pi)=(r_1,\ldots,r_m)$, with
\[
  \bigcup_{\ell,k}Q_{\ell k}\subseteq\bigcup_t r_t
       \subseteq\Omega_Z,\qquad
  r_t\cap r_s=\varnothing\quad(t\ne s).
\]
Each nonempty rectangular region carries the construction's alignment
plan for each operand: absent, restrict a specified block, or materialize
over the region. We also use $r$ for the region's box. Merging may make
the first inclusion strict; the plan is retained even when a merged
region fits within one original block.

Alignment follows these plans and returns
\[
  \operatorname{Align}(r,\vec\pi)
     =((\beta_{1r},\beta_{2r}),\alpha_{1r};\alpha_{2r}).
\]
For an absent operand, $\beta_{\ell r}=\underline{d_\ell}$ and
$\alpha_{\ell r}=\epsilon$. For a restriction plan,
$\beta_{\ell r}=(\tau_{\ell r},u_{\ell r})$, with the restricted,
reindexed type selected as in Lemma~\ref{lem:gb-widen};
$\alpha_{\ell r}$ records the prescribed restriction and conversion
calls. A materialization plan instead selects $\dblock$ on $\Omega_r$
and records calls producing its core. Within an operand's block boxes,
materialization uses the blocks' full denotations, including structural
zeros; only positions outside those boxes take $d_\ell$. Output core
names are fresh unless the fixed plan reuses an existing block.

\paragraph{Example.}
On the index interval $[0,4)$, let the first operand have one block on
$[0,2)$ and background $0$, and the second one block on $[1,3)$ and
background $2$. Refinement gives $r_0=[0,1)$, $r_1=[1,2)$, and
$r_2=[2,3)$. Alignment records the corresponding block restrictions.
On $r_0$, it returns the first block's restriction and $\underline 2$;
for either addition or multiplication, lifting constructs a constant
core $(2)$ and \textsc{B-Compute} records the operation. On $r_1$,
both operands supply blocks, so \textsc{B-Compute} needs no lifting calls.
On $r_2$, addition uses \textsc{B-Identity}, while multiplication uses
\textsc{B-Annihilator}; its alignment calls remain recorded in both cases.
\textsc{T-Elementwise} places the retained results in their regions and
sets the background to $2$ for addition or $0$ for multiplication. The
regions and rule choices do not depend on any block's numerical entries.

\subsubsection{Element-wise Operations on Tensors}\label{sec:gb-elementwise}

We now relate the structural rules to the existing \sparse operation.
Algorithm~\ref{alg:gb-binary} records its operational steps under fixed
structural policies. Unlike the rules, the procedure executes each
numerical call when it encounters it. It retains redundant result
blocks so that the correspondence can identify both descriptors and
payloads.

Let $\mathcal{T}_1 = (d_1, S_1, B_1, Z)$ and
$\mathcal{T}_2 = (d_2, S_2, B_2, Z)$ be two \sparse tensors and let
$\mathcal{T}_3 = \mathcal{T}_1 \odot \mathcal{T}_2$ for a binary operator
$\odot$. To compute $\mathcal{T}_3$ we must determine which blocks of the
operands participate in the computation, where the blocks of
$\mathcal{T}_3$ lie, the type of each of them, and the background $d_3$.

The following procedure makes the refinement and identity, annihilator,
and constant-lifting cases explicit. The block rules of
Fig.~\ref{fig:gb-elementwise-rules} and the tensor rule of
Fig.~\ref{fig:gb-tensor-operation} are defined independently on
descriptors and symbolic cores. Lemma~\ref{lem:gb-reference-rules}
compares their chosen representation with this procedure under the same
fixed policies, with redundant blocks retained.

\begin{algorithm}[tb]
\caption{Element-wise operations over \sparse tensors}
\label{alg:gb-binary}
\begin{flushleft}
\textbf{Input:} Two \sparse tensors $\mathcal{T}_1 = (d_1, S_1, B_1, Z_1)$ and
$\mathcal{T}_2 = (d_2, S_2, B_2, Z_2)$ of the same shape; an element-wise
operator $\odot$ such that every scalar operation of
$\sem{\mathcal{T}_1} \odot \sem{\mathcal{T}_2}$ is defined. \\
\textbf{Output:} $\mathcal{T}_3$ with
$\sem{\mathcal{T}_3} = \sem{\mathcal{T}_1} \odot \sem{\mathcal{T}_2}$.
\end{flushleft}
\begin{algorithmic}[1]
\State $\mathcal{G} \gets$ \Call{Refine}{$S_1, B_1, S_2, B_2$}
  \CommentRight{The cells of a common refinement}
\State $d_3 \gets d_1 \odot d_2$, or $0$ if it is undefined
  \CommentRight{Background of the result}
\State $S_3, B_3 \gets \emptyset, \emptyset$; \ $Z_3 \gets Z_1$
\For{each cell $r \in \mathcal{G}$}
  \State $\beta_1 \gets$ \Call{Align}{$r, \mathcal{T}_1$}; \
         $\beta_2 \gets$ \Call{Align}{$r, \mathcal{T}_2$}
    \CommentRight{The operand over $r$; $\bot$ if no block of it meets $r$}
  \If{$\beta_1 = \bot$ or $\beta_2 = \bot$}
    \State $j \gets$ the index of the absent operand; \
           $\beta \gets \beta_{3-j}$; \ $d \gets d_j$
    \If{$\mathsf{id}_j(\odot,d)$}
      \State Append $(r, \beta)$ to $(S_3, B_3)$; \textbf{Continue}
        \CommentRight{Copy the block}
    \ElsIf{$\mathsf{ann}(\odot,d)$}
      \State \textbf{Continue}
        \CommentRight{The cell equals $d_3$ throughout; emit no block}
    \Else
      \State Lift $d$ to a $\cblock$ over $r$ and let it play the absent operand
    \EndIf
  \EndIf
  \State $\tau \gets \tau(\beta_1) \mathbin{\odot^\tau} \tau(\beta_2)$
    \CommentRight{\S\ref{sec:gb-result-types}}
  \State $b \gets$ \Call{Kernel}{$\odot, \tau(\beta_1), \tau(\beta_2), \beta_1, \beta_2$}
    \CommentRight{Core of the result}
  \State Append $(r, (\tau, b))$ to $(S_3, B_3)$
\EndFor
\State \Return $(d_3, S_3, B_3, Z_3)$
\end{algorithmic}
\end{algorithm}

Algorithm~\ref{alg:gb-binary} forms the regions on line~1, determines
$d_3$ on line~2, and applies the regional computation to each cell.
Alignment on line~5 follows the plans of \S\ref{app:gb-refine}. It
restricts an operand block to the cell or materializes its contribution
there, including the background outside its block boxes. In this
reference procedure, $\beta$ denotes the resulting numerical block and
$\tau(\beta)$ its type. The structural rules instead represent that
block by its type and a symbolic core name. Lines~16--17 select the
result type using $\odot^\tau$ and compute its core. In addition to the
alignment calls, each cell requires at most one result-computation
kernel call. The block-type calculation is the one developed in
\S\ref{sec:gb-result-types}.

\paragraph{The background value.}
The background of the result is $d_3 = d_1 \odot d_2$, since every entry
that is background in both operands takes this value. The combination
$d_1 \odot d_2$ can be undefined even when the dense operation is defined,
because the blocks can cover every position at which both backgrounds
would apply. For example, let $A = [2, 4]$ and $B = [1, 2]$ each be one
block with background $0$: the quotient $A \div B = [2, 2]$ is defined,
while $0 \div 0$ is not. Under the definedness assumption of
Theorem~\ref{thm:gb-correct}, no position then reads both backgrounds, and
line~2 sets $d_3 = 0$. The choice of $0$ also gives the correct value in
the cells that line~11 omits because the background of one operand is
the annihilator $0$ of multiplication (\S\ref{app:gb-proof-correct}).
The identities and annihilators in Table~\ref{tab:gb-units} determine
when an operand block can be reused or omitted. The remaining cases
lift a background to a constant block and use the result-type rule.
These are precisely the three cases of Fig.~\ref{fig:gb-elementwise-rules}.

\subsubsection{Tensor Multiplication}\label{sec:gb-matmul}

Tensor multiplication combines block pairs whose contracted ranges
overlap. Unlike an element-wise operation, an output position can
receive contributions from several such pairs. For
$\mathcal{T}_3 = \mathcal{T}_1 \tmul \mathcal{T}_2$, each entry is
\begin{equation}
\label{eq:gb-matmul}
  \mathcal{T}_3[\mathbf{i}, \mathbf{k}]
    \;=\; \sum_j \mathcal{T}_1[\mathbf{i}, j] \cdot
                 \mathcal{T}_2[j, \mathbf{k}] ,
\end{equation}
so a block of $\mathcal{T}_1$ interacts with a block of $\mathcal{T}_2$
whenever the column range of the first meets the row range of the second.
For a pair $(\beta_1,\beta_2)$ with nonempty contracted intersection
$J=\operatorname{cols}(\beta_1)\cap\operatorname{rows}(\beta_2)$, its output
box is $F=\operatorname{rows}(\beta_1)\times\operatorname{cols}(\beta_2)$.
These boxes can partially overlap. \Call{Meet}{} cuts each free axis at
every endpoint of the output boxes and retains the resulting cells
covered by at least one box. For each cell $r$, the list $\mathcal P[r]$
contains exactly the pairs whose output box contains all of $r$.
Contributing pairs are therefore constant throughout a cell; cells are
not merged across a boundary where this list changes.

For each pair in $\mathcal P[r]$, \Call{Trim}{} returns the restrictions
$\gamma_1=\beta_1|_{\operatorname{rows}(r)\times J}$ and
$\gamma_2=\beta_2|_{J\times\operatorname{cols}(r)}$.
Their product covers the whole cell $r$. We compute its core and sum the
partial products into the output block of $r$, using the join $\sqcup$
of their types for the sum. This construction depends only on block
positions and extents.

The background value requires more care than in the element-wise case. The
enumeration over pairs of blocks computes only those terms of
Equation~\ref{eq:gb-matmul} in which both factors lie in a block, so it
agrees with the dense product only when both backgrounds are $0$. For
instance, if $\mathcal{T}_1$ is the single block $[2]$ with background $0$
and $\mathcal{T}_2$ has no blocks and background $1$, the dense product is
$[2]$, whereas the enumeration finds no pair and returns $[0]$. We
therefore fall back to the dense product unless $d_1 = d_2 = 0$. The
fallback stores this product as one $\dblock$ with background $0$
and shape $Z_3$. In practice, the tensors that backsubstitution multiplies have background
$0$, so this case does not arise. The identity shortcut of
Algorithm~\ref{alg:gb-binary} also does not apply: a background of $1$ is
the identity for element-wise multiplication, but the all-ones matrix is
not the identity for tensor multiplication,
\[
  \begin{pmatrix}1&2\\3&4\end{pmatrix}
  \begin{pmatrix}1&1\\1&1\end{pmatrix}
  \;=\;
  \begin{pmatrix}3&3\\7&7\end{pmatrix} .
\]

Algorithm~\ref{alg:gb-matmul} is the reference procedure for the
multiplication construction of \S\ref{sec:gb-algorithms};
Lemma~\ref{lem:gb-reference-rules} proves their correspondence.
Lines~1--3 test the backgrounds,
lines~5 and~9 compute the regions, line~10 computes the type,
and line~11 computes the core. During structural evaluation,
this numerical work is recorded for runtime, together with any
conversions and additions needed for accumulation. The dense fallback
on line~2 stores its product as one \dblock covering the output shape,
with background $0$. It therefore also returns a \sparse tensor.

The accumulation $\boxplus$ in Algorithm~\ref{alg:gb-matmul} adds
blocks over the same output region using the addition type rule and
its numerical routine. The accumulator $\bot$ denotes the absence of
a contribution, with $\bot\boxplus B=B$. Subsequent additions convert
the cores to the selected result type as needed before summing them.

\begin{algorithm}[t]
\small
\algrenewcommand\algorithmicindent{0.8em}
\caption{Multiplication of \sparse tensors}
\label{alg:gb-matmul}
\begin{flushleft}
\textbf{Input:} $\mathcal T_\ell=(d_\ell,S_\ell,B_\ell,Z_\ell)$, $\ell\in\{1,2\}$,
with compatible contracted extents. \\
\textbf{Output:} $\mathcal{T}_3$ with
$\sem{\mathcal{T}_3} = \sem{\mathcal{T}_1} \tmul \sem{\mathcal{T}_2}$.\\
Write $Z_3$ for the output shape formed from the non-contracted axes
of $Z_1$ and $Z_2$.
\end{flushleft}
\begin{algorithmic}[1]
\If{$d_1 \neq 0$ \textbf{or} $d_2 \neq 0$}
  \State \Return \Call{Dense}{$\mathcal{T}_1$} $\tmul$ \Call{Dense}{$\mathcal{T}_2$}
\EndIf
\State $d_3 \gets 0$; \ $S_3, B_3 \gets \emptyset, \emptyset$
\State $\mathcal{P} \gets$ \Call{Meet}{$S_1, B_1, S_2, B_2$}
\For{$r\in\operatorname{dom}(\mathcal P)$}
  \State $A \gets \bot$
  \For{$(\beta_1,\beta_2)\in\mathcal P[r]$}
    \State $(\gamma_1, \gamma_2) \gets$ \Call{Trim}{$r, \beta_1, \beta_2$}
    \State $\tau \gets \tau(\gamma_1) \mathbin{\tmul^\tau} \tau(\gamma_2)$
    \State $b \gets$ \Call{Kernel}{$\tmul, \tau(\gamma_1), \tau(\gamma_2), \gamma_1, \gamma_2$}
    \State $A \gets A \mathbin{\boxplus} (\tau, b)$
  \EndFor
  \State Append $(r, A)$ to $(S_3, B_3)$
\EndFor
\State \Return $(d_3, S_3, B_3, Z_3)$
\end{algorithmic}
\end{algorithm}

It remains to give the type of the result. The parameters of the two
operands compose, so we treat the convolutional types first. Call a tensor
a \emph{window operator} with parameters $\theta = (k, s, p)$ if its
support is contained in $\mathcal{S}_\theta$, which holds for both
$\stblock_\theta$ and $\wblock_\theta$.

\begin{lemma}[Window composition]
\label{thm:gb-composition}
Let $M_{\mathrm i}$ be a window operator from an input space to a middle
space with parameters $(k_{\mathrm i}, s_{\mathrm i}, p_{\mathrm i})$, and
let $M_{\mathrm o}$ be a window operator from that middle space to an
output space with $(k_{\mathrm o}, s_{\mathrm o}, p_{\mathrm o})$. If the product is defined,
$M_{\mathrm o} \tmul M_{\mathrm i}$ is a window operator with parameters
\begin{equation}
\label{eq:gb-compose}
  s'' = s_{\mathrm o}s_{\mathrm i},
  \qquad
  p'' = p_{\mathrm o}s_{\mathrm i} + p_{\mathrm i},
  \qquad
  k'' = (k_{\mathrm o}-1)s_{\mathrm i} + k_{\mathrm i}.
\end{equation}
\end{lemma}

The core of the composite is
\[
  P''[(\kappa, A), (\gamma, f)]
  = \sum_{\substack{(e,d)\,:\;es_{\mathrm i}+d=f\\ m(A,e) \in [M]}}
    \ \sum_\mu
    P_{\mathrm o}[(\kappa,A),(\mu,e)]\;P_{\mathrm i}[(\mu, m(A,e)),(\gamma,d)],
\]
where $m(A,e) = As_{\mathrm o} - p_{\mathrm o} + e$ and $M$ is the extent of
the middle space.

\begin{proof}[Proof of Lemma~\ref{thm:gb-composition}]
A term of the matrix product is non-zero only if
$m \in \mathcal{W}_{\mathrm o}(A)$ and $n \in \mathcal{W}_{\mathrm i}(m)$.
Write $m = As_{\mathrm o} - p_{\mathrm o} + e$ and
$n = ms_{\mathrm i} - p_{\mathrm i} + d$, both forced and unique by
injectivity of the window map in its second argument. Substituting,
\[
  n = As_{\mathrm o}s_{\mathrm i}
      - (p_{\mathrm o}s_{\mathrm i} + p_{\mathrm i})
      + (es_{\mathrm i} + d),
\]
which gives $s''$ and $p''$. As $e$ and $d$ range over their intervals the
offset $es_{\mathrm i} + d$ takes values in
$[0, (k_{\mathrm o}-1)s_{\mathrm i} + k_{\mathrm i} - 1] = [k'']$. Hence
the support is contained in the window pattern of $\theta''$, and since a
window type imposes no sharing this gives membership. Collecting the terms
landing on each offset yields the core.
\end{proof}

\begin{figure}[htbp]
    \centering
    \fitwidth{
\begin{tikzpicture}[>=stealth, font=\small,line width=.3pt,
    alt={An input axis. Three brackets of width k-inner, their left edges
      s-inner apart, one per tap of the outer window. Above them a single
      bracket spanning all three, annotated as k-outer minus one gaps of
      size s-inner plus one inner window.}]
  \def\u{0.6}
  \draw[->] (-0.25,0) -- ({8*\u},0) node[right,font=\scriptsize] {input};
  \foreach \i in {0,...,7} \draw ({\i*\u},0.06) -- ({\i*\u},-0.06);

  \foreach \s/\lbl in {0/{$e=0$}, 2/{$e=1$}, 4/{$e=2$}}{
    \draw[diagramcolor, line width=.35pt]
      ({\s*\u},0.30) -- ({\s*\u},0.50) -- ({(\s+3)*\u},0.50) -- ({(\s+3)*\u},0.30);
    \node[font=\scriptsize, diagramcolor] at ({(\s+1.5)*\u},0.68) {\lbl};}

  \foreach \s in {0,2}
    \draw[<->, black] ({\s*\u},0.14) -- ({(\s+2)*\u},0.14)
      node[midway, font=\scriptsize, black, fill=white, inner sep=1pt] {$s_{\mathrm i}$};
  \draw[<->, black] ({4*\u},0.14) -- ({7*\u},0.14)
    node[midway, font=\scriptsize, black, fill=white, inner sep=1pt] {$k_{\mathrm i}$};

  \draw[line width=.35pt] (0,1.28) -- (0,1.08) -- ({7*\u},1.08) -- ({7*\u},1.28);
  \node[font=\scriptsize, anchor=west] at ({7.2*\u},1.18)
    {$k'' = (k_{\mathrm o}-1)\,s_{\mathrm i} + k_{\mathrm i}$};
  \node[font=\scriptsize, anchor=west] at ({7.2*\u},0.50)
    {one per outer tap};
\end{tikzpicture}}
    \caption[Composing windows]{Window extent under composition: the
        composite window has extent
        $(k_{\mathrm o}-1)s_{\mathrm i} + k_{\mathrm i}$. Each tap of the
        outer window contributes one inner window; consecutive inner windows
        are $s_{\mathrm i}$ apart, and the composite spans all of them.}
    \label{fig:gb-window}
\end{figure}
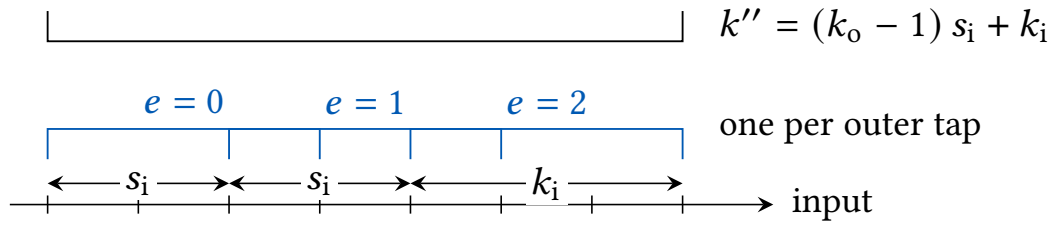

The three parameter laws follow from
one geometric observation (Fig.~\ref{fig:gb-window}): an output position
reads $k_{\mathrm o}$ consecutive middle positions, and the windows of
consecutive output positions start $s_{\mathrm o}$ apart; each middle
position reads $k_{\mathrm i}$ input positions, and the windows of
consecutive middle positions start $s_{\mathrm i}$ apart, so converting a
middle offset to an input offset multiplies it by $s_{\mathrm i}$. Strides
therefore multiply, paddings accumulate with a stride weighting, and the
composite window spans $k_{\mathrm o}-1$ gaps of size $s_{\mathrm i}$
together with the extent of one inner window. In the running example this
gives $2 \cdot 1 + 3 = 5$, the width observed in \S\ref{sec:gb-ov-ops}.

The composite weights are obtained by accumulating products into offset
slots: the pairs of taps $(e, d)$ that land on composite offset $f$ are
those with $es_{\mathrm i} + d = f$, and several pairs may land on one
slot, which produces the sum of three products seen in
\S\ref{sec:gb-ov-ops}. For $k_{\mathrm i} \geq s_{\mathrm i}$, which holds
whenever consecutive inner windows overlap or abut, every offset in
$[k'']$ is of the form $es_{\mathrm i} + d$. For
$k_{\mathrm i} < s_{\mathrm i}$ the offsets of this form are
$k_{\mathrm o}$ runs of length $k_{\mathrm i}$ spaced $s_{\mathrm i}$
apart, and the composite weights at the remaining offsets are $0$.

Lemma~\ref{thm:gb-composition} states a containment, and the support of
the product can be smaller than $\mathcal{S}_{\theta''}$ even when
$k_{\mathrm i} \geq s_{\mathrm i}$. At an output position that reads a
middle position outside the middle space, the inputs that only that middle
position would read have coefficient $0$. With input extent $6$, inner parameters
$(3, 2, 0)$ and outer parameters $(3, 1, 1)$, the window of the second
output position contains input $5$, which neither of the middle positions
it reads depends on.

The sharing relation requires a side condition. Call the composition
\emph{unclipped} if every middle position that an output position reads
lies inside the middle space. When the outer operator has the output extent
of its convolution, with $p_{\mathrm o} \geq 0$ and at least one output
position, this holds exactly when $p_{\mathrm o} = 0$.

\begin{lemma}[Tying under composition]
\label{thm:gb-tying}
Assume the product is defined. If the composition is unclipped, the product
of two $\stblock$ operators is
a $\stblock_{\theta''}$ operator.

Without this hypothesis it is a $\wblock_{\theta''}$ operator, and its output positions that read only
middle positions inside the middle space share a common kernel.
\end{lemma}

\begin{proof}[Proof of Lemma~\ref{thm:gb-tying}]
By Lemma~\ref{thm:gb-composition}, the product has the composite window
support. Let $A$ be an output position such that $m(A,e)\in[M]$ for
every outer offset $e$. The two input cores are independent of their
output-position indices. Hence the composite-core sum ranges over all
pairs $(e,d)$ with $es_{\mathrm i}+d=f$, independently of $A$, and
all such output positions share a kernel. If the composition is
unclipped, this holds at every output position and gives the sharing
relation of $\stblock_{\theta''}$.
\end{proof}

The examples below show that padding can make the boundary rows
differ from the interior rows, so this hypothesis cannot be dropped. With no outer padding, every output position
reads middle positions that exist, so neither the index set of the
accumulation nor the composite weights depend on the output position; this
is the sharing relation of $\stblock_{\theta''}$. With outer padding, the
output positions near the border read middle positions outside the middle
space and acquire weights that differ from the interior ones, so the
sharing holds on the interior alone. When the outer operator has the output
extent of its convolution, at most $\lceil p_{\mathrm o}/s_{\mathrm o}
\rceil$ output positions at each end of a spatial axis read a middle
position outside the middle space. In two dimensions these positions form
a border of the output, and the interior is the product of the interiors
of the two axes.

A product with a $\diagblock$ scales rows or columns. Scaling cannot make a
zero non-zero, so the support and the parameters are unchanged, while two
positions related by the sharing relation lie in different rows, acquire
different scale factors, and are no longer constrained to be equal.
Multiplying a $\stblock_\theta$ by a $\diagblock$ on either side therefore
yields a $\wblock_\theta$, a product of two diagonals is diagonal, and a
product with a $\dblock$ is dense. Table~\ref{tab:gb-matmul} collects the
rules and cites the result that establishes each.

\begin{table}[htbp]
    \centering\small
    \caption[Result types for tensor multiplication]{Result types for
        $\tau_1 \tmul \tau_2$, with $\tau_1$ down the side and $\tau_2$
        across the top. The composite parameters $\theta''$ are those of
        Equation~\ref{eq:gb-compose}, and $\dagger$ marks the entry that
        requires the composition to be unclipped
        (Lemma~\ref{thm:gb-tying}); without that hypothesis it is
        $\wblock_{\theta''}$. A product of window operators with
        incompatible channel extents, and any product involving a
        $\dblock$, is represented as $\dblock$. The preceding lemmas
        establish the window and diagonal cases.}
    \label{tab:gb-matmul}
    \begin{tabular}{@{}l|cccc@{}}
    \toprule
    $\tmul$ & $\stblock_{\theta_2}$ & $\wblock_{\theta_2}$ & $\diagblock$ & $\dblock$ \\
    \midrule
    $\stblock_{\theta_1}$ & $\stblock_{\theta''}{}^{\dagger}$ & $\wblock_{\theta''}$ & $\wblock_{\theta_1}$ & $\dblock$ \\
    $\wblock_{\theta_1}$  & $\wblock_{\theta''}$ & $\wblock_{\theta''}$ & $\wblock_{\theta_1}$ & $\dblock$ \\
    $\diagblock$          & $\wblock_{\theta_2}$ & $\wblock_{\theta_2}$ & $\diagblock$ & $\dblock$ \\
    $\dblock$             & $\dblock$ & $\dblock$ & $\dblock$ & $\dblock$ \\
    \bottomrule
    \end{tabular}
\end{table}

Table~\ref{tab:gb-matmul} gives the products that the
running example forms. A product of two $\stblock$ operators is a
$\stblock$ when the outer operator is unpadded and a $\wblock$ otherwise
(Lemmas~\ref{thm:gb-composition} and~\ref{thm:gb-tying}); a product of a
$\stblock$ and a $\wblock$ is a $\wblock$, since
Lemma~\ref{thm:gb-composition} supplies the support and the untied operand
admits no sharing; and a product of a $\stblock$ and a $\diagblock$ is a
$\wblock$. The $\wblock$ family is required for closure: a certifier whose
operands are convolutions and diagonals leaves the $\stblock$ family at the
first \relu and remains inside $\wblock$ thereafter, so the products it
forms along such chains of layers are
representable. \S\ref{app:gb-compose} proves in addition that the
parameter laws of Equation~\ref{eq:gb-compose} are associative, which
allows the composite parameters of a chain of layers to be computed once
and reused.

\subsubsection{Further Properties of Window Composition}\label{app:gb-compose}
The composition laws above give a result type that contains every
possible product. We now examine when the resulting window contains
positions that cannot be reached, prove associativity of the parameter
laws, and account for boundary clipping.

For $k_{\mathrm o} \geq 2$, the map $(e,d) \mapsto es_{\mathrm i} + d$ is
injective exactly when $k_{\mathrm i} \leq s_{\mathrm i}$; for
$k_{\mathrm o} = 1$ it is injective whatever $k_{\mathrm i}$ and
$s_{\mathrm i}$, and its image is $[k'']$. For $k_{\mathrm o} \geq 2$ and
$k_{\mathrm i} < s_{\mathrm i}$, the reachable offsets form the union
$\bigcup_{e < k_{\mathrm o}} \{ e s_{\mathrm i}, \ldots,
 e s_{\mathrm i} + k_{\mathrm i} - 1 \}$,
that is $k_{\mathrm o}$ runs of length $k_{\mathrm i}$ spaced
$s_{\mathrm i}$ apart, which is a proper subset of $[k'']$ and is not in
general an arithmetic progression: for $k_{\mathrm i} = 2$,
$s_{\mathrm i} = 3$ and $k_{\mathrm o} = 2$ it is $\{0,1,3,4\}$.
Lemma~\ref{thm:gb-composition} is therefore a strict over-approximation in
that regime.

\begin{lemma}[Window monoid]
\label{lem:gb-monoid}
The triples $(k,s,p)$ with $k,s\in\mathbb Z_{>0}$ and $p\in\mathbb Z$ form a monoid
under the composition laws of Equation~\ref{eq:gb-compose}, with identity
$(1,1,0)$. The composite window
extent is therefore associative.
\end{lemma}

\begin{proof}
Read $(k,s,p)$ as the map sending an output index $A$ to the interval
$[As-p,\ As-p+k-1]$, and define composition as the bounding interval of the
set of input positions reachable from $A$. The inner map is increasing, so
the minimum is attained at the first outer tap and the maximum at the last,
giving the stated laws. Associativity follows because both bracketings of a
triple product describe the bounding interval of the same reachable set, and
$(1,1,0)$ sends $A$ to $[A,A]$.
\end{proof}

In this reading, which ignores clipping, the bounding interval equals the
reachable set when that set is an interval; in particular, this holds
when $k_{\mathrm i} \geq s_{\mathrm i}$.

\paragraph{Clipping at the boundary.}
Let the outer operator have the output extent of its convolution,
$o = \lfloor (M + 2p_{\mathrm o} - k_{\mathrm o})/s_{\mathrm o} \rfloor + 1$,
with $p_{\mathrm o} \geq 0$ and at least one output position. Position $A$ is interior exactly when
$As_{\mathrm o} \geq p_{\mathrm o}$ and
$As_{\mathrm o} - p_{\mathrm o} + k_{\mathrm o} \leq M$. The first fails for
the at most $\lceil p_{\mathrm o}/s_{\mathrm o} \rceil$ positions
$A < p_{\mathrm o}/s_{\mathrm o}$. Writing
$X = M + p_{\mathrm o} - k_{\mathrm o}$, the second fails for
$A > X/s_{\mathrm o}$, and $A \leq o - 1 =
\lfloor (X + p_{\mathrm o})/s_{\mathrm o} \rfloor$ leaves
$\lfloor (X + p_{\mathrm o})/s_{\mathrm o} \rfloor - \lfloor X/s_{\mathrm o}
\rfloor$ such positions. With $X = qs_{\mathrm o} + \rho$ and
$0 \leq \rho < s_{\mathrm o}$, this is
$\lfloor (\rho + p_{\mathrm o})/s_{\mathrm o} \rfloor \leq
\lceil p_{\mathrm o}/s_{\mathrm o} \rceil$. At most
$\lceil p_{\mathrm o}/s_{\mathrm o} \rceil$ positions at each end of the
axis are therefore not interior, and for $p_{\mathrm o} = 0$ none is, so
the composition is unclipped exactly when $p_{\mathrm o} = 0$. In two
dimensions a position is interior when it is interior along both axes.

The hypothesis cannot be dropped. Take one channel,
$k_{\mathrm i} = 3$, $s_{\mathrm i} = 1$, $p_{\mathrm i} = 0$ and
$k_{\mathrm o} = 2$, $s_{\mathrm o} = 1$, $p_{\mathrm o} = 1$, so
$k'' = 4$. For interior $A$ both middle positions are legal and
$P''[A,f] = W_{\mathrm o}[0]W_{\mathrm i}[f] + W_{\mathrm o}[1]W_{\mathrm i}[f-1]$,
while for $A = 0$ the position $m(0,0) = -1$ is clipped and
$P''[0,f] = W_{\mathrm o}[1]W_{\mathrm i}[f-1]$. At $f = 1$ the two differ by
$W_{\mathrm o}[0]W_{\mathrm i}[1]$, and the corresponding input indices $0$
and $1$ are both in range, so the discrepancy appears in the matrix
whenever $W_{\mathrm o}[0]W_{\mathrm i}[1] \neq 0$.

Nor does the weaker condition $k_{\mathrm i} \leq s_{\mathrm i} +
p_{\mathrm i}$ suffice. It makes a middle position
clipped \emph{on the left} read only inputs that are themselves out of
range, so that clipping is invisible; it says nothing about the right
boundary, where a middle position beyond the last is clipped for a
different reason. Take input extent $6$, inner
$(k,s,p) = (3,2,1)$ and outer $(k,s,p) = (3,1,1)$, with every weight $1$.
Here $k_{\mathrm i} = 3 = s_{\mathrm i} + p_{\mathrm i}$, so the condition
holds, and the middle extent is $3$. The last output position reads middle
positions $1, 2, 3$, of which $3$ does not exist. Composite offset $4$ then
carries coefficient $2$ at the first two output positions and $1$ at the
last, so tying fails. Thus $p_{\mathrm o}=0$ is sufficient for preservation of weight
sharing. We do not claim that it is necessary for particular cores.

\subsection{Correctness and Composition}\label{app:gb-sep}
We prove that executing the recorded numerical calls produces the
values prescribed by dense computation. The proof first establishes
correctness of the regional block rules and then combines their results
for complete tensors. We also relate the result to the reference
procedures and extend the argument to a fixed sequence of operations.
All computations use exact arithmetic over $\EReal$, with Boolean
operations interpreted over Boolean values. The expressions
$\infty-\infty$, $0\cdot(\pm\infty)$, $x/0$, and
$(\pm\infty)/(\pm\infty)$ are undefined. A dense computation is defined
when every scalar operation it requires is defined. We write
$e\Downarrow v$ for defined, terminating evaluation of $e$ to $v$.

\paragraph{Primitive contracts.}
The structural routines use fixed policies for refinement, ordering,
type selection, and fresh names. Their contracts are as follows.
\begin{enumerate}[topsep=2pt,itemsep=1pt,label=(\roman*)]
\item $\operatorname{Refine}$ terminates with finitely many ordered,
    disjoint regions covering every operand block box. Each region records
    the chosen restriction or materialization plan used by
    $\operatorname{Align}$. Outside their union, both operands have their
    background values. After alignment,
    each region has at least one block operand. The construction in
    \S\ref{app:gb-refine} establishes these properties, including
    when merging enlarges a region.
\item $\operatorname{Align}$ records a finite sequence that, on a
    well-formed input payload, terminates with correctly typed cores
    denoting each operand restricted to the region. An absent-block
    marker denotes the operand's constant background. Restriction,
    reindexing, and materialization preserve these regional values.
\item $\operatorname{Lift}_r$ is defined exactly when neither
    \textsc{B-Identity} nor \textsc{B-Annihilator} applies. It preserves existing blocks
    and records any calls needed to construct constant blocks for
    background markers. On well-formed inputs, these calls terminate
    with correctly typed constant cores. Its outputs denote the same
    regional arrays.
\item The selected result type admits every defined dense regional
    result. Lemmas~\ref{lem:gb-supp}, \ref{lem:gb-share}, and
    \ref{lem:gb-widen} establish the support, sharing, and widening
    properties used in this selection.
\item Numerical block, conversion, and accumulation routines terminate
    on well-typed inputs whenever their specified dense operation is
    defined. They return a compatible core denoting that result through
    a finite sequence of calls selected by descriptors. These are
    contracts of the numerical primitives, rather than a verification of
    their backend implementation.
\end{enumerate}
Recorded assignments bind fresh core names and preserve existing
bindings. In particular, alignment calls remain in the recorded program
when the subsequent regional rule is \textsc{B-Identity} or \textsc{B-Annihilator}.
Correctness of a numerical routine only conditional on its successful
execution would not suffice for the termination conclusion below.

\paragraph{Interpreting symbolic blocks.}
Let $\nu$ assign numerical cores to core names. A symbolic block
$\beta=(\tau,u)$ is well formed under $\nu$ when $\nu(u)$ is a
compatible core for $\tau$. Its regional array is
$V_\nu(\beta)=\sem{(\tau,\nu(u))}$, where $(\tau,b)$ abbreviates a
block of type $\tau$ and core $b$. For a background marker,
$V_\nu(\underline d)[\iota]=d$ throughout the region.
Write $\langle\kappa,\nu\rangle\Downarrow\nu'$ for execution of a
regional fragment. Such a fragment updates the core-name environment;
the return in \textsc{T-Elementwise} subsequently selects its output tuple.

\begin{lemma}[Correctness of the block rules]\label{lem:gb-regional-rules}
Suppose $\vec\beta\xRightarrow[r]{\odot}(\beta,\kappa)$, the operands
are well formed under $\nu$, and
$V_\nu(\beta_1)\odot V_\nu(\beta_2)\Downarrow D_r$. There is an
extension $\nu'$ of $\nu$ such that
$\langle\kappa,\nu\rangle\Downarrow\nu'$ and
\[
\begin{aligned}
  \beta=(\tau,u)&\implies
      \nu'(u)\text{ is compatible with }\tau
      \ \land\ V_{\nu'}(\beta)=D_r,\\
  \beta=\bot&\implies D_r[\iota]=d'
      \quad\text{for every position }\iota\text{ in }r.
\end{aligned}
\]
\end{lemma}
\begin{proof}
We distinguish the last rule of the derivation.

\emph{B-Identity.} The missing operand has an identity background on its
operand side. Thus the regional operation has the value of the other
operand at every position, including its structural zeros. The rule
returns that operand's type and core name and emits $\epsilon$.
Taking $\nu'=\nu$ gives a well-formed block denoting $D_r$.

\emph{B-Annihilator.} The missing operand has an annihilator background $d_j$.
Every defined scalar result in the region is therefore $d_j$. The
premise $d_j=d'$ ensures that this is exactly the result background,
so no result block is required. The emitted fragment is empty and
$\nu'=\nu$ establishes the second conclusion.

\emph{B-Compute.} The contract of $\operatorname{Lift}_r$ gives a
finite fragment $\lambda$ that preserves both regional arrays and
produces well-typed block operands. The selected type $\tau$ admits
their dense result $D_r$. Since that result is defined, the numerical
contract of $K_{\odot,\vec\tau,\tau}$ gives a terminating execution
whose core represents $D_r$ in $\tau$. Its fresh output name $u$
extends the environment without changing either operand or any earlier
result. Executing $\lambda$ and then this call yields the required
$\nu'$ and output block $(\tau,u)$.
\end{proof}

\subsubsection{The Reference Procedures}\label{sec:gb-correct}\label{app:gb-proof-correct}

We prove that the operations of the \sparse library compute the same
result as their dense counterparts. The argument uses the region and type lemmas established above.

The first concerns the cells over which Algorithm~\ref{alg:gb-binary}
iterates. Each cell of the common refinement lies in at most one block of
each operand, and the cells cover every index that either operand covers
(Lemma~\ref{lem:gb-cells}). For a merged group, the block of an operand is
the one that materialization produces over the bounding box, so the
property holds for the merge policy as well. A case analysis over an index
is therefore exhaustive: either no cell contains the index, or exactly one
does, and the contribution of each operand within it is determinate.

The second concerns the restriction and extension of a block to a cell
(Lemma~\ref{lem:gb-widen}). Restricting a block of type $\tau$ on a box
$r$ to a sub-box $r'$ gives a block whose type is the restriction of $\tau$
to $r'$, and the type that \Call{Align}{} assigns to it lies above that
restriction. Extending the block to a larger box with a background of $0$
keeps its type, read in the larger index space, and extending it with any
other background gives a block that $\dblock$ admits.

These two properties establish that the type the algorithms assign to a
block admits the tensor that the block must hold. The kernels then compute its
core. We take the following as their specification: for blocks
$\beta_1, \beta_2$ on a common box and the type $\tau$ assigned on line~16
of Algorithm~\ref{alg:gb-binary}, \Call{Kernel}{} returns the core $b$ for
which the block $(\tau, b)$ denotes $\sem{\beta_1} \odot \sem{\beta_2}$;
on line~11 of Algorithm~\ref{alg:gb-matmul} it returns the core for which
$(\tau, b)$ denotes $\sem{\gamma_1} \tmul \sem{\gamma_2}$. Such a core
exists exactly when the tensor lies in $\sem{\tau}$, and it is then unique,
since a tensor of type $\tau$ has one value per $\approx$-class.

In the following statement, operations on \sparse tensors denote the
reference procedures, and operations on arrays denote their dense
counterparts. The statement restates the dense correctness property
of g-BCSR operations~\cite{neuton} under the specified kernel contracts.
Removing a redundant result block also preserves these equalities,
since its entries already equal the tensor background.

\begin{theorem}[Correctness]
\label{thm:gb-correct}
Let $\mathcal{T}_1, \mathcal{T}_2$ be \sparse tensors of appropriate
shapes, and let $\odot$ be an element-wise binary operator of
\S\ref{sec:gb-result-types}. Assume the reference algorithms use kernels that compute their dense
operations in the assigned result types. If every scalar operation in the dense
computation of the right-hand side is defined, then
\begin{align*}
  \sem{\mathcal{T}_1 \odot \mathcal{T}_2}
    &\;=\; \sem{\mathcal{T}_1} \odot \sem{\mathcal{T}_2}, \\
  \sem{\mathcal{T}_1 \tmul \mathcal{T}_2}
    &\;=\; \sem{\mathcal{T}_1} \tmul \sem{\mathcal{T}_2}.
\end{align*}
\end{theorem}

\begin{proof}
Both equalities reduce to pointwise identities under the tensor lookup
of \S\ref{sec:background-gbcsr} and block denotation of
Definition~\ref{def:gb-block}. By hypothesis every scalar operation
of the dense computation is defined.

\emph{Element-wise.}
Fix $\mathbf{i}$. By Lemma~\ref{lem:gb-cells}, the cells cover every
operand block and are pairwise disjoint. Alignment represents each
operand over a cell by one block or its background. In a merged cell,
these are the materialized blocks, whose values include the original
background outside the original block boxes. The following cases refer
to these aligned representations, which preserve the operand values.

\emph{No cell contains $\mathbf{i}$.} Both operands give their
backgrounds, and the result gives $d_3$. The dense result
is $d_1 \odot d_2$, which is defined by hypothesis, so line~2 sets $d_3$ to
it.

\emph{Only a block $B$ of $\mathcal{T}_1$ covers $\mathbf{i}$.} The second
operand gives $d_2$, and Algorithm~\ref{alg:gb-binary} distinguishes three
subcases.

If $d_2 = \mathbf{1}^{\mathrm R}_\odot$ the block is copied (line~9), which
agrees since $x \odot \mathbf{1}^{\mathrm R}_\odot = x$ at every position
of the cell, including those the index map of $B$ omits, where the stored
value is $0$ and $0 \odot \mathbf{1}^{\mathrm R}_\odot = 0$.

If $d_2 = \mathbf{0}_\odot$ no block is emitted (line~11) and the result
reads $d_3$. The dense result is
$\mathcal{T}_1[\mathbf{i}] \odot \mathbf{0}_\odot = \mathbf{0}_\odot$, so it
remains to show $d_3 = \mathbf{0}_\odot$. If $d_1 \odot \mathbf{0}_\odot$
is defined it equals $\mathbf{0}_\odot$. Otherwise $\odot$ is
multiplication, the only operator with an annihilator that is undefined at
some arguments, and line~2 sets $d_3 = 0$, which is the annihilator of
multiplication.

Otherwise line~13 lifts $d_2$ to a $\cblock$ over the cell and the general
case applies to the resulting pair of blocks, with $B$ on the left. The
result can be non-zero at positions outside the support of $B$, where $B$
holds $0$ and the result holds $0 \odot d_2$, so retaining the type of
$B$ is not sound in general. Taking $B$ diagonal, $\odot$ addition and
$d_2 = 1$, the result has $1$ at every off-diagonal position of the cell,
while the diagonal index map would report $0$ there. The emitted block is
given the type $\tau(B) \odot^\tau \cblock$ of line~16, or a named type
above it. For an operator disjunctive at $0$, Lemma~\ref{lem:gb-supp}
bounds its support by the union of the two supports, which is the whole
cell. For an operator conjunctive at $0$, it bounds the support by that of
$B$; multiplying a diagonal block by the background $2$, for instance,
gives a diagonal block. In both cases Lemma~\ref{lem:gb-share} gives the
sharing. For any other operator, the type is $\dblock$, which admits every
tensor. By the specification of the kernels, line~17 computes the core for which every
position of the cell, those $B$ stores and those it omits alike, reads
$\mathcal{T}_1[\mathbf{i}] \odot d_2$.

\emph{Only a block $B$ of $\mathcal{T}_2$ covers $\mathbf{i}$.} The first
operand gives $d_1$, and the dense result is
$d_1 \odot \mathcal{T}_2[\mathbf{i}]$. The argument is that of the previous
case with the operands in their order: the block is copied only when
$d_1 = \mathbf{1}^{\mathrm L}_\odot$, the annihilator case is unchanged,
and otherwise $d_1$ is lifted to a $\cblock$ on the left of $B$. For
subtraction and division the copy does not apply, since they have no left
identity, and the cell is computed as $d_1 - B$ or $d_1 \div B$.

\emph{Both cover $\mathbf{i}$.} The cell containing $\mathbf{i}$ is
contained in both boxes by Definition~\ref{def:gb-refine}, so both
contributions are determinate throughout the cell. For an operator
disjunctive or conjunctive at $0$, line~16 assigns the type
$\tau(\beta_1) \odot^\tau \tau(\beta_2)$, or a named type above it, which
admits $\sem{\beta_1} \odot \sem{\beta_2}$ by Lemmas~\ref{lem:gb-supp}
and~\ref{lem:gb-share}. Lemma~\ref{lem:gb-supp} does not apply to the
remaining operators ($\leq$, $\geq$, $=$ and $\div$), and line~16 assigns
them $\dblock$, which admits every tensor. By the specification of
the kernels, line~17 computes the core of that tensor, which reads
$\mathcal{T}_1[\mathbf{i}] \odot \mathcal{T}_2[\mathbf{i}]$ at
$\mathbf{i}$.

The cells are disjoint, so the result is well formed, and
Lemma~\ref{lem:gb-widen} guarantees that the alignment of line~5 gives each
operand a type that admits its values on the cell.

\emph{Multiplication.}
The block enumeration visits only pairs of blocks, so it is sound only when
indices outside every block contribute nothing, which requires both
backgrounds to be $0$. Lines~1--3 dispatch on this: if either background
is non-zero, both operands are materialized densely and multiplied
directly, which computes the sum by definition. When both backgrounds are $0$, an index uncovered in either
operand contributes a product with $0$, so only block pairs contribute.

Fix an output index $(\mathbf{i}, \mathbf{k})$. If no region of
\Call{Meet}{} contains it, then no pair of blocks covers both
$(\mathbf{i}, j)$ and $(j, \mathbf{k})$ for any $j$, since such a pair
would meet on the contracted axis and its output region would contain
$(\mathbf{i}, \mathbf{k})$. Every term of Equation~\ref{eq:gb-matmul} then
has a factor $0$, so the dense result is $0$, which is the background
$d_3$ that the result reads. Otherwise, let $r$ be the region of
\Call{Meet}{} that contains $(\mathbf{i}, \mathbf{k})$. Let $J(\mathbf{i}, \mathbf{k})$ be the set
of contraction indices $j$ at which both $(\mathbf{i}, j)$ and
$(j, \mathbf{k})$ are covered. For each $j\in J$, the covering block pair
is unique because each operand's blocks are disjoint. Its output box
contains $(\mathbf{i},\mathbf{k})$ and, by the endpoint partition of
\S\ref{sec:gb-matmul}, contains all of $r$. Thus \Call{Meet}{} associates
exactly the contributing pairs with $r$, without duplicating any term.
For $j\notin J$, at least one factor is $0$. Line~9 restricts each pair
to the rows and columns of $r$ and its shared range on the contracted
axis, yielding a product over the whole cell. Restricting the free axes of two factors restricts their product to
the same rows and columns, so the partial product of line~11 is the
product of the two blocks over their shared range, read on $r$. By the
specification of the kernels, its core represents that tensor in the type
of line~10, which admits it by Lemmas~\ref{thm:gb-composition}
and~\ref{thm:gb-tying} and the rule for $\diagblock$ in
\S\ref{sec:gb-matmul}. Summing the partial products of $r$ on line~12
gives, at $(\mathbf{i}, \mathbf{k})$,
\[
  \sum_{j \in J(\mathbf{i}, \mathbf{k})}
    \mathcal{T}_1[\mathbf{i}, j]\, \mathcal{T}_2[j, \mathbf{k}]
  \;=\; \sum_j \mathcal{T}_1[\mathbf{i}, j]\, \mathcal{T}_2[j, \mathbf{k}],
\]
which is Equation~\ref{eq:gb-matmul}.
\end{proof}

Theorem~\ref{thm:gb-correct} covers the two binary operations, together
with the restriction and extension that \Call{Align}{} performs. The
library also provides unary operations, which apply a scalar function $f$
to each entry. A block stores $0$ at the positions its index map omits, so
applying $f$ to the core of a block is correct only when $f(0) = 0$. For
other $f$, the library widens the block to represent the values at those
positions. We do not prove the unary operations correct here.

Widening the type of a block at any point preserves the conclusion of
Theorem~\ref{thm:gb-correct}, since the wider type admits the same tensor.
It can lose compression and increase the work of the kernels. This is the
fallback of \S\ref{sec:gb-types}.

\subsubsection{Correctness of the Structural Rules}

\begin{proof}[Proof of Theorem~\ref{thm:gb-separate}]
First consider an element-wise operation whose dense evaluation yields
$D$. The descriptor rules terminate without reading a payload.
Refinement produces a finite ordered collection $\mathcal G$.
Alignment fixes each region's operand types or background markers and
records its calls. There is at least one block operand in every region.
If the other operand is absent, its descriptor background selects
\textsc{B-Identity}, \textsc{B-Annihilator}, or constant lifting followed by
\textsc{B-Compute}. With two block operands, only \textsc{B-Compute}
applies. These mutually exclusive cases exhaust the aligned operand
pairs. All type and call choices follow the fixed policies, including
fresh-name allocation. Thus the result descriptor and finite numerical
fragment depend only on $\vec\pi$.

For \textsc{T-Elementwise}, give each input core name the corresponding
core of $\vec{\mathcal T}$. Process the regions in refinement order. For each
$r$, the contract of alignment gives a terminating execution of
$\alpha_r$ whose symbolic operands denote the original inputs restricted
to $r$. Their dense result is $D|_r$, so
Lemma~\ref{lem:gb-regional-rules} gives a terminating execution of
$\kappa_r$ and a correct block or omission. Fresh bindings preserve
all input cores and earlier regional results. Induction over the regions
therefore establishes termination of their concatenated fragments.
The local return collects the retained core names into a compatible
payload $\rho'$ for the descriptor $\pi'$.

Let $\mathcal T'=\operatorname{Pack}(\pi',\rho')$ and consider an
arbitrary output position. If it belongs to a retained region, its value
in $\mathcal T'$ equals $D$ by the regional lemma. If it belongs to an
omitted region, the lemma gives $D=d'$ throughout that region, so the
output background is correct. Outside every region, both input arrays
have their backgrounds. Since $D$ is defined, $d_1\odot d_2$ must then
be defined and equal to $d'$. If this background combination is
undefined, the defined dense computation therefore has no such outside
position, and choosing fallback background zero is harmless. These
cases cover the domain without overlap, proving $\sem{\mathcal T'}=D$.

For multiplication, a nonzero background selects the dense fallback,
whose contract gives a terminating execution and one output \dblock.
With zero backgrounds, a summand can be nonzero only when both factors
lie in blocks. Because each operand's blocks are disjoint, every such
summand belongs to exactly one pair with a common contracted range.
Partitioning the pairs' output boxes assigns its output position to
exactly one region. Restricting a pair to that region and its shared
contracted range therefore counts every contributing summand once.
The other summands have a zero factor, and positions outside all output
regions have value zero, the result background. The product rules admit
each partial product and the addition rules admit its accumulated
result. Definedness of the dense product makes every scalar product
defined. A defined finite sum over $\EReal$ contains no mixture of
$+\infty$ and $-\infty$, so the partial sums and their fixed-order
accumulation are also defined. The primitive contracts and induction
through this finite order give terminating calls and correctly typed
cores. Reconstruction consequently has the dense product's value.
The regions, types, and calls again depend only on descriptors.

It remains to compare with the existing operation $f_{\mathrm{gb}}$.
Lemma~\ref{lem:gb-model-correspondence} identifies the operand arrays
with those given by the original index-map representation. The dense
correctness property of g-BCSR operations~\cite{neuton}, restated under
the numerical contracts in Theorem~\ref{thm:gb-correct}, gives
$\sem{f_{\mathrm{gb}}(\vec{\mathcal T})}
 =f(\sem{\vec{\mathcal T}})=D$.
Removing a block whose entries equal its background preserves every
lookup, so this equality also covers the operation's redundant-block
removal. This comparison establishes the theorem's final denotational
equality; the preceding proof of the structural rules uses only the
block model, region properties, and primitive contracts.
\end{proof}

\paragraph{Correspondence with the reference procedures.}
Write $f^{\mathrm{ref}}$ for Algorithm~\ref{alg:gb-binary} or
Algorithm~\ref{alg:gb-matmul}, interpreted in the structural block model
with the same fixed policies and with redundant blocks retained.
This notation distinguishes their chosen representations from the
existing operation $f_{\mathrm{gb}}$, which may remove redundant blocks.
The following equality concerns abstract blocks and quotient-indexed
cores; Lemma~\ref{lem:gb-model-correspondence} relates them to physical
index-map storage by denotation.

\begin{lemma}[Reference correspondence]\label{lem:gb-reference-rules}
Under the primitive contracts and fixed policies, suppose
$\vec\pi\xRightarrow{f}(\pi',\kappa)$,
$\descr(\vec{\mathcal T})=\vec\pi$, and
$f(\sem{\vec{\mathcal T}})\Downarrow D$.
For the payload $\rho'$ produced by
$\kappa(\payload(\vec{\mathcal T}))$, we have
\[
  f^{\mathrm{ref}}(\vec{\mathcal T})
       \Downarrow\operatorname{Pack}(\pi',\rho').
\]
The reference procedure executes the numerical calls recorded in
$\kappa$, up to fresh names and output binding. It satisfies
Definition~\ref{def:gb-supported}(i)--(iii).
\end{lemma}
\begin{proof}
For an element-wise operation, the rules and reference procedure use
the same refinement, alignment, background, and type policies.
The algorithm's identity branch is \textsc{B-Identity}; its annihilator
branch is \textsc{B-Annihilator}; and its remaining branch is \textsc{B-Compute},
including any constant lifting. The extra $d_j=d'$ premise of
\textsc{B-Annihilator} holds for the annihilators of the supported operators
(Table~\ref{tab:gb-units}). If the background combination is defined,
annihilation gives this equality. Its only undefined case with an
annihilator is multiplication, whose annihilator and fallback background
are both zero. The branches therefore agree, including their priority.
Both constructions retain alignment calls before these branches and
retain result blocks without testing their cores. They select the same
output positions and types and invoke the same numerical routines in
the same order. The calls terminate by Theorem~\ref{thm:gb-separate};
on the same input payload, they produce the same quotient-indexed cores.

For multiplication, both constructions use the same background test,
output regions, restrictions, selected product types, and accumulation
order. They therefore select the same output representation and
numerical calls. Their calls terminate and return the same cores by
the primitive contracts, establishing the displayed equality in both
cases. In particular,
\[
  \descr(f^{\mathrm{ref}}(\vec{\mathcal T}))=\pi',\qquad
  \payload(f^{\mathrm{ref}}(\vec{\mathcal T}))
       =\kappa(\payload(\vec{\mathcal T})).
\]
Theorem~\ref{thm:gb-separate} supplies dense equivalence, and the
fixed-policy correspondence supplies descriptor separation and the
fixed numerical calls. Thus the reference procedures, equivalently the
operations defined by the structural rules, are supported. Equality of
their representations does not assert equality with a representation
from which $f_{\mathrm{gb}}$ has removed redundant blocks.
\end{proof}

\subsubsection{Sequences of Operations}

\begin{corollary}[Descriptor propagation]
\label{cor:gb-prop}
Let $f_1, \ldots, f_N$ be a fixed sequence of supported
operations, and let the
descriptors of the operands it starts from be
determined by the DNN architecture and the input and batch shapes. Then the
descriptor of every tensor the sequence produces is determined by them as
well, independently of the payloads.
\end{corollary}

\begin{proof}
We induct on $j$. The initial descriptors are fixed by hypothesis.
Suppose that the descriptors of the first $j-1$ results are fixed. The
operands of $f_j$ are inputs or earlier results, so their descriptor
tuple $\vec\pi_j$ is also fixed. The judgment
$\vec\pi_j\xRightarrow{f_j}(\pi_j,\kappa_j)$ determines $\pi_j$
without reading any payload. By condition~(ii) of
Definition~\ref{def:gb-supported}, this is the descriptor of the actual
result. This establishes the induction step.
\end{proof}

Correctness of the composed numerical code follows from
Lemma~\ref{lem:spec-rules}. Its \textsc{S-Seq} case establishes that the
result of one generated fragment supplies the correct inputs to the next.

\section{Architecture Specialization: Definitions and Proofs}\label{app:spec}

This appendix gives the definitions and proofs for program specialization
in \S\ref{sec:spec}: retained redundant blocks, descriptor-determined
control, structural traversals, and the generated specialized program.
Appendix~\ref{app:gbcsr} establishes correctness of the tensor
operations. Here we prove preservation under program composition and
correctness of the complete specialized program.

\subsection{Redundant Blocks}\label{app:gb-canon}

The fragment excludes guards that read payloads, and it excludes one
operation of the general implementation. A
block all of whose entries equal the background of its tensor is redundant,
since dropping it leaves $\sem{\mathcal{T}}$ unchanged, and the general
implementation drops it. The test reads the core, so the number of blocks
in a result, which belongs to the descriptor by
Equation~\eqref{eq:gb-descr}, depends on the payload. A specializing
implementation therefore omits the step, at the cost of retaining the
redundant blocks.

The two implementations then differ. A retained block is invisible to
$\sem{\cdot}$: every index inside it reads $d$ whether the block is
retained or removed, while lookup outside the block is unchanged. Block counts, however, belong to the descriptor, so the two
implementations need not produce equal descriptors, and a block count is
data that a descriptor-determined guard is permitted to read. In a program
with such a guard, the two implementations can take different branches,
and from that point they perform different sequences of operations.
Theorem~\ref{thm:gb-correct} does not close this gap by itself: it states
that each operation agrees with the dense operation it replaces, which
gives agreement along a \emph{fixed} sequence and says nothing about two
executions that diverge.

Theorem~\ref{thm:gb-res} below is therefore stated about the
non-canonicalizing implementation. For the binary operations,
its fixed representations are those of the structural rules, equivalently
the retained-block reference procedures of Lemma~\ref{lem:gb-reference-rules}.
For a generated program the gap does not arise. Apart from tensor-library dispatch, the tests in a generated
program observe tensors only through their dense entries. The expression
replacement of \S\ref{sec:generated-certifiers} likewise depends on the
coefficients, not their block partition. The proof of Theorem~\ref{thm:gb-spec}
uses this to relate $G_\sigma$ to the canonicalizing implementation as
well.

\begingroup
\subsection{Structural Execution}\label{app:gb-dd}

We give the definitions and proofs for the structural execution rules
of \S\ref{sec:gb-dd}. In its structural environment $\Gamma$, a tensor
binding contains a descriptor, and a structural binding contains a known
value. Layer indices,
traversal sets, and known masks are structural bindings. Each variable
has a fixed tensor or structural classification. Thus a tensor target
uses \textsc{S-Assign}, and a structural target uses \textsc{S-Static}.
An assignment to a structural variable must be evaluated entirely from
these bindings and tensor descriptors. Assigning a numerical tensor entry to a counter
does not make that entry structural.

We write $\sem a_\Gamma$ for the value of a structural expression $a$.
This evaluation reads constants, known bindings, and descriptors, and
applies the corresponding deterministic operations to them. It neither
executes numerical calls nor reads unknown cores. Descriptor queries
must refer to the same supported representations as tensor evaluation.
For every ordinary environment $V$ consistent with $\Gamma$, evaluating
$a$ in $V$ must therefore give $\sem a_\Gamma$. This agreement is also
required of structural operation parameters. Values of uniform masks
are known only when established by the fixed inputs or known-value
computations. The type of a uniform block alone does not establish its
value. These requirements make the guard and structural-assignment
conditions of Definition~\ref{def:gb-dguard} explicit.

\subsubsection{Tensor Expressions}\label{app:gb-expressions}

The expression rules of Fig.~\ref{fig:gb-expression-rules} apply the
operation semantics to the tensor programs of \neuton. We use $V$ for an
ordinary execution environment. It assigns tensors to tensor variables
and values to structural variables. The structural environment $\Gamma$
contains the descriptors of these tensors and the known structural
values. The payload environment $R$ contains the cores used by the
generated code. The environments $V$ and $\Gamma$ have the same live
variable domain, partitioned into tensor and structural variables, and
$R$ has exactly its tensor variables. We write $\Gamma\sim(V,R)$ when
\[
\begin{aligned}
  V(x)&=\operatorname{Pack}(\Gamma(x),R(x))
       &&\text{for every tensor variable }x,\\
  V(z)&=\Gamma(z)
       &&\text{for every structural variable }z.
\end{aligned}
\]
The variable classes are fixed by the program. This relation
uses the supported operations' fixed representations. For the binary
operations, these are the retained-block reference representations of
Lemma~\ref{lem:gb-reference-rules}. It does not identify representations
that differ by numerical removal of redundant blocks.

Tensor expressions are variables, tensor constants, and applications
of supported operations. The generic notation $f(\vec i_e)$ includes
tensor operations with their axis, shape, and other structural
parameters. The instance $f_\Gamma$ replaces those parameters by their
values under $\Gamma$. To insert a known structural value as tensor
data, $\operatorname{const}(a)$ permits a structural expression $a$
whose value is a tensor in a fixed representation. We write
$\sem a_\Gamma$ for its evaluation using the known bindings and
descriptors in $\Gamma$, as defined above.
The rule is
\[
  \frac{\mathcal T_0=\sem a_\Gamma}
     {\Gamma\vdash\operatorname{const}(a)
      \leadsto
      (\descr(\mathcal T_0),\payload(\mathcal T_0))}
      \quad\textsc{E-Const}.
\]
A literal tensor is the special case $a=\mathcal T_0$. The payload in
the conclusion is a literal in the generated expression. Thus a mask
computed from a traversal set is fixed separately at each occurrence,
using that iteration's environment.
A known mask is permitted here only if its entries follow from the fixed
inputs or a computation on known structural values. A \cblock type
specifies sharing but does not determine the value of its core.

The numerical expression $\kappa(\vec i_e')$ evaluates
its arguments once, in operand order, and binds them to the input cores
of $\kappa$. It then executes the local numerical calls and returns the
result payload. Intermediate names are local and fresh. Structural
parameters are constants in this expression, so subsequent updates to
$\Gamma$ cannot change an already generated occurrence. Expressions
have no effects on program variables. We write
$R\vdash i_e'\Downarrow\rho$ for this payload evaluation
and $V\vdash i_e\Downarrow\mathcal T$ for ordinary
execution using the supported tensor operations.

\begin{lemma}[Correctness of tensor-expression evaluation]
\label{lem:gb-expression-correct}
Suppose $\Gamma\vdash i_e\leadsto(\pi,i_e')$,
$\Gamma\sim(V,R)$, and every operation in $i_e$ is supported. If
$V\vdash i_e\Downarrow\mathcal T$ and its direct dense
evaluation is defined in exact arithmetic, then there is a payload
$\rho$ such that
\[
  R\vdash i_e'\Downarrow\rho,
  \qquad \operatorname{Pack}(\pi,\rho)=\mathcal T,
  \qquad \sem{\mathcal T}=i_e^{\mathrm{dense}}(V).
\]
Here $i_e^{\mathrm{dense}}(V)$ evaluates the tensor expression using
the dense values of $V$ and the same structural parameters.
\end{lemma}

\begin{proof}
We induct on the structural derivation. For \textsc{E-Var}, the
conclusion records $\Gamma(x)$ and the numerical expression $x$.
Its evaluation returns $R(x)$, and the environment relation reconstructs
$V(x)$. For \textsc{E-Const}, structural-value agreement gives the
same $\mathcal T_0$ in $V$ and $\Gamma$. Its descriptor and literal
payload reconstruct that tensor. Both cases also give equality of
dense values.

For \textsc{E-Op}, apply the induction hypothesis to each operand.
The generated operand expressions return payloads $\vec\rho$ that,
with descriptors $\vec\pi$, reconstruct the ordinary operand tensors
$\vec{\mathcal T}$. The known structural bindings agree in $\Gamma$
and $V$, and every descriptor read returns the same descriptor.
Consequently the two evaluations instantiate the same operation
$f_\Gamma$. The premise
$\vec\pi\xRightarrow{f_\Gamma}(\pi',\kappa)$ and supportedness imply
that $\kappa(\vec\rho)$ terminates with the payload of
$f_\Gamma(\vec{\mathcal T})$, whose descriptor is $\pi'$. This proves
exact reconstruction of the supported result. Dense equivalence gives
\[
  \sem{\operatorname{Pack}(\pi',\kappa(\vec\rho))}
   =f_\Gamma(\sem{\vec{\mathcal T}}).
\]
The operand induction hypotheses identify this with direct dense
expression evaluation. For binary operations these properties follow
from Theorem~\ref{thm:gb-separate} and
Lemma~\ref{lem:gb-reference-rules}. For other operations they are the
supportedness contracts. Local bindings in the generated expression
preserve the operand order and do not change $R$, completing the
induction.
\end{proof}

\subsubsection{Statements}

\begin{definition}[Structural execution]\label{def:gb-sexec}
A structural execution of a statement $i_s$ from $\Gamma$ is a finite
derivation of
$\Gamma\vdash i_s\leadsto(\Gamma',K)$ using
Fig.~\ref{fig:gb-program-rules} and the rules below. Its result consists
of the updated structural environment $\Gamma'$ and the numerical code
$K$. A program body has terminating structural execution when such a
derivation exists.
\end{definition}

A structural execution follows the syntax of the statement. The
\textsc{S-Assign} rule uses the tensor-expression judgment to generate
a numerical assignment. The right-hand side reads the old payload
bindings before the assignment updates its target. Thus successive
iterations may assign the same tensor variable without introducing a
new program-level name. Fresh names are needed only for local
intermediate results inside numerical expressions. The
\textsc{S-Static} rule updates a known binding and emits no code.
\textsc{S-Seq} concatenates the two generated fragments. The while
rules recursively process the body and remaining iterations, testing
the guard in the updated environment each time.

Figure~\ref{fig:gb-program-rules} gives the assignment, sequence,
conditional, and while rules. The empty-statement rule is
\[
\frac{}{\Gamma\vdash\operatorname{skip}
  \leadsto(\Gamma,\epsilon)}
  \quad\textsc{S-Skip}
\]
The rules for ite are
\[
\frac{\sem a_\Gamma=\true \quad
\Gamma\vdash i_{s_1}\leadsto(\Gamma',K)}{\Gamma\vdash\operatorname{ite}(a,i_{s_1},i_{s_2}) \leadsto(\Gamma',K)\\}
  \quad\textsc{S-If-T}
\]
\[
\frac{\sem a_\Gamma=\false \quad
\Gamma\vdash i_{s_2}\leadsto(\Gamma',K)}{\Gamma\vdash\operatorname{ite}(a,i_{s_1},i_{s_2}) \leadsto(\Gamma',K)\\}
  \quad\textsc{S-If-F}
\]
Here $\operatorname{ite}$ is a conditional statement. Tensor-valued
selection whose condition depends on numerical values remains a
numerical operation. The structural rules cannot select a branch using
an unknown guard. Section~\ref{sec:gb-relax} instead specializes
the regions invoked by such runtime decisions.

\paragraph{Ordinary execution and generated code.}
Let $V$ be an environment of tensors and structural values, and $R$ a
payload environment, as in Appendix~\ref{app:gb-expressions}. Ordinary
statement execution is written
$V\vdash i_s\Downarrow V'$. It uses \neuton's
assignment, sequence, and while semantics, with the supported tensor
operations and the structural helpers of $P_s$. Generated-code
execution is written $R\vdash K\Downarrow^{\mathrm{num}}R'$.
An empty fragment leaves $R$ unchanged. An assignment evaluates its
numerical expression and then replaces the target binding. Composition
executes the fragments in order. Intermediate names remain local, so
they cannot overwrite a live program variable. A numerical execution
is defined when every scalar operation in the corresponding dense
computation of each executed tensor operation is defined, and its
structural expressions are defined. Sparse execution alone is
insufficient for this condition, since omitting a zero entry may avoid
an undefined operation in the dense counterpart.

\subsubsection{Correctness and Reuse}

\begin{lemma}[Structural determinacy]\label{lem:gb-trace}
For fixed operation policies, the results of a finite structural
execution are determined by its input statement and environment.
Specifically, if
$\Gamma\vdash i_s\leadsto(\Gamma_1,K_1)$ and
$\Gamma\vdash i_s\leadsto(\Gamma_2,K_2)$, then
$\Gamma_1=\Gamma_2$ and $K_1=K_2$ up to renaming local intermediate
variables. Neither result depends on numerical payloads.
\end{lemma}

\begin{proof}
We induct on the first derivation and compare its final rule with the
second derivation. The statement form fixes the assignment or sequence
rule. For a conditional or while statement, $\sem a_\Gamma$ fixes
which rule applies. Tensor-expression evaluation is deterministic
because operand evaluation and the operation policies are fixed.
Structural assignments evaluate deterministic expressions over known
data. In a sequence, the induction hypothesis gives the same result
of the first statement, and hence the same input environment for the
second. In a true while case it gives the same environment and code
for the body, followed by the same result for the recursive loop.
False while and skip cases both return the input environment and empty
code. Each rule reads only $\Gamma$, establishing independence from
payloads. This argument proves uniqueness of finite derivations, not
that every structural execution terminates.
\end{proof}

\begin{lemma}[Execution of generated code]\label{lem:spec-rules}
Let $\Gamma\vdash i_s\leadsto(\Gamma',K)$ be a
finite structural execution of a descriptor-determined statement with
supported operations. Suppose $\Gamma\sim(V,R)$ and
$V\vdash i_s\Downarrow V'$ is defined. Then there is
$R'$ such that
\[
  R\vdash K\Downarrow^{\mathrm{num}}R',
  \qquad \Gamma'\sim(V',R').
\]
In particular, for each tensor variable $x$ in the final environment,
$\sem{\operatorname{Pack}(\Gamma'(x),R'(x))}=\sem{V'(x)}$.
\end{lemma}

\begin{proof}
We induct on the structural derivation. The relation
$\Gamma\sim(V,R)$ is defined in Appendix~\ref{app:gb-expressions}.
It gives exact reconstruction for tensor bindings and equality for
known structural bindings. Exact reconstruction concerns the supported
representations. Equality with a backend that prunes redundant blocks
is established separately through dense values.

For \textsc{S-Assign}, Lemma~\ref{lem:gb-expression-correct} gives a
payload $\rho$ for the generated right-hand side and a descriptor
$\pi$ reconstructing the ordinary expression result $\mathcal T$.
The structural rule stores $\pi$ at $x$, ordinary execution stores
$\mathcal T$, and the emitted assignment stores $\rho$. All other
bindings are unchanged, so the relation holds for the updated
environments. This also covers assignments whose right-hand side reads
$x$. Expression evaluation precedes the update in both executions.

For \textsc{S-Static}, the structural expression has the same value
$v$ in $\Gamma$ and $V$. Both environments update the corresponding
structural binding to $v$. The rule emits $\epsilon$, leaving $R$
unchanged, which preserves the relation. For \textsc{S-Seq}, apply
the induction hypothesis to the first premise to obtain the intermediate
payload environment. The resulting consistency relation supplies the
hypothesis for the second premise. Executing the concatenated code
therefore produces the required final environment.

For \textsc{S-While-F}, guard agreement implies that ordinary execution
also exits without executing the body. Both sides preserve their
environments. For \textsc{S-While-T}, the ordinary guard is true.
Apply the induction hypothesis to the body premise. It yields related
updated environments, which give agreement on the next guard and allow
the induction hypothesis for the recursive while premise. The two
fragments compose to produce the final payload environment. These
premises are smaller derivations even though the second premise has
the same while syntax. The conditional cases use guard agreement and
the induction hypothesis for the selected branch. The skip case
preserves all bindings. This completes the induction.
\end{proof}

\begin{corollary}[Architecture dependence]\label{cor:gb-arch}
Fix the abstract interpreter, operation policies, and structural
constants. If the initial environment of a descriptor-determined
program is determined by the DNN architecture and input and batch
shapes, its finite structural execution generates the same code for
all compatible numerical inputs. At corresponding assignments, the
tensor descriptors are the same, while the computed payloads may differ.
\end{corollary}

\begin{proof}
The fixed information gives the same initial $\Gamma$.
Lemma~\ref{lem:gb-trace} determines the structural derivation and its
output independently of payloads. Lemma~\ref{lem:spec-rules} relates
each defined numerical execution to this derivation, including the
descriptors assigned in its premises.
\end{proof}

\begin{theorem}[Specialized program]\label{thm:gb-res}
Let $P=(i_s,\vec o)$ have body $i_s$ and returned tensor variables
$\vec o$. Fix specialization inputs $\sigma$ and initial environment
$\Gamma_0(\sigma)$. Assume supported tensor operations,
descriptor-determined control, and a finite derivation
\[
  \Gamma_0(\sigma)\vdash i_s\leadsto(\Gamma_f,K).
\]
Let $G_\sigma$ execute $K$ on the input payloads and return
$\operatorname{Pack}(\Gamma_f(\vec o),\vec o)$. For every compatible
$\mathbf b$ on which $P$ has a defined execution,
\[
  \sem{G_\sigma(\mathbf b)}=\sem{P(\mathbf b)}.
\]
The generated code has a fixed finite sequence of numerical calls
and output reconstruction. Its structural parameters are constants
determined by $\sigma$, and it contains no structural dispatch or
host guard inspecting numerical values.
\end{theorem}

\begin{proof}
Fix a compatible numerical input $\mathbf b$, and let $V_0$ be its
ordinary initial environment and $R_0$ its tensor payloads. Compatibility
means that $\Gamma_0(\sigma)\sim(V_0,R_0)$. By
Lemma~\ref{lem:spec-rules}, executing $K$ gives $R_f$ consistent with
$\Gamma_f$ and the ordinary final environment $V_f$. Reconstructing
the returned variables therefore gives the same tensors and hence the
same dense bounds. By Lemma~\ref{lem:gb-trace}, the derivation and $K$
are fixed by $\sigma$. Each tensor-expression rule emits finite
numerical code with fixed structural arguments. Structural assignments
and guard evaluations emit no code, and sequence and loop rules only
concatenate the fragments of their finite premises. Thus $K$ has the
claimed form. Numerical choices inside its calls remain permitted.
\end{proof}

The finite-derivation premise is necessary. A loop whose guard depends
only on a shape may still run forever. For the transformed programs
$P_s$ of the main theorem, the fixed layer sequence and the structural
traversal termination argument below establish this premise. The proof
is relative to the tensor program produced by \neuton. Its existing
semantic-preservation theorem supplies the connection to the original
\cf specification.
\par\endgroup

\subsection{Relaxing Descriptor Determinacy for Traversals}\label{app:gb-relax}

\begin{algorithm}[tb]
\caption{The structural loop of a \traverse}
\label{alg:gb-strav}
\begin{flushleft}
\textbf{Input:} The loop-carried expression, whose coefficient tensor is
$C$; the direction $\mathit{dir}$, the priority function $\mathit{pri}$,
the stopping condition $\mathit{stop}$ and the replacement function $g$ of
the \traverse. We write $\overline{S}$ for the set of lanes $j$ at which
$\mathit{stop}(n_j)$ does not hold, and $\mathrm{nbr}(\Sigma)$ for the set
of lanes whose neurons are neighbors, in the direction $\mathit{dir}$, of
the neurons of the lanes of $\Sigma$. \\
\textbf{Output:} The expression with which the loop of
\S\ref{sec:generated-certifiers} ends (Lemma~\ref{lem:gb-strav}).
\end{flushleft}
\begin{algorithmic}[1]
\State $B \gets \{\, j \in \overline{S} :
  \text{a block of } C \text{ covers lane } j \,\}$, or
  $B \gets \overline{S}$ if the background of $C$ is not $0$
\While{$B \neq \emptyset$}
  \State $\Sigma \gets \{\, j \in B : \mathit{pri}(n_j) =
    \max_{k \in B} \mathit{pri}(n_k) \,\}$
    \CommentRight{Select}
  \State Replace each term $c_{i,j}\, n_j$, $j \in \Sigma$, of the
    expression of every neuron $i$ by $g(n_j, c_{i,j})$
  \State $B \gets (B \setminus \Sigma) \cup
    (\mathrm{nbr}(\Sigma) \cap \overline{S})$
    \CommentRight{Advance}
\EndWhile
\State \Return the expression
\end{algorithmic}
\end{algorithm}

\begin{definition}[Descriptor-determined up to traversals]
\label{def:gb-ddtrav}
Let $P$ be a program that \compiler generates, and let $P_s$ resolve its
fixed layer dispatch, use its non-canonicalizing library, and replace
the loop of every \traverse by Algorithm~\ref{alg:gb-strav}, whose
set $B$ is a known binding in the structural environment. $P$ is \emph{descriptor-determined
up to its traversals} if $P_s$ is descriptor-determined
(Definition~\ref{def:gb-dguard}). If $P$ already has expanded layer
dispatch, contains no \traverse, and retains redundant blocks, then
$P_s=P$, and the condition is ordinary descriptor determinacy.
\end{definition}

\paragraph{Coverage of active terms.}
A replacement can refer to a neuron beyond an immediate neighbor, as
allowed by Definition~\ref{def:replacement-reach}. The set $B$ advances
along individual edges, so it need not contain every active index at
every iteration. We use the following coverage relation to account for
these paths. Let $\mathcal U=\{j:\neg\mathit{stop}(n_j)\}$, which is
$\overline S$ in Algorithm~\ref{alg:gb-strav}. For $B\subseteq\mathcal U$,
an index $j\in\mathcal U$ is \emph{covered} by $B$ if $j\in B$, or if
there is a path $j_0,\ldots,j_k=j$, $k\ge1$, with $j_0\in B$ and
$j_{t+1}\in\mathrm{nbr}(\{j_t\})$, such that for every $0\le t<k$,
$j_t\in\mathcal U$ and $\mathit{pri}(n_{j_t})>\mathit{pri}(n_j)$.
Write $\operatorname{Cov}(B)$ for the set of covered indices. Thus every
vertex before the destination, including the starting vertex, has a
strictly greater priority. When $j\in B$, no path is required.

\begin{lemma}[Coverage]\label{lem:spec-coverage}
Let $B\subseteq\mathcal U$ be nonempty, let
$p=\max_{j\in B}\mathit{pri}(n_j)$, and let
$\Sigma=\{j\in B:\mathit{pri}(n_j)=p\}$ and
$B'=(B\setminus\Sigma)\cup(\mathrm{nbr}(\Sigma)\cap\mathcal U)$.
Then:
\begin{enumerate}[topsep=2pt,itemsep=0pt,label=(\roman*)]
\item If $j\in\operatorname{Cov}(B)$, then
    $\mathit{pri}(n_j)\le p$; equality implies $j\in\Sigma$.
\item $\operatorname{Cov}(B)\setminus\Sigma
    \subseteq\operatorname{Cov}(B')$.
\item If $g$ follows the DNN and $j\in\Sigma$, every unstopped neuron
    mentioned by $g(n_j,c)$ has its index in $\operatorname{Cov}(B')$.
\end{enumerate}
\end{lemma}

\begin{proof}
For (i), an index in $B$ has priority at most $p$. A covered index outside
$B$ has priority strictly smaller than that of the starting vertex of
its path, which belongs to $B$. Equality therefore requires membership
in $B$ and hence in $\Sigma$.

For (ii), an index in $B\setminus\Sigma$ remains in $B'$. Otherwise,
choose a path witnessing coverage. If its starting vertex is not
selected, it remains in $B'$ and the same path witnesses coverage.
If it is selected, the next vertex belongs to $B'$: it is a neighbor of
a selected vertex and is unstopped. That vertex either is the
destination, or starts the remaining path, whose vertices still have
priority strictly greater than the destination.

For (iii), let $m$ be an unstopped neuron mentioned by $g(n_j,c)$.
Definition~\ref{def:replacement-reach} supplies a nonempty path from
$n_j$ to $m$ whose intermediate vertices are unstopped and have priority
strictly greater than $m$. Its first vertex after $n_j$ belongs to $B'$.
It either is $m$, or starts a path witnessing coverage of $m$ by $B'$.
\end{proof}

\begin{lemma}[Structural loop]
\label{lem:gb-strav}
Let $g$ satisfy the conditions of \S\ref{sec:generated-certifiers}. Then
Algorithm~\ref{alg:gb-strav} halts. If, in addition, both the original
and structural executions are defined, then it ends with the expression
with which the loop of \S\ref{sec:generated-certifiers} ends.
\end{lemma}

\begin{proof}
The coefficient axis is indexed by the neurons of the DNN, and $n_j$
is the neuron of lane $j$. Both executions use the same replacement
function $g$ and the same abstract state, which remains fixed throughout
the traversal. For the original loop, $E^\ell$ is the tuple of
expressions after $\ell$ iterations, $A_i^\ell$ the active lanes of
neuron $i$, $\pi^\ell$ the largest priority over $\bigcup_i A_i^\ell$,
and $\Sigma_i^\ell$ the selected lanes. Let $\ell^*$ denote its
termination index, whose existence we establish below. For the
structural loop, $F^m$ is the tuple after $m$ iterations, $B^m$ the set
of lanes, $\mu^m$ the largest priority over $B^m$, and $\Sigma^m$ the
selected lanes. The update is
\[
  B^{m+1}=(B^m\setminus\Sigma^m)
    \cup(\mathrm{nbr}(\Sigma^m)\cap\mathcal U).
\]

\emph{Termination.} Number the neurons in a topological order of the DNN,
reversed for forward traversal, so that every neighbor in the traversal
direction precedes its source. Let $r(j)$ be the distinct number of
$n_j$, and let $L^m=\sum_{j\in B^m}2^{r(j)}$. If $B^m\ne\emptyset$,
then $\Sigma^m\ne\emptyset$. Choose $j^*\in\Sigma^m$ of greatest
number. Every index in $\mathrm{nbr}(\Sigma^m)$ has number below
$r(j^*)$, so $j^*$ is not added back. The total contribution of newly
added indices is at most $\sum_{k<r(j^*)}2^k<2^{r(j^*)}$, while
removing $j^*$ subtracts $2^{r(j^*)}$. Thus $L^{m+1}<L^m$, proving
termination. The same measure on $\bigcup_i A_i^\ell$ proves termination
of the original loop: every new active neuron introduced by $g$ is
reachable in the traversal direction and therefore has smaller number
than its source. Stopped neurons do not enter the measure. In
particular, $\ell^*$ exists. This argument does not require priority
to decrease along every DNN edge.

\emph{Agreement.} Call structural iteration $m$ \emph{effective} if
$\Sigma^m$ contains an index whose coefficient in $F_i^m$ is nonzero
for some $i$, and a zero-work iteration otherwise. Let $\lambda(m)$
count the effective iterations before $m$. We prove by induction that
\begin{equation}\label{eq:traversal-invariant}
  F^m=E^{\lambda(m)},\qquad
  A_i^{\lambda(m)}\subseteq\operatorname{Cov}(B^m)
  \quad\text{for every }i.
\end{equation}
Initially, $F^0=E^0$. A nonzero coefficient lies in a stored block of
$C$, or the background is nonzero. Its index therefore belongs to
$B^0$ whenever it is unstopped, giving
$A_i^0\subseteq B^0\subseteq\operatorname{Cov}(B^0)$.
For the induction step, suppose $B^m\ne\emptyset$ and write
$\ell=\lambda(m)$, $E=F^m=E^\ell$, and $A_i=A_i^\ell$.

\emph{A zero-work iteration.} Every selected coefficient is zero, so
its replacement is $g(n_j,0)=0$, defined by hypothesis. Thus
$F^{m+1}=E^\ell$ and $\lambda(m+1)=\ell$. No active index belongs to
$\Sigma^m$, so Lemma~\ref{lem:spec-coverage}(ii) gives
$A_i\subseteq\operatorname{Cov}(B^{m+1})$.

\emph{An effective iteration.} Choose $j^\circ\in\Sigma^m$ whose
coefficient is nonzero in some $E_{i^\circ}$. Since $B^m\subseteq\mathcal U$,
$j^\circ\in A_{i^\circ}$, and hence
$\pi^\ell\ge\mathit{pri}(n_{j^\circ})=\mu^m$.
Conversely, coverage and Lemma~\ref{lem:spec-coverage}(i) imply
$\pi^\ell\le\mu^m$, with every active index of that priority in
$\Sigma^m$. Therefore $\ell<\ell^*$ and
\[
  \pi^\ell=\mu^m,\qquad
  \Sigma_i^\ell=A_i\cap\Sigma^m\quad\text{for every }i.
\]
A lane of $\Sigma^m\setminus\Sigma_i^\ell$ is unstopped but not
active in $E_i$, so its coefficient is zero. The additional
replacements therefore contribute zero, and the structural update
of $E_i$ equals the original update. Consequently,
$F^{m+1}=E^{\ell+1}$ and $\lambda(m+1)=\ell+1$.
Every active index after the update is either a surviving active index
outside $\Sigma^m$, or is introduced by a replacement of a selected
term. Lemma~\ref{lem:spec-coverage}(ii) covers the first case and
part~(iii) covers the second, preserving the invariant.

Let $M$ be the terminating structural iteration. Since
$B^M=\emptyset$ and $\operatorname{Cov}(\emptyset)=\emptyset$, every
$A_i^{\lambda(M)}$ is empty, giving $\lambda(M)\ge\ell^*$.
Each effective step starts before original termination, so
$\lambda(M)\le\ell^*$. Thus $\lambda(M)=\ell^*$ and
$F^M=E^{\ell^*}$, as required.
\end{proof}

\begingroup
\begin{proof}[Proof of Theorem~\ref{thm:gb-spec}]
Let $D_\sigma$ denote the compatible numerical inputs on which both
$P$ and $P_s$ have defined dense executions in exact arithmetic.

\emph{Structural execution of $P_s$.} The fixed DNN has finitely many
layers. Resolving layer dispatch therefore gives a finite sequence of
transformers in the prescribed topological order. The remaining loops
are structural traversals. Their known traversal sets evolve as in
Algorithm~\ref{alg:gb-strav}, and Lemma~\ref{lem:gb-strav} proves that
this evolution terminates. All remaining statements are finite
sequences, assignments, and conditionals. The descriptor-determinacy
premise supplies the values of their structural expressions and guards.
Supportedness supplies terminating structural evaluation of every
tensor operation. Recursing over these statements and the finite
traversal iterations constructs a finite derivation
$\Gamma_0(\sigma)\vdash i_s\leadsto(\Gamma_f,K)$
for the body of $P_s$.

A structural traversal performs the tensor body with its selected-layer
or selected-lane mask in place of the numerical selection mask. The
structural set determines the entries of this mask, so it can be
inserted as a known tensor constant. This is distinct from assuming
that an unknown uniform block has a known value. Supportedness of the
remaining tensor operations is an explicit premise of the theorem.
Theorem~\ref{thm:gb-res} now applies to $P_s$, and the whole-program
rule returns its generated code with fixed output reconstruction.
For every $\mathbf b\in D_\sigma$, it gives
$\sem{G_\sigma(\mathbf b)}=\sem{P_s(\mathbf b)}$.

\emph{From $P_s$ to $P$.} Fix $\mathbf b\in D_\sigma$. Resolving
layer dispatch and inlining the selected transformers preserves their
order and value dependencies. Outside traversal loops, the two
programs perform the same tensor operations, each with its specified
dense meaning. They therefore reach corresponding traversals with the
same dense coefficients and bounds. Within a traversal, the tensor
body implements the simultaneous replacements specified in
\S\ref{sec:gb-relax}. Its coefficient indices describe tensor
entries, rather than separate host executions for individual neurons.
Under the traversal hypotheses, Lemma~\ref{lem:gb-strav} proves equality
of the returned bound expressions. Additional structural iterations
replace only zero-coefficient terms. Both numerical executions are
defined by the choice of $D_\sigma$, including these extra replacements.
Equality of final expressions alone is not used to establish
definedness. Induction over the surrounding statements consequently
gives $\sem{P_s(\mathbf b)}=\sem{P(\mathbf b)}$.

\emph{Redundant blocks.} The supported implementation used for $P_s$
retains blocks that a general implementation of $P$ may remove.
Dropping a block equal to the background preserves tensor values.
Outside tensor-library dispatch, $P$ observes tensor entries through
their dense values, including the numerical conditions in its traversal
control. Thus removing such blocks changes neither its numerical
branches nor the bounds it returns. Exact representation agreement in
Lemma~\ref{lem:spec-rules} is needed only between structural execution
and the supported execution of $P_s$. Combining the two dense
equalities establishes the theorem.
\end{proof}
\par\endgroup

\subsection{The Implementation of \scepter}\label{app:gb-impl}\label{app:spec-implementation}

\paragraph{Generating the program.}
The implementation performs partial evaluation in two passes: it
records structural information and then constructs the specialized
program. Both passes run before certification. The first executes the
generated code on tensors whose
blocks carry a shape and an element type but no numerical payload, and
keeps the constants that the fixed inputs determine, such as the value of a
stopping condition over a layer. This run records the descriptor of every
tensor that it produces, the layer sizes that each call receives, and the
number of iterations of each traversal loop. Running the generated code
itself lets the library evaluate its own descriptor rules, so no separate
analyzer has to be kept in agreement with it. \scepter then unrolls the
flow over the layers into one call per layer, unrolls each traversal loop
to its recorded iteration count, substitutes the recorded sizes as integer
literals, and replaces the aggregates that carry the structural data by
the tensors they contain. The result is a program whose only operations
are tensor operations, and it computes the payloads during certification.

\paragraph{Descriptor data.}
Part of the data that the run on tensors without data records has a
closed form. The window parameters compose along a chain of layers by
Equation~\ref{eq:gb-compose}, which is associative
(Lemma~\ref{lem:gb-monoid}), so the parameters of an $L$-layer
backsubstitution are the fold of Equation~\ref{eq:gb-compose} over the
layers, and output extents and the result types of
Table~\ref{tab:gb-lattice} are local rules. The rest depends on the
execution: the number of blocks and their boxes after refinement, merging
and alignment depend on the current block lists, the shapes that pass
through the reshape, permute and contraction nodes of the generated code
are below the level of the block algebra, and the counts of symbolic
variables accumulate along the flow. The run obtains all of them.

\paragraph{Tests on tensor values.}
The implementation does not store the set $B$ of
Algorithm~\ref{alg:gb-strav}. It reads the lanes that may be active from
the blocks of the Boolean tensor of active lanes, where, in the run on
tensors without data, a block without data counts as active and a \cblock with a value known from the fixed inputs
counts by that value. The masks of the stopping condition and of the
selected lanes consist of \cblock{}s, so a stopped lane, and a selected
lane once it has been replaced, lies in a \cblock of value false or $0$
and does not count. The compiler evaluates a condition that selects the
result of a transformer once for the whole layer, which is correct when
the condition takes the same value at every neuron of the layer. The run
answers such a condition in the same way, from its blocks, and the answer
is exact when every block of the condition is a \cblock whose value is known from
the fixed inputs, as for the
conditions of \crownibp, \reusecert and \zid, which test the layer index.
A condition that reads a numerical value cannot be resolved by this
structural execution: recording the branch taken by a shape-only execution is
not sufficient. Such a condition must either be known from the fixed
inputs or remain in the surrounding runtime control. The structural traversal
also requires each layer priority to be a \cblock with a value known
from the fixed inputs.

\paragraph{The variant of the structural loop.}
The loop that the implementation runs selects whole layers, and the lanes
that a replacement adds are those of the blocks that it produces, rather
than the neighbors of the selected neurons. The proof of
Lemma~\ref{lem:gb-strav} extends to this variant when it meets the four
obligations that the proof uses for Algorithm~\ref{alg:gb-strav}:
\emph{coverage}, every active lane lies in the set of lanes that the
variant may select; \emph{selection}, an iteration that selects any active lane selects
exactly the maximum-priority active lanes of the original loop; \emph{harmlessness}, every further lane that it selects
has coefficient $0$, so that its replacement leaves the expression
unchanged; and \emph{progress}, the blocks that a replacement produces lie
over neurons that precede the selected ones in the order of the proof, so
that the loop halts. Under these obligations, the iterations that select
an active lane are the iterations of the original loop and the others are
extra iterations, as in the proof of the lemma. These obligations must be checked for the generated program; observed
termination of a structural execution does not by itself establish them. The
prototype does not automatically verify them for arbitrary specifications.

\paragraph{Specializing regions.}
For a region $H$ with fixed input descriptors, supported operations,
and terminating descriptor-determined control, Theorem~\ref{thm:gb-res}
produces a specialized program $G_H$ satisfying
$\sem{G_H(\mathbf b)}=\sem{H(\mathbf b)}$ on their common domain of defined execution.
The region interface must preserve all inputs and results observed by
the enclosing program. Replacing calls to $H$ with calls to $G_H$ then
preserves subsequent numerical decisions: induction over calls shows
that each call returns the same values to the enclosing computation.
The sequence of region calls may depend on those values at runtime.
Fixed input shapes alone are insufficient for specialization; varying
numerical inputs must also preserve all intermediate descriptors.

\paragraph{Selective refinement.}
In \boxpoly, interval bounds select the neurons whose bounds are refined
by a polyhedral analysis. For finite, ordered interval bounds, let $\underline h_{b,j}$ and
$\overline h_{b,j}$ be the interval bounds of ReLU input $j$ in batch
member $b$. The refinement set is
\[
  U=\{j:\exists b,\ \underline h_{b,j}<0<\overline h_{b,j}\}.
\]
Entries that are non-finite or do not form an ordered interval are also
refined. Selection is shared across the batch: selecting $j$ refines it
for all batch members. Neurons stable for every batch member retain their
interval bounds, and the output property is bounded in full. We specialize the refinement region for groups of fixed
shape. Each call supplies a tensor of selected row indices as a numerical
input: its shape is fixed, but its entries and the number of calls are
runtime choices. Row gathering operates on the numerical payload tensors
of the specialized program. Its result layouts must be fixed by the
region dimensions, so changing the selected indices changes values but
not intermediate descriptors. A sparse gather whose block partition
depends on these indices would not satisfy this requirement. The region
gathers the selected rows, computes their refined bounds, and returns
them to their positions. An incomplete group
is padded by repeating its last index, and the padded results are
discarded. Row independence ensures that these additional rows do not
change the retained bounds. This preserves the
selective analysis while applying specialization within each region.

\paragraph{Subsequent tensor optimizations.}
After specialization, the generated program can be optimized by early
reductions, compact block layouts, and fusion of sign-split convolutions
and affine substitutions before tensor compilation. These transformations
operate on the specialized program.
Theorem~\ref{thm:gb-spec} proves equivalence of the specialized program
before these transformations; it does not establish floating-point
identity of the subsequently compiled kernels. 

\section{Memory Transformations of the Specialized Program}\label{app:optimizations}
This appendix describes the semantic requirements of the transformations
in \S\ref{sec:gb-relax}. Their allocation estimates guide a choice of
transformation; the dependence and lifetime checks justify that choice.
We reason about the specialized program supplied to the transformations,
using exact arithmetic and defined operations.

\subsection{Specialized Programs and Region Interfaces}\label{app:memory-analysis}

\label{sec:memory-analysis}
We analyze the specialized program after expanding its numerical routines
into tensor statements,
\[
  G_\sigma=s_1;\cdots;s_m,
  \qquad s_t\equiv x_t\gets f_t(\vec y_t),
\]
where each $f_t$ includes its fixed parameters. Each computed value has
one definition, and its operands are program inputs or earlier results.
Updates are represented by distinct value versions, including dependencies
through shared storage. We reason about the resulting operations as pure
functions of their operands. The dependence graph determines which
statements can be grouped and which values must remain available across
groups.

Let $H$ be a region of this graph, and let $\operatorname{Def}(H)$ and
$\operatorname{Use}(H)$ be its defined and read variables. For the set
$\mathcal O$ of program outputs, the region's input and output interfaces
are
\begin{align*}
  \operatorname{In}(H)
    &=\operatorname{Use}(H)\setminus\operatorname{Def}(H),\\
  \operatorname{Out}(H)
    &=\operatorname{Def}(H)\cap
      \bigl(\operatorname{Use}(G_\sigma\setminus H)\cup\mathcal O\bigr).
\end{align*}
Here $G_\sigma\setminus H$ denotes the statements outside $H$.
Every value in $\operatorname{Out}(H)$ must be retained or reconstructed
by a transformation of $H$. Other intermediate values can be released
after their last use. Storage shared by several values remains live
until all uses of those values have completed.

The shapes and element types determine storage estimates for materialized
results. Let $\widehat a(s_t)$ estimate the new storage allocated by
statement $s_t$. A view contributes no new allocation, and shared storage
is counted once. We use
\[
  \widehat A(H)=\sum_{s_t\in H}\widehat a(s_t)
\]
to guide the selection of regions and their sizes. This quantity estimates
new allocations; it is not a bound on peak memory, which also includes
live inputs, retained outputs, and compiler temporaries.

\subsection{Restriction of Independent Rows}\label{app:row-chunking}\label{sec:row-chunking}
Consider a region $H$ in single-assignment form, with shared inputs $W$
and row-indexed inputs $C$. The region's operations are pure. A view may
share storage, but the region does not modify an input through that view.
An in-place update must first be represented by its value dependencies,
including any dependencies through aliases, before the reasoning below
applies.

For each value $x$, the dependence analysis either establishes that $x$ is
shared, establishes a row axis $d_x$, or rejects the candidate region.
These are analysis outcomes, not a claim that tensor dimensions alone
form an abstract domain for dependence. For a set $R$ of row indices,
define the projection $P_R^x$ to be the identity on a shared value and
restriction to $R$ along $d_x$ on a row-indexed value. A statement
$x=f(y_1,\ldots,y_k)$ is accepted with a restricted implementation $f_R$
only if
\begin{equation}\label{eq:chunk-local}
 P_R^x\bigl(f(y_1,\ldots,y_k)\bigr)
 = f_R\bigl(P_R^{y_1}(y_1),\ldots,P_R^{y_k}(y_k)\bigr).
\end{equation}
The fixed parameters of $f_R$ include the group size and the map from
group indices to their original row indices. A tensor kept in shared
storage can be gathered using this map inside $f_R$; its original storage
remains available to all groups. In particular, reading a repeated source
requires the original row indices, not just the group size.

The supported cases have direct justifications:
\begin{itemize}[topsep=2pt,itemsep=1pt]
\item An element-wise operation reads corresponding entries. Operands
that vary along the output row axis are restricted; operands broadcast
along that axis remain whole.
\item A reshape is accepted only when it preserves the contiguous set
of elements belonging to each row. A reshape that mixes the row and
coefficient indices does not satisfy this rule.
\item A reduction is accepted only when it does not reduce the row axis.
\item A matrix product or convolution is accepted only when the tracked
rows are independent output rows or batches, rather than an axis being
contracted. Shared weights are read whole; operands varying with the
output row are restricted consistently.
\item A repeated tensor can be read from its source using the original
index map. If its source has period $p$, row $i$ is read from row
$i\bmod p$. This gives the same rows as materializing the repetition and
then restricting it.
\end{itemize}
Other operations require their own restriction argument. A matching
shape is insufficient to accept them.

Writing $H(C,W)$ for the region's exported row-indexed results and
$H_R$ for the computation restricted to group $R$, these operation rules
establish
\begin{equation}\label{eq:row-restriction}
  H_R(C|_R,W)=H(C,W)|_R.
\end{equation}
The fixed parameters and neuron indices of $H_R$ are adjusted to $R$;
shared results are computed once. The following lemma lifts this
restriction property to an ordered partition of the neurons.

\begin{lemma}[Independent row groups]\label{lem:row-groups}
Suppose every statement in $H$ satisfies Equation~\eqref{eq:chunk-local},
and every result read outside $H$ is either a shared result computed once
or an exported row-indexed result. Let $R_1,\ldots,R_q$ be an ordered
partition of its rows into consecutive ranges. Executing $H_{R_j}$ on
each $R_j$ and concatenating
each exported result along its row axis computes the same exported values
as $H$.
\end{lemma}

\begin{proof}
Fix $R_j$. The supplied operands are the projections of the original
inputs. Induction over the statements, using
Equation~\eqref{eq:chunk-local}, shows that every restricted intermediate
is the projection of its original value. This applies to each exported
result. The groups partition the row axis in order, so concatenating
those projections reconstructs the original value. Shared results are
unchanged.
\end{proof}

The surrounding rewrite treats each group as one unit and orders the
units and unmodified statements by their value dependencies. A cycle
rejects the rewrite. Every value used outside a unit must be exported,
and deallocation follows its last use. Lemma~\ref{lem:row-groups} then
establishes equivalence of the rewritten specialized program by induction
over this order. To reduce memory, \sys selects regions with small
per-neuron exports; returning all large intermediate coefficient tensors
would preserve the values but defeat the purpose of the transformation.

\sys chooses a group size from estimated storage requirements and a
memory budget. Each group computes its exported bounds and releases
its large intermediate tensors before the next group starts. The
candidate region must include the computations that consume those
intermediates: grouping one operation and reconstructing its entire
large result would retain that storage cost. The dependence checks
above apply to the complete region and its exported values.

\subsection{Consecutive Regions}\label{app:region-partitioning}\label{sec:region-partitioning}
Let $s_1;\ldots;s_m$ be a specialized program with unique definitions.
Partition its statements into consecutive nonempty regions
$H_1,\ldots,H_q$. A region receives every value that it reads whose
definition lies outside the region, and exports each value that it defines
and that a later region or the program's result reads. An update to shared
storage is modeled as a new storage version; its later uses are included
when constructing these interfaces.

Writing $H_j=s_{t_{j-1}+1};\cdots;s_{t_j}$ for boundaries
$0=t_0<t_1<\cdots<t_q=m$, define
$I_j=\operatorname{In}(H_j)$ and $O_j=\operatorname{Out}(H_j)$.

\begin{lemma}[Region partitioning]\label{lem:region-partitioning}
Calling $H_1,\ldots,H_q$ in order with input interfaces $I_j$ and output
interfaces $O_j$ computes the same outputs as $G_\sigma$.
\end{lemma}

\begin{proof}
The first region receives the original program inputs. Assume the inputs
of $H_j$ have their original values. Its statements execute in their
original order, and therefore compute their original results. Every result
needed by a later region is exported. Consequently, the inputs of
$H_{j+1}$ have their original values as well. Induction gives the claim
for the final outputs. Deleting a value after its last use cannot affect
any later statement.
\end{proof}

To choose region boundaries, \sys scans statements in order and
accumulates their estimated new allocations $\widehat a(s_t)$. It
starts a new region when the current region is nonempty and adding the
next statement would exceed the allocation budget. A statement remains
indivisible: if its estimated allocation alone exceeds the budget, it
occupies its own region unless independent row grouping reduces its size. Region boundaries restrict fusion but provide points
at which dead storage can be released, subject to the lifetime and
alias checks above. Within each specialized program, the boundaries and region call
sequence are fixed before certification.

Boundary selection is independent of this equivalence argument. The
allocation heuristic counts newly introduced storage, avoiding duplicate
counts for aliases. It does not include all storage live on entry to a
region or all temporaries chosen by a tensor compiler. Likewise, the
chunk-size heuristic need not account for every allocation outside its
selected region. The budget is therefore a planning parameter rather
than a theorem about peak device memory.

Both transformations preserve outputs in exact arithmetic. Changes to
tensor shapes and compilation boundaries can affect floating-point
rounding; the transformations do not establish bitwise equality of
compiled executions.

\section{The Running Example}\label{app:example}
This appendix expands the running example of \S\ref{sec:spec-overview}
and Fig.~\ref{fig:gb-running}.
For readability, the coefficient matrices use one spatial dimension and
one channel. The network consists of two width-three convolutions separated
by a ReLU; its layer widths are $7$, $5$, $5$ and $3$.

\subsection{Operations over Structured Blocks}\label{sec:gb-ov-ops}

We consider two kinds of operation over these tensors: element-wise
operations, written $\odot$, and tensor multiplication, written $\tmul$ as
in the generated code.

\paragraph{Element-wise operations.}
Let two coefficient tensors be combined by a binary operator. Their blocks
need not coincide, so the first step is to partition the index space into
regions on which both operands are \emph{uniform}, meaning that each
region is covered throughout by a single block of that operand or lies
entirely in its background; only such a region can be passed to one
library call. We call such a partition a \emph{common refinement} of the
two block lists. Layer alignment supplies a common
refinement at no cost whenever both operands are aligned to the layers of
the same network, which is the usual case, and the general construction of
\S\ref{sec:gb-elementwise} is required only when an earlier operation has
changed a block's extent so that a boundary falls inside a layer.

Two questions remain once the regions are fixed. The first is whether a
region requires any computation at all. A region covered by one operand
alone meets the other operand's background, and the algebraic role of that
scalar frequently determines the result without reading data: a background
of $0$ under addition copies the block, a background of $0$ under
multiplication deletes it, and a background of $-\infty$ under maximum
copies it. The second is the type of the result. The product of a diagonal
and a dense tensor is diagonal, since multiplication by zero yields zero,
whereas their sum is dense, since addition preserves the off-diagonal
entries of the dense operand. \S\ref{sec:gb-elementwise} states the
rule.

\paragraph{Matrix multiplication.}
The whole-tensor call $L_4\tmul L$ that computes $L_4^2$ in the running example
selects the block product $L_4^3\tmul L_3^2$, which scales column $j$ of
$L_4^3$ by $\lambda_j$:
\[
L_4^2=L_4^3\tmul L_3^2 =
\begin{pmatrix}
w_0\lambda_0 & w_1\lambda_1 & w_2\lambda_2 & & \\
 & w_0\lambda_1 & w_1\lambda_2 & w_2\lambda_3 & \\
 & & w_0\lambda_2 & w_1\lambda_3 & w_2\lambda_4
\end{pmatrix}
\]
The support is preserved, since scaling a zero yields zero. The sharing is
not: row 0 holds $w_0\lambda_0$ where row 1 holds $w_0\lambda_1$, and the
two are equal only by coincidence. The product occupies the same positions
as $L_4^3$ but stores nine independent values where $L_4^3$ stored three; it
is a $\wblock$. Multiplication by the slope matrix $L_3^2$ therefore preserves
the support of $L_4^3$ and removes its weight sharing. This product arises
whenever backsubstitution passes a \relu between two convolutions.

For the nonnegative weights of the second convolution in the running example,
multiplying by $L_2^1$ and expanding the lower-bound expression
$\underline y_0 = w_0\lambda_0 h_0 + w_1\lambda_1 h_1 + w_2\lambda_2 h_2$
for $y_0$, with each
$h_j$ substituted, gives
\begin{align*}
\underline y_0 \;=\;& w_0\lambda_0v_0\,x_0
 \;+\; (w_0\lambda_0v_1 + w_1\lambda_1v_0)\,x_1
 \;+\; (w_0\lambda_0v_2 + w_1\lambda_1v_1 + w_2\lambda_2v_0)\,x_2 \\
 &+\; (w_1\lambda_1v_2 + w_2\lambda_2v_1)\,x_3
 \;+\; w_2\lambda_2v_2\,x_4 .
\end{align*}
Two properties follow. The support has widened from three inputs to five,
because the three $h_j$ that $y_0$ reads each read three of the $x_i$ and
their windows overlap; and the coefficient of $x_2$ is a sum of three
products, because three pairs of taps map to the same input offset.
\S\ref{sec:gb-matmul} shows that this is general: the product of two
convolution-supported operators is convolution-supported, with window
width $(k_{\mathrm o}-1)s_{\mathrm i} + k_{\mathrm i}$, so the family is
closed under the products a certifier forms.

\section{Benchmark Inventory}\label{app:benchmarks}
\begingroup
\footnotesize
\setlength{\tabcolsep}{2pt}
\begin{longtable}{@{}rllllrrr@{}}
\caption{The 44 evaluated networks in plot order. Plot indices 1--28 are small, 29--35 medium, 36--42 large, and 43--44 very large. Source IDs preserve the identifiers in the experimental records. Neuron counts use the source inventory's ReLU-neuron convention. Training methods are given in parentheses. $\epsilon$ is the perturbation radius.}
\label{tab:eval-networks}\\
\toprule
Plot & Source ID & Name & Architecture & Dataset & Neurons & Batch & $\epsilon$\\
\midrule
\endfirsthead
\toprule
Plot & Source ID & Name & Architecture & Dataset & Neurons & Batch & $\epsilon$\\
\midrule
\endhead

1 & $N_{1}$ & 3$\times$50 & Fully connected & MNIST & $110$ & $100$ & $0.004$ \\
2 & $N_{2}$ & 3$\times$100 & Fully connected & MNIST & $210$ & $100$ & $0.004$ \\
3 & $N_{3}$ & 5$\times$100 & Fully connected & MNIST & $510$ & $100$ & $0.04$ \\
4 & $N_{4}$ & 6$\times$100 & Fully connected & MNIST & $510$ & $100$ & $0.004$ \\
5 & $N_{5}$ & 9$\times$100 & Fully connected & MNIST & $810$ & $100$ & $0.004$ \\
6 & $N_{6}$ & 6$\times$200 & Fully connected & MNIST & $1{,}010$ & $100$ & $0.004$ \\
7 & $N_{7}$ & 9$\times$200 & Fully connected & MNIST & $1{,}610$ & $100$ & $0.004$ \\
8 & $N_{8}$ & 6$\times$500 (Point) & Fully connected & MNIST & $3{,}000$ & $100$ & $0.004$ \\
9 & $N_{9}$ & 6$\times$500 (PGD 0.1) & Fully connected & MNIST & $3{,}000$ & $100$ & $0.005$ \\
10 & $N_{10}$ & 6$\times$500 (PGD 0.3) & Fully connected & MNIST & $3{,}000$ & $100$ & $0.005$ \\
11 & $N_{11}$ & 4$\times$1024 & Fully connected & MNIST & $3{,}072$ & $100$ & $0.004$ \\
12 & $N_{12}$ & ConvSmall (Point) & Convolutional & MNIST & $3{,}604$ & $10$ & $0.004$ \\
13 & $N_{13}$ & ConvSmall (PGD) & Convolutional & MNIST & $3{,}604$ & $100$ & $0.005$ \\
14 & $N_{14}$ & ConvSmall (DiffAI) & Convolutional & MNIST & $3{,}604$ & $100$ & $0.1$ \\
15 & $N_{15}$ & ConvMed (Point) & Convolutional & MNIST & $5{,}704$ & $10$ & $0.004$ \\
16 & $N_{19}$ & 4$\times$100 & Fully connected & CIFAR-10 & $410$ & $100$ & $4\times10^{-5}$ \\
17 & $N_{20}$ & 6$\times$100 & Fully connected & CIFAR-10 & $610$ & $100$ & $4\times10^{-5}$ \\
18 & $N_{21}$ & 9$\times$200 & Fully connected & CIFAR-10 & $1{,}810$ & $100$ & $4\times10^{-5}$ \\
19 & $N_{22}$ & 6$\times$500 (Point) & Fully connected & CIFAR-10 & $3{,}000$ & $100$ & $4\times10^{-5}$ \\
20 & $N_{23}$ & 6$\times$500 (PGD 0.0078) & Fully connected & CIFAR-10 & $3{,}000$ & $100$ & $4.9\times10^{-4}$ \\
21 & $N_{24}$ & 6$\times$500 (PGD 0.0313) & Fully connected & CIFAR-10 & $3{,}000$ & $100$ & $4.9\times10^{-4}$ \\
22 & $N_{25}$ & 7$\times$1024 & Fully connected & CIFAR-10 & $6{,}144$ & $100$ & $4\times10^{-5}$ \\
23 & $N_{26}$ & ConvSmall (Point) & Convolutional & CIFAR-10 & $4{,}852$ & $100$ & $4\times10^{-5}$ \\
24 & $N_{27}$ & ConvSmall (PGD) & Convolutional & CIFAR-10 & $4{,}852$ & $100$ & $4.9\times10^{-4}$ \\
25 & $N_{28}$ & ConvSmall (DiffAI) & Convolutional & CIFAR-10 & $4{,}852$ & $100$ & $0.00392$ \\
26 & $N_{29}$ & ConvMed (Point) & Convolutional & CIFAR-10 & $7{,}144$ & $100$ & $4\times10^{-5}$ \\
27 & $N_{30}$ & ConvMed (PGD 0.0078) & Convolutional & CIFAR-10 & $7{,}144$ & $10$ & $4.9\times10^{-4}$ \\
28 & $N_{31}$ & ConvMed (PGD 0.0313) & Convolutional & CIFAR-10 & $7{,}144$ & $100$ & $4.9\times10^{-4}$ \\
29 & $N_{16}$ & ConvBig (DiffAI) & Convolutional & MNIST & $48{,}064$ & $10$ & $0.2$ \\
30 & $N_{17}$ & ConvSuper (DiffAI) & Convolutional & MNIST & $88{,}544$ & $10$ & $0.14$ \\
31 & $N_{18}$ & Skip (DiffAI) & Residual & MNIST & $71{,}650$ & $10$ & $0.14$ \\
32 & $N_{32}$ & ConvBig (DiffAI) & Convolutional & CIFAR-10 & $62{,}464$ & $10$ & $0.00196$ \\
33 & $N_{49}$ & ResNet-small & Residual & CIFAR-100 & $55{,}460$ & $1$ & $0.00392$ \\
34 & $N_{50}$ & ResNet-medium & Residual & CIFAR-100 & $55{,}460$ & $1$ & $0.00392$ \\
35 & $N_{52}$ & ResNet-super & Residual & CIFAR-100 & $68{,}836$ & $1$ & $0.00392$ \\
36 & $N_{33}$ & ResNetTiny (DiffAI) & Residual & CIFAR-10 & $311{,}796$ & $1$ & $0.001$ \\
37 & $N_{34}$ & ResNetTiny (PGD) & Residual & CIFAR-10 & $311{,}796$ & $1$ & $4\times10^{-5}$ \\
38 & $N_{35}$ & ResNet18 (DiffAI) & Residual & CIFAR-10 & $558{,}080$ & $1$ & $0.001$ \\
39 & $N_{36}$ & ResNet18 (PGD) & Residual & CIFAR-10 & $558{,}080$ & $1$ & $4\times10^{-5}$ \\
40 & $N_{37}$ & ResNet34 (DiffAI) & Residual & CIFAR-10 & $967{,}680$ & $1$ & $0.001$ \\
41 & $N_{38}$ & ResNet-medium & Residual & TinyImageNet & $172{,}296$ & $1$ & $0.00392$ \\
42 & $N_{51}$ & ResNet-large & Residual & CIFAR-100 & $286{,}820$ & $1$ & $0.00392$ \\
43 & $N_{54}$ & VGG-16 & Convolutional & ImageNet & $13{,}555{,}712$ & $1$ & $7.5\times10^{-5}$ \\
44 & $N_{40}$ & ResNet50 (PGD) & Residual & CIFAR-10 & $2{,}940{,}928$ & $1$ & $4\times10^{-5}$ \\
\bottomrule
\end{longtable}
\endgroup

\clearpage
\section{Additional Per-Network Results}\label{app:spec-eval}
\begin{figure}[h]
\centering
\includegraphics[width=\linewidth]{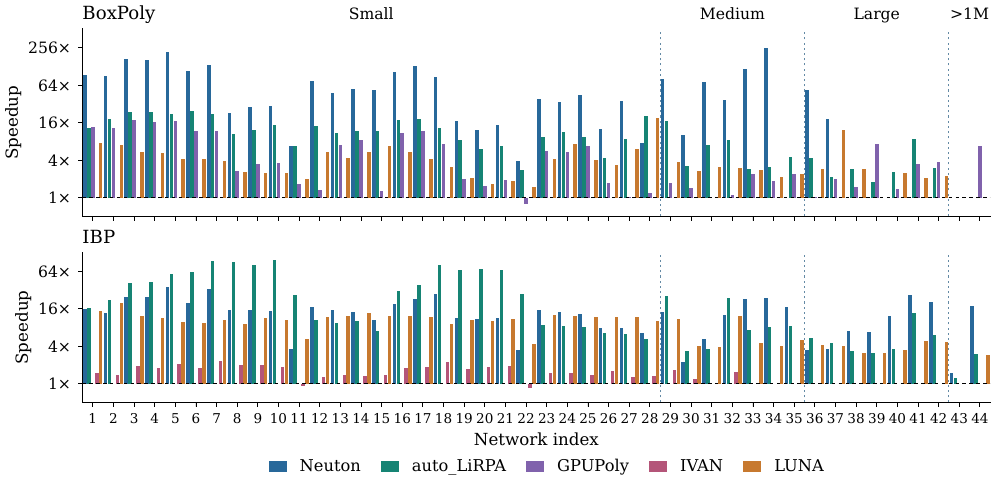}
\caption{Complete BoxPoly and IBP per-network speedups, including configurations where Scepter is slower than a baseline. Bars below $1\times$ indicate slowdowns. Missing bars indicate configurations without a paired result. The network indices and logarithmic spacing follow the main plots.}
\Description{BoxPoly and IBP plots retain all paired results, including the BoxPoly slowdown against GPUPoly and two IBP slowdowns against IVAN.}
\end{figure}

\begin{figure}[h]
\centering
\includegraphics[width=\linewidth]{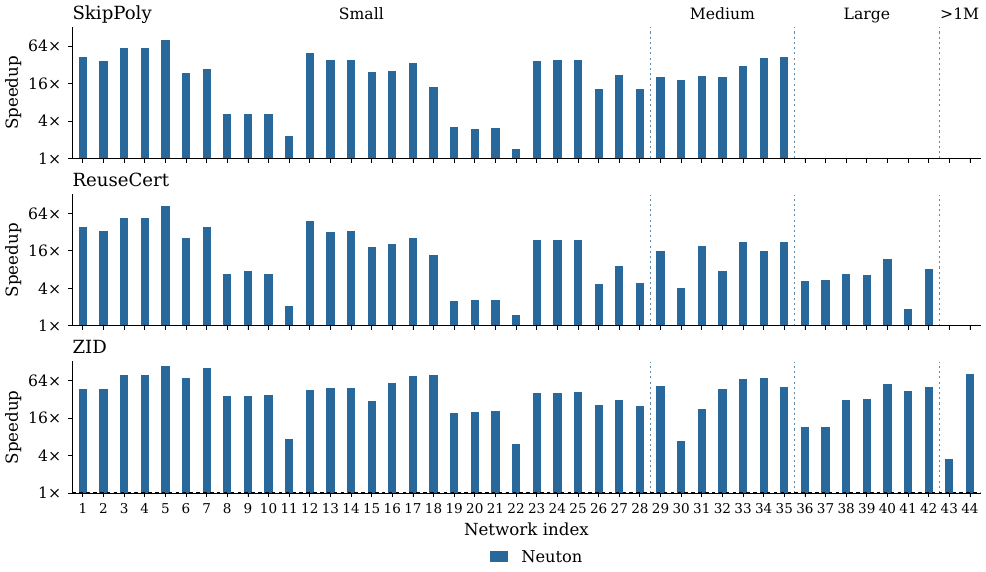}
\caption{Per-network speedups for SkipPoly, ReuseCert, and ZID over Neuton. The network order and size-group separators match the main evaluation plots. Indices map to original network identifiers in the inventory. Missing bars indicate configurations without a paired result.}
\Description{Three aligned bar plots show SkipPoly, ReuseCert, and ZID speedups over Neuton on networks completed by both systems.}
\end{figure}

\clearpage
\section{CPU Results}\label{app:cpu-results}
\begingroup
\setlength{\tabcolsep}{6pt}
\renewcommand{\arraystretch}{1.15}
\begin{longtable}{@{}l@{\extracolsep{\fill}}lr@{}}
\caption{CPU speedups for each abstract interpreter and baseline on the 28 small, seven medium, and seven large networks. Entries are geometric means of baseline runtime divided by Scepter runtime on networks completed by both systems.}
\label{tab:ch8-cpu}\\
\toprule
Abstract interpreter & Baseline & Geometric-mean speedup\\
\midrule
\endfirsthead
\multicolumn{3}{l}{\tablename\ \thetable\ continued}\\
\toprule
Abstract interpreter & Baseline & Geometric-mean speedup\\
\midrule
\endhead
\midrule
\multicolumn{3}{r}{Continued on the next page}\\
\endfoot
\bottomrule
\endlastfoot
DeepPoly & Neuton & $5.58\times$ \\
 & auto\_LiRPA & $2.04\times$ \\
 & IVAN & $19.7\times$ \\
 & ERAN & $87.4\times$ \\
\addlinespace[6pt]
BoxPoly & auto\_LiRPA & $1.78\times$ \\
 & LUNA & $2.90\times$ \\
\addlinespace[6pt]
CROWN-IBP & Neuton & $7.74\times$ \\
 & auto\_LiRPA & $2.24\times$ \\
 & LUNA & $2.90\times$ \\
\addlinespace[6pt]
IBP & Neuton & $5.30\times$ \\
 & auto\_LiRPA & $3.43\times$ \\
 & IVAN & $1.14\times$ \\
 & LUNA & $5.01\times$ \\
\addlinespace[6pt]
DeepZ & Neuton & $4.02\times$ \\
 & IVAN & $4.66\times$ \\
 & ERAN & $522\times$ \\
\addlinespace[6pt]
SkipPoly & Neuton & $5.33\times$ \\
\addlinespace[6pt]
ReuseCert & Neuton & $2.87\times$ \\
\addlinespace[6pt]
ZID & Neuton & $7.79\times$ \\
\midrule
All analyses & Neuton & $5.27\times$ \\
 & auto\_LiRPA & $2.30\times$ \\
 & IVAN & $4.49\times$ \\
 & ERAN & $182\times$ \\
 & LUNA & $3.48\times$ \\
\end{longtable}
The CPU suite consists of the small, medium, and large networks, plot indices 1--42 in Table~\ref{tab:eval-networks}; the two very large networks are not run on the CPU. The final five rows aggregate all completed certifier--network comparisons with each baseline. Only evaluated analysis--baseline combinations are listed.
\endgroup

\section{Memory and CPU--GPU Interaction by Certifier}\label{app:profiles}
\begingroup
\footnotesize
\setlength{\tabcolsep}{4pt}
\begin{longtable}{llrrrrrr}
\caption{Memory use and execution profiles per baseline and certifier (Table~\ref{tab:eval-profiles} for all analyses). Ratios compare the baseline with \scepter. Memory is over the networks both systems complete; idle time, transfer time, and copies are over the networks profiled with both systems, whose number is in the last column. Each Overall row aggregates all of the baseline's certifier--network pairs: geometric means for the ratios and medians for the copies. Copies are median device-to-host (D2H) and host-to-device (H2D) copies per batch, baseline / \scepter. GPUPoly is not included in the memory comparison.}
\label{tab:app-profiles}\\
\toprule
& & \multicolumn{3}{c}{Reduction} & \multicolumn{2}{c}{Copies (median)} &\\
\cmidrule(lr){3-5}\cmidrule(lr){6-7}
Baseline & Certifier & Memory & Idle time & Transfer & D2H & H2D & Networks\\
\midrule
\endfirsthead
\multicolumn{8}{l}{\tablename\ \thetable\ continued}\\
\toprule
& & \multicolumn{3}{c}{Reduction} & \multicolumn{2}{c}{Copies (median)} &\\
\cmidrule(lr){3-5}\cmidrule(lr){6-7}
Baseline & Certifier & Memory & Idle time & Transfer & D2H & H2D & Networks\\
\midrule
\endhead
\bottomrule
\endlastfoot
Neuton & \deeppoly & $3.14\times$ & $128\times$ & $2.34\times$ & $127$ / $0$ & $35$ / $16$ & $38$ \\
 & CROWN-IBP & $1.87\times$ & $50.4\times$ & $1.63\times$ & $29$ / $0$ & $35$ / $16$ & $44$ \\
 & IBP & $1.23\times$ & $13.0\times$ & $1.16\times$ & $1$ / $0$ & $23$ / $16$ & $44$ \\
 & \deepz & $4.61\times$ & $49.8\times$ & $12.9\times$ & $1$ / $0$ & $40$ / $15$ & $34$ \\
 & SkipPoly & $2.51\times$ & $85.2\times$ & $9.22\times$ & $73$ / $0$ & $23$ / $16$ & $35$ \\
 & ReuseCert & $2.96\times$ & $74.2\times$ & $4.44\times$ & $15$ / $0$ & $23$ / $16$ & $41$ \\
 & ZID & $3.81\times$ & $53.7\times$ & $6.63\times$ & $29$ / $0$ & $41$ / $16$ & $44$ \\
\cmidrule(l){2-8}
 & \textit{Overall} & $2.60\times$ & $52.7\times$ & $3.74\times$ & $27$ / $0$ & $32$ / $16$ & $280$ \\
\grpline
auto\_LiRPA & \deeppoly & $1.40\times$ & $14.2\times$ & $11.3\times$ & $281$ / $0$ & $77$ / $16$ & $41$ \\
 & BoxPoly & $1.44\times$ & $10.5\times$ & $11.1\times$ & $423$ / $5$ & $77$ / $16$ & $42$ \\
 & CROWN-IBP & $1.49\times$ & $10.6\times$ & $11.3\times$ & $319$ / $0$ & $67$ / $16$ & $44$ \\
 & IBP & $1.01\times$ & $11.0\times$ & $11.5\times$ & $315$ / $0$ & $65$ / $16$ & $44$ \\
\cmidrule(l){2-8}
 & \textit{Overall} & $1.32\times$ & $11.4\times$ & $11.3\times$ & $356$ / $0$ & $69$ / $16$ & $171$ \\
\grpline
GPUPoly & \deeppoly & --- & $3.52\times$ & $2.69\times$ & $100$ / $0$ & $208$ / $16$ & $41$ \\
 & BoxPoly & --- & $2.59\times$ & $6.08\times$ & $700$ / $5$ & $808$ / $16$ & $42$ \\
\cmidrule(l){2-8}
 & \textit{Overall} & --- & $3.02\times$ & $4.06\times$ & $100$ / $2$ & $208$ / $16$ & $83$ \\
\grpline
IVAN & \deeppoly & $11.2\times$ & $2.74\times$ & $2.83\times$ & $6$ / $0$ & $41$ / $14$ & $26$ \\
 & IBP & $1.10\times$ & $1.37\times$ & $1.76\times$ & $12$ / $0$ & $17$ / $14$ & $31$ \\
 & \deepz & $8.97\times$ & $1.30\times$ & $1.79\times$ & $19$ / $0$ & $17$ / $14$ & $29$ \\
\cmidrule(l){2-8}
 & \textit{Overall} & $4.51\times$ & $1.66\times$ & $2.04\times$ & $10$ / $0$ & $19$ / $14$ & $86$ \\
\grpline
LUNA & BoxPoly & $2.43\times$ & $4.40\times$ & $4.14\times$ & $14$ / $5$ & $61$ / $16$ & $41$ \\
 & CROWN-IBP & $1.45\times$ & $4.48\times$ & $3.20\times$ & $2$ / $0$ & $49$ / $16$ & $43$ \\
 & IBP & $1.51\times$ & $4.97\times$ & $3.44\times$ & $2$ / $0$ & $48$ / $16$ & $43$ \\
\cmidrule(l){2-8}
 & \textit{Overall} & $1.74\times$ & $4.61\times$ & $3.56\times$ & $2$ / $0$ & $49$ / $16$ & $127$ \\
\end{longtable}
\endgroup

\section{GPU Runtime by Configuration}
\label{app:runtime-configs}
\begingroup
\footnotesize
\setlength{\tabcolsep}{3pt}
\renewcommand{\arraystretch}{1.08}
\setlength{\LTleft}{\fill}
\setlength{\LTright}{\fill}
\begin{longtable}{@{}llrrrrrrrrrrr@{}}
\caption[GPU time per image and speedup by configuration]{GPU time per image (ms/image) and speedup by configuration (the $44$ networks of Table~\ref{tab:eval-networks}). Each speedup is the baseline runtime divided by \scepter's runtime for the same configuration, as in Table~\ref{tab:ch8-main}. Each certifier's Geomean row summarizes the completed configurations: times use each system's completed configurations, and speedups use configurations completed by both systems. A dash marks a comparison not included, and NC means a run did not complete.}\label{tab:ch8-runtime-configs}\\
    \toprule
    & & \scepter & \multicolumn{2}{c}{\neuton} & \multicolumn{2}{c}{auto\_LiRPA} & \multicolumn{2}{c}{IVAN} & \multicolumn{2}{c}{GPUPoly} & \multicolumn{2}{c}{LUNA} \\
    \cmidrule(lr){3-3}\cmidrule(lr){4-5}\cmidrule(lr){6-7}\cmidrule(lr){8-9}\cmidrule(lr){10-11}\cmidrule(lr){12-13}
    Certifier & Config & Time & Time & $\times$ & Time & $\times$ & Time & $\times$ & Time & $\times$ & Time & $\times$ \\
    \midrule
\endfirsthead
\multicolumn{13}{c}{\tablename\ \thetable\ (continued)}\\
    \toprule
    & & \scepter & \multicolumn{2}{c}{\neuton} & \multicolumn{2}{c}{auto\_LiRPA} & \multicolumn{2}{c}{IVAN} & \multicolumn{2}{c}{GPUPoly} & \multicolumn{2}{c}{LUNA} \\
    \cmidrule(lr){3-3}\cmidrule(lr){4-5}\cmidrule(lr){6-7}\cmidrule(lr){8-9}\cmidrule(lr){10-11}\cmidrule(lr){12-13}
    Certifier & Config & Time & Time & $\times$ & Time & $\times$ & Time & $\times$ & Time & $\times$ & Time & $\times$ \\
    \midrule
\endhead
\midrule
\multicolumn{13}{r}{Continued on next page}\\
\endfoot
\bottomrule
\endlastfoot
\ibp & $N_{1}$ & $0.0167$ & $0.259$ & $15.6$ & $0.269$ & $16.2$ & $0.0247$ & $1.48$ & --- & --- & $0.251$ & $14.8$ \\
\ibp & $N_{2}$ & $0.0185$ & $0.255$ & $13.9$ & $0.401$ & $21.8$ & $0.0257$ & $1.39$ & --- & --- & $0.375$ & $20.0$ \\
\ibp & $N_{3}$ & $0.0229$ & $0.551$ & $24.3$ & $0.930$ & $41.0$ & $0.0431$ & $1.89$ & --- & --- & $0.287$ & $12.3$ \\
\ibp & $N_{4}$ & $0.0227$ & $0.558$ & $24.7$ & $0.969$ & $42.8$ & $0.0413$ & $1.82$ & --- & --- & $0.256$ & $11.2$ \\
\ibp & $N_{5}$ & $0.0267$ & $0.943$ & $35.4$ & $1.53$ & $57.4$ & $0.0557$ & $2.09$ & --- & --- & $0.268$ & $9.97$ \\
\ibp & $N_{6}$ & $0.0282$ & $0.561$ & $20.0$ & $1.72$ & $61.3$ & $0.0506$ & $1.79$ & --- & --- & $0.262$ & $9.31$ \\
\ibp & $N_{7}$ & $0.0295$ & $0.956$ & $32.6$ & $2.73$ & $93.1$ & $0.0692$ & $2.35$ & --- & --- & $0.318$ & $10.6$ \\
\ibp & $N_{8}$ & $0.0440$ & $0.661$ & $15.2$ & $3.97$ & $91.2$ & $0.0890$ & $2.00$ & --- & --- & $0.400$ & $9.16$ \\
\ibp & $N_{9}$ & $0.0453$ & $0.679$ & $15.0$ & $3.60$ & $80.1$ & $0.0930$ & $2.03$ & --- & --- & $0.533$ & $11.2$ \\
\ibp & $N_{10}$ & $0.0455$ & $0.656$ & $14.5$ & $4.36$ & $96.6$ & $0.0842$ & $1.83$ & --- & --- & $0.483$ & $10.4$ \\
\ibp & $N_{11}$ & $0.130$ & $0.482$ & $3.66$ & $3.45$ & $26.6$ & $0.123$ & $0.908$ & --- & --- & $0.701$ & $5.30$ \\
\ibp & $N_{12}$ & $0.193$ & $3.25$ & $16.9$ & $2.14$ & $10.6$ & $0.249$ & $1.29$ & --- & --- & $2.30$ & $11.9$ \\
\ibp & $N_{13}$ & $0.0209$ & $0.316$ & $15.2$ & $0.198$ & $9.56$ & $0.0288$ & $1.38$ & --- & --- & $0.260$ & $12.3$ \\
\ibp & $N_{14}$ & $0.0222$ & $0.319$ & $14.4$ & $0.246$ & $10.2$ & $0.0292$ & $1.31$ & --- & --- & $0.318$ & $13.9$ \\
\ibp & $N_{15}$ & $0.358$ & $3.84$ & $10.7$ & $2.49$ & $6.94$ & $0.488$ & $1.36$ & --- & --- & $4.37$ & $12.2$ \\
\ibp & $N_{16}$ & $0.461$ & $6.59$ & $14.3$ & $12.0$ & $25.9$ & $0.770$ & $1.66$ & --- & --- & $5.03$ & $10.9$ \\
\ibp & $N_{17}$ & $6.99$ & $15.7$ & $2.24$ & $24.0$ & $3.41$ & $8.26$ & $1.18$ & --- & --- & $27.8$ & $3.97$ \\
\ibp & $N_{18}$ & $2.92$ & $15.0$ & $5.15$ & $10.8$ & $3.61$ & NC & NC & --- & --- & $11.4$ & $3.86$ \\
\ibp & $N_{19}$ & $0.0254$ & $0.479$ & $19.0$ & $0.776$ & $30.7$ & $0.0456$ & $1.78$ & --- & --- & $0.318$ & $12.4$ \\
\ibp & $N_{20}$ & $0.0302$ & $0.700$ & $23.3$ & $1.17$ & $38.9$ & $0.0556$ & $1.84$ & --- & --- & $0.362$ & $11.7$ \\
\ibp & $N_{21}$ & $0.0404$ & $1.13$ & $27.9$ & $3.22$ & $79.7$ & $0.0914$ & $2.26$ & --- & --- & $0.377$ & $9.12$ \\
\ibp & $N_{22}$ & $0.0636$ & $0.716$ & $11.3$ & $4.30$ & $68.1$ & $0.112$ & $1.75$ & --- & --- & $0.675$ & $10.7$ \\
\ibp & $N_{23}$ & $0.0634$ & $0.701$ & $11.1$ & $4.40$ & $69.9$ & $0.118$ & $1.84$ & --- & --- & $0.636$ & $10.1$ \\
\ibp & $N_{24}$ & $0.0607$ & $0.695$ & $11.5$ & $4.04$ & $67.2$ & $0.118$ & $1.94$ & --- & --- & $0.660$ & $11.0$ \\
\ibp & $N_{25}$ & $0.341$ & $1.14$ & $3.44$ & $9.09$ & $27.5$ & $0.290$ & $0.861$ & --- & --- & $1.45$ & $4.37$ \\
\ibp & $N_{26}$ & $0.0233$ & $0.348$ & $15.0$ & $0.201$ & $8.68$ & $0.0351$ & $1.50$ & --- & --- & $0.301$ & $12.7$ \\
\ibp & $N_{27}$ & $0.0240$ & $0.342$ & $14.3$ & $0.202$ & $8.47$ & $0.0357$ & $1.48$ & --- & --- & $0.292$ & $12.0$ \\
\ibp & $N_{28}$ & $0.0261$ & $0.339$ & $13.1$ & $0.209$ & $8.06$ & $0.0364$ & $1.38$ & --- & --- & $0.313$ & $11.7$ \\
\ibp & $N_{29}$ & $0.0484$ & $0.382$ & $7.98$ & $0.318$ & $6.65$ & $0.0782$ & $1.58$ & --- & --- & $0.571$ & $11.9$ \\
\ibp & $N_{30}$ & $0.445$ & $3.52$ & $7.96$ & $2.83$ & $6.18$ & $0.576$ & $1.30$ & --- & --- & $5.17$ & $11.7$ \\
\ibp & $N_{31}$ & $0.0582$ & $0.379$ & $6.55$ & $0.299$ & $5.17$ & $0.0789$ & $1.32$ & --- & --- & $0.595$ & $10.3$ \\
\ibp & $N_{32}$ & $0.523$ & $6.67$ & $12.8$ & $12.7$ & $24.2$ & $0.799$ & $1.52$ & --- & --- & $6.40$ & $12.2$ \\
\ibp & $N_{33}$ & $112$ & $396$ & $3.54$ & $610$ & $5.45$ & NC & NC & --- & --- & $459$ & $4.11$ \\
\ibp & $N_{34}$ & $112$ & $401$ & $3.59$ & $520$ & $4.59$ & NC & NC & --- & --- & $460$ & $4.11$ \\
\ibp & $N_{35}$ & $78.9$ & $551$ & $6.98$ & $269$ & $3.40$ & NC & NC & --- & --- & $250$ & $3.16$ \\
\ibp & $N_{36}$ & $81.3$ & $550$ & $6.77$ & $254$ & $3.12$ & NC & NC & --- & --- & $256$ & $3.15$ \\
\ibp & $N_{37}$ & $117$ & $1{,}446$ & $12.3$ & $428$ & $3.59$ & NC & NC & --- & --- & $405$ & $3.46$ \\
\ibp & $N_{38}$ & $17.0$ & $447$ & $26.8$ & $354$ & $13.5$ & NC & NC & --- & --- & $81.6$ & $4.87$ \\
\ibp & $N_{40}$ & $129$ & $2{,}286$ & $17.7$ & $390$ & $3.02$ & NC & NC & --- & --- & $382$ & $2.94$ \\
\ibp & $N_{49}$ & $9.18$ & $211$ & $23.1$ & $69.5$ & $7.27$ & NC & NC & --- & --- & $41.2$ & $4.53$ \\
\ibp & $N_{50}$ & $15.2$ & $362$ & $23.9$ & $140$ & $8.03$ & NC & NC & --- & --- & $60.2$ & $4.01$ \\
\ibp & $N_{51}$ & $18.3$ & $368$ & $20.3$ & $112$ & $6.17$ & NC & NC & --- & --- & $82.9$ & $4.60$ \\
\ibp & $N_{52}$ & $28.2$ & $467$ & $16.8$ & $280$ & $8.49$ & NC & NC & --- & --- & $141$ & $5.07$ \\
\ibp & $N_{54}$ & $2{,}381$ & $3{,}523$ & $1.48$ & $2{,}906$ & $1.22$ & NC & NC & --- & --- & NC & NC \\
\ibp & \textbf{Geomean} & $0.457$ & $5.42$ & $11.9$ & $6.90$ & $14.8$ & $0.102$ & $1.58$ & --- & --- & $3.01$ & $8.01$ \\
\midrule
\crownibp & $N_{1}$ & $0.0277$ & $1.21$ & $43.9$ & $0.301$ & $10.9$ & --- & --- & --- & --- & $0.368$ & $13.0$ \\
\crownibp & $N_{2}$ & $0.0286$ & $1.16$ & $40.8$ & $0.590$ & $20.0$ & --- & --- & --- & --- & $0.300$ & $10.4$ \\
\crownibp & $N_{3}$ & $0.0385$ & $2.96$ & $77.1$ & $0.972$ & $25.3$ & --- & --- & --- & --- & $0.289$ & $7.49$ \\
\crownibp & $N_{4}$ & $0.0385$ & $2.93$ & $76.1$ & $1.11$ & $28.5$ & --- & --- & --- & --- & $0.309$ & $7.95$ \\
\crownibp & $N_{5}$ & $0.0482$ & $5.29$ & $110$ & $1.60$ & $33.0$ & --- & --- & --- & --- & $0.393$ & $8.01$ \\
\crownibp & $N_{6}$ & $0.0414$ & $2.90$ & $70.4$ & $1.76$ & $42.7$ & --- & --- & --- & --- & $0.300$ & $7.27$ \\
\crownibp & $N_{7}$ & $0.0524$ & $5.17$ & $98.8$ & $2.78$ & $53.1$ & --- & --- & --- & --- & $0.341$ & $6.50$ \\
\crownibp & $N_{8}$ & $0.0699$ & $3.29$ & $47.2$ & $3.97$ & $57.0$ & --- & --- & --- & --- & $0.463$ & $6.63$ \\
\crownibp & $N_{9}$ & $0.0686$ & $3.27$ & $47.8$ & $3.82$ & $55.7$ & --- & --- & --- & --- & $0.491$ & $6.97$ \\
\crownibp & $N_{10}$ & $0.0681$ & $3.28$ & $48.4$ & $4.33$ & $63.9$ & --- & --- & --- & --- & $0.457$ & $6.72$ \\
\crownibp & $N_{11}$ & $0.168$ & $1.57$ & $9.36$ & $3.63$ & $21.7$ & --- & --- & --- & --- & $0.710$ & $4.20$ \\
\crownibp & $N_{12}$ & $0.345$ & $13.8$ & $40.1$ & $2.25$ & $6.47$ & --- & --- & --- & --- & $2.76$ & $7.99$ \\
\crownibp & $N_{13}$ & $0.0361$ & $1.44$ & $40.0$ & $0.228$ & $6.35$ & --- & --- & --- & --- & $0.279$ & $7.72$ \\
\crownibp & $N_{14}$ & $0.0363$ & $1.41$ & $38.9$ & $0.228$ & $6.32$ & --- & --- & --- & --- & $0.291$ & $7.99$ \\
\crownibp & $N_{15}$ & $0.497$ & $14.8$ & $29.8$ & $2.91$ & $5.78$ & --- & --- & --- & --- & $5.09$ & $10.2$ \\
\crownibp & $N_{16}$ & $0.740$ & $31.8$ & $43.1$ & $13.0$ & $17.5$ & --- & --- & --- & --- & $4.77$ & $6.44$ \\
\crownibp & $N_{17}$ & $7.48$ & $41.5$ & $5.54$ & $25.0$ & $3.31$ & --- & --- & --- & --- & $28.5$ & $3.81$ \\
\crownibp & $N_{18}$ & $3.30$ & $69.3$ & $21.1$ & $10.6$ & $3.19$ & --- & --- & --- & --- & $11.4$ & $3.41$ \\
\crownibp & $N_{19}$ & $0.0402$ & $2.26$ & $56.3$ & $0.824$ & $20.5$ & --- & --- & --- & --- & $0.340$ & $8.34$ \\
\crownibp & $N_{20}$ & $0.0498$ & $3.58$ & $72.0$ & $1.22$ & $24.6$ & --- & --- & --- & --- & $0.330$ & $6.61$ \\
\crownibp & $N_{21}$ & $0.0673$ & $6.13$ & $91.4$ & $3.18$ & $47.5$ & --- & --- & --- & --- & $0.536$ & $7.30$ \\
\crownibp & $N_{22}$ & $0.0991$ & $3.34$ & $33.8$ & $4.44$ & $45.0$ & --- & --- & --- & --- & $0.713$ & $7.21$ \\
\crownibp & $N_{23}$ & $0.0962$ & $3.33$ & $34.7$ & $4.38$ & $45.7$ & --- & --- & --- & --- & $0.707$ & $7.37$ \\
\crownibp & $N_{24}$ & $0.0988$ & $3.38$ & $34.4$ & $4.39$ & $44.6$ & --- & --- & --- & --- & $0.705$ & $7.16$ \\
\crownibp & $N_{25}$ & $0.423$ & $3.86$ & $9.30$ & $8.81$ & $21.2$ & --- & --- & --- & --- & $1.60$ & $3.82$ \\
\crownibp & $N_{26}$ & $0.0463$ & $1.43$ & $31.0$ & $0.233$ & $5.04$ & --- & --- & --- & --- & $0.331$ & $7.07$ \\
\crownibp & $N_{27}$ & $0.0462$ & $1.43$ & $31.0$ & $0.232$ & $5.04$ & --- & --- & --- & --- & $0.305$ & $6.59$ \\
\crownibp & $N_{28}$ & $0.0467$ & $1.41$ & $30.2$ & $0.303$ & $6.04$ & --- & --- & --- & --- & $0.294$ & $6.31$ \\
\crownibp & $N_{29}$ & $0.0798$ & $1.51$ & $19.0$ & $0.349$ & $4.40$ & --- & --- & --- & --- & $0.538$ & $6.72$ \\
\crownibp & $N_{30}$ & $0.549$ & $14.4$ & $26.2$ & $3.28$ & $5.70$ & --- & --- & --- & --- & $5.79$ & $10.5$ \\
\crownibp & $N_{31}$ & $0.0905$ & $1.46$ & $16.3$ & $0.319$ & $3.54$ & --- & --- & --- & --- & $0.576$ & $6.35$ \\
\crownibp & $N_{32}$ & $0.792$ & $32.9$ & $41.5$ & $13.4$ & $16.8$ & --- & --- & --- & --- & $6.26$ & $7.86$ \\
\crownibp & $N_{33}$ & $121$ & $1{,}712$ & $14.2$ & $586$ & $4.85$ & --- & --- & --- & --- & $456$ & $3.78$ \\
\crownibp & $N_{34}$ & $119$ & $1{,}684$ & $14.1$ & $651$ & $5.37$ & --- & --- & --- & --- & $466$ & $3.90$ \\
\crownibp & $N_{35}$ & $86.4$ & $3{,}152$ & $36.5$ & $246$ & $2.84$ & --- & --- & --- & --- & $249$ & $2.87$ \\
\crownibp & $N_{36}$ & $88.8$ & $2{,}999$ & $33.8$ & $296$ & $3.26$ & --- & --- & --- & --- & $244$ & $2.75$ \\
\crownibp & $N_{37}$ & $132$ & $8{,}655$ & $65.8$ & $567$ & $4.11$ & --- & --- & --- & --- & $408$ & $3.10$ \\
\crownibp & $N_{38}$ & $33.1$ & $2{,}035$ & $61.9$ & $478$ & $12.0$ & --- & --- & --- & --- & $99.2$ & $3.01$ \\
\crownibp & $N_{40}$ & $156$ & $14{,}470$ & $93.1$ & $434$ & $2.77$ & --- & --- & --- & --- & $408$ & $2.62$ \\
\crownibp & $N_{49}$ & $13.4$ & $1{,}028$ & $77.7$ & $81.8$ & $5.78$ & --- & --- & --- & --- & $48.5$ & $3.65$ \\
\crownibp & $N_{50}$ & $22.4$ & $1{,}988$ & $89.3$ & $108$ & $4.82$ & --- & --- & --- & --- & $66.4$ & $2.98$ \\
\crownibp & $N_{51}$ & $30.2$ & $2{,}022$ & $67.6$ & $130$ & $4.27$ & --- & --- & --- & --- & $91.1$ & $3.04$ \\
\crownibp & $N_{52}$ & $40.2$ & $2{,}467$ & $62.2$ & $537$ & $10.7$ & --- & --- & --- & --- & $148$ & $3.72$ \\
\crownibp & $N_{54}$ & $2{,}396$ & $8{,}094$ & $3.38$ & $3{,}014$ & $1.26$ & --- & --- & --- & --- & NC & NC \\
\crownibp & \textbf{Geomean} & $0.675$ & $24.9$ & $37.0$ & $7.57$ & $11.0$ & --- & --- & --- & --- & $3.20$ & $5.71$ \\
\midrule
\deepz & $N_{1}$ & $0.0207$ & $0.651$ & $31.8$ & --- & --- & $0.0451$ & $2.20$ & --- & --- & --- & --- \\
\deepz & $N_{2}$ & $0.0213$ & $0.664$ & $31.4$ & --- & --- & $0.0520$ & $2.45$ & --- & --- & --- & --- \\
\deepz & $N_{3}$ & $0.0337$ & $2.22$ & $65.9$ & --- & --- & $0.0799$ & $2.38$ & --- & --- & --- & --- \\
\deepz & $N_{4}$ & $0.0341$ & $2.21$ & $65.0$ & --- & --- & $0.0801$ & $2.36$ & --- & --- & --- & --- \\
\deepz & $N_{5}$ & $0.0493$ & $5.64$ & $115$ & --- & --- & $0.108$ & $2.18$ & --- & --- & --- & --- \\
\deepz & $N_{6}$ & $0.0643$ & $2.19$ & $34.1$ & --- & --- & $0.124$ & $1.93$ & --- & --- & --- & --- \\
\deepz & $N_{7}$ & $0.122$ & $5.61$ & $46.1$ & --- & --- & $0.193$ & $1.59$ & --- & --- & --- & --- \\
\deepz & $N_{8}$ & $0.355$ & $2.69$ & $7.60$ & --- & --- & $0.465$ & $1.31$ & --- & --- & --- & --- \\
\deepz & $N_{9}$ & $0.354$ & $2.80$ & $7.87$ & --- & --- & $0.487$ & $1.37$ & --- & --- & --- & --- \\
\deepz & $N_{10}$ & $0.357$ & $2.69$ & $7.53$ & --- & --- & $0.407$ & $1.14$ & --- & --- & --- & --- \\
\deepz & $N_{11}$ & $0.483$ & $1.39$ & $2.88$ & --- & --- & $0.860$ & $1.78$ & --- & --- & --- & --- \\
\deepz & $N_{12}$ & $0.249$ & $8.44$ & $33.8$ & --- & --- & $0.500$ & $2.01$ & --- & --- & --- & --- \\
\deepz & $N_{13}$ & $0.0495$ & $1.14$ & $23.0$ & --- & --- & $0.293$ & $5.93$ & --- & --- & --- & --- \\
\deepz & $N_{14}$ & $0.0502$ & $1.12$ & $22.4$ & --- & --- & $0.298$ & $5.95$ & --- & --- & --- & --- \\
\deepz & $N_{15}$ & $0.701$ & $8.86$ & $12.6$ & --- & --- & $0.973$ & $1.39$ & --- & --- & --- & --- \\
\deepz & $N_{16}$ & $16.7$ & $132$ & $7.93$ & --- & --- & $18.4$ & $1.10$ & --- & --- & --- & --- \\
\deepz & $N_{17}$ & $67.3$ & NC & NC & --- & --- & NC & NC & --- & --- & --- & --- \\
\deepz & $N_{18}$ & $7.59$ & $91.0$ & $12.0$ & --- & --- & NC & NC & --- & --- & --- & --- \\
\deepz & $N_{19}$ & $0.0497$ & $1.57$ & $31.7$ & --- & --- & $0.363$ & $7.31$ & --- & --- & --- & --- \\
\deepz & $N_{20}$ & $0.0692$ & $3.16$ & $45.8$ & --- & --- & $0.394$ & $5.70$ & --- & --- & --- & --- \\
\deepz & $N_{21}$ & $0.285$ & $7.89$ & $27.6$ & --- & --- & $0.716$ & $2.51$ & --- & --- & --- & --- \\
\deepz & $N_{22}$ & $0.735$ & $3.17$ & $4.32$ & --- & --- & $1.40$ & $1.91$ & --- & --- & --- & --- \\
\deepz & $N_{23}$ & $0.737$ & $3.20$ & $4.34$ & --- & --- & $1.55$ & $2.10$ & --- & --- & --- & --- \\
\deepz & $N_{24}$ & $0.734$ & $3.15$ & $4.29$ & --- & --- & $1.54$ & $2.10$ & --- & --- & --- & --- \\
\deepz & $N_{25}$ & $3.52$ & $7.19$ & $2.05$ & --- & --- & $3.91$ & $1.11$ & --- & --- & --- & --- \\
\deepz & $N_{26}$ & $0.0986$ & $1.91$ & $19.4$ & --- & --- & $1.66$ & $16.8$ & --- & --- & --- & --- \\
\deepz & $N_{27}$ & $0.0997$ & $1.89$ & $18.9$ & --- & --- & $1.84$ & $18.5$ & --- & --- & --- & --- \\
\deepz & $N_{28}$ & $0.0987$ & $1.90$ & $19.2$ & --- & --- & $1.91$ & $19.3$ & --- & --- & --- & --- \\
\deepz & $N_{29}$ & $1.04$ & $4.23$ & $4.06$ & --- & --- & $2.71$ & $2.61$ & --- & --- & --- & --- \\
\deepz & $N_{30}$ & $1.11$ & $10.1$ & $9.06$ & --- & --- & $3.23$ & $2.91$ & --- & --- & --- & --- \\
\deepz & $N_{31}$ & $1.04$ & $4.44$ & $4.27$ & --- & --- & $3.48$ & $3.34$ & --- & --- & --- & --- \\
\deepz & $N_{32}$ & $21.6$ & $174$ & $8.02$ & --- & --- & NC & NC & --- & --- & --- & --- \\
\deepz & $N_{33}$ & $2{,}582$ & NC & NC & --- & --- & NC & NC & --- & --- & --- & --- \\
\deepz & $N_{34}$ & $2{,}582$ & NC & NC & --- & --- & NC & NC & --- & --- & --- & --- \\
\deepz & $N_{35}$ & $20{,}745$ & NC & NC & --- & --- & NC & NC & --- & --- & --- & --- \\
\deepz & $N_{36}$ & $20{,}755$ & NC & NC & --- & --- & NC & NC & --- & --- & --- & --- \\
\deepz & $N_{37}$ & NC & NC & NC & --- & --- & NC & NC & --- & --- & --- & --- \\
\deepz & $N_{38}$ & $4{,}555$ & NC & NC & --- & --- & NC & NC & --- & --- & --- & --- \\
\deepz & $N_{40}$ & NC & NC & NC & --- & --- & NC & NC & --- & --- & --- & --- \\
\deepz & $N_{49}$ & $214$ & $1{,}513$ & $7.06$ & --- & --- & NC & NC & --- & --- & --- & --- \\
\deepz & $N_{50}$ & $822$ & $2{,}686$ & $3.27$ & --- & --- & NC & NC & --- & --- & --- & --- \\
\deepz & $N_{51}$ & $6{,}034$ & NC & NC & --- & --- & NC & NC & --- & --- & --- & --- \\
\deepz & $N_{52}$ & $1{,}126$ & $4{,}191$ & $3.72$ & --- & --- & NC & NC & --- & --- & --- & --- \\
\deepz & $N_{54}$ & NC & NC & NC & --- & --- & NC & NC & --- & --- & --- & --- \\
\deepz & \textbf{Geomean} & $2.45$ & $7.23$ & $13.1$ & --- & --- & $0.609$ & $2.83$ & --- & --- & --- & --- \\
\midrule
\deeppoly & $N_{1}$ & $0.0299$ & $1.84$ & $61.7$ & $0.334$ & $11.2$ & $0.239$ & $8.03$ & $0.451$ & $15.1$ & --- & --- \\
\deeppoly & $N_{2}$ & $0.0306$ & $1.88$ & $61.7$ & $0.476$ & $15.6$ & $0.265$ & $8.71$ & $0.462$ & $15.2$ & --- & --- \\
\deeppoly & $N_{3}$ & $0.0518$ & $6.68$ & $129$ & $1.09$ & $21.0$ & $0.522$ & $10.1$ & $1.09$ & $21.0$ & --- & --- \\
\deeppoly & $N_{4}$ & $0.0525$ & $6.59$ & $126$ & $1.15$ & $22.0$ & $0.519$ & $9.90$ & $1.09$ & $20.7$ & --- & --- \\
\deeppoly & $N_{5}$ & $0.0843$ & $15.6$ & $185$ & $2.04$ & $24.0$ & $0.812$ & $9.63$ & $2.01$ & $23.9$ & --- & --- \\
\deeppoly & $N_{6}$ & $0.108$ & $6.65$ & $61.7$ & $1.91$ & $17.7$ & $0.965$ & $8.96$ & $1.22$ & $11.3$ & --- & --- \\
\deeppoly & $N_{7}$ & $0.212$ & $16.6$ & $78.3$ & $3.09$ & $14.5$ & $1.69$ & $7.95$ & $2.37$ & $11.1$ & --- & --- \\
\deeppoly & $N_{8}$ & $0.660$ & $9.46$ & $14.3$ & $4.69$ & $7.10$ & $3.99$ & $6.04$ & $2.30$ & $3.49$ & --- & --- \\
\deeppoly & $N_{9}$ & $0.663$ & $9.39$ & $14.2$ & $4.39$ & $6.61$ & $3.99$ & $6.01$ & $2.31$ & $3.48$ & --- & --- \\
\deeppoly & $N_{10}$ & $0.659$ & $9.69$ & $14.7$ & $5.19$ & $7.87$ & $3.99$ & $6.05$ & $2.30$ & $3.49$ & --- & --- \\
\deeppoly & $N_{11}$ & $0.819$ & $4.22$ & $5.15$ & $4.45$ & $5.44$ & $4.73$ & $5.78$ & $1.52$ & $1.85$ & --- & --- \\
\deeppoly & $N_{12}$ & $0.436$ & $26.8$ & $61.6$ & $2.76$ & $6.19$ & $9.59$ & $22.0$ & $0.952$ & $2.18$ & --- & --- \\
\deeppoly & $N_{13}$ & $0.0806$ & $3.06$ & $38.0$ & $0.410$ & $5.09$ & $8.07$ & $100$ & $0.800$ & $9.93$ & --- & --- \\
\deeppoly & $N_{14}$ & $0.0811$ & $3.08$ & $38.0$ & $0.413$ & $5.10$ & $8.07$ & $99.7$ & $0.802$ & $9.90$ & --- & --- \\
\deeppoly & $N_{15}$ & $0.964$ & $28.3$ & $29.4$ & $3.64$ & $3.77$ & $26.6$ & $27.6$ & $2.24$ & $2.32$ & --- & --- \\
\deeppoly & $N_{16}$ & $11.9$ & $263$ & $22.0$ & $45.0$ & $3.77$ & NC & NC & $61.5$ & $5.15$ & --- & --- \\
\deeppoly & $N_{17}$ & $39.2$ & $522$ & $13.3$ & $96.2$ & $2.45$ & NC & NC & $316$ & $8.04$ & --- & --- \\
\deeppoly & $N_{18}$ & $7.18$ & $177$ & $24.7$ & $28.5$ & $3.95$ & NC & NC & NC & NC & --- & --- \\
\deeppoly & $N_{19}$ & $0.0722$ & $4.63$ & $64.1$ & $0.953$ & $13.2$ & $3.56$ & $49.4$ & $0.894$ & $12.4$ & --- & --- \\
\deeppoly & $N_{20}$ & $0.107$ & $9.33$ & $87.0$ & $1.41$ & $13.2$ & $4.80$ & $44.7$ & $1.47$ & $13.7$ & --- & --- \\
\deeppoly & $N_{21}$ & $0.477$ & $20.3$ & $42.6$ & $3.82$ & $8.01$ & $12.5$ & $26.1$ & $3.09$ & $6.48$ & --- & --- \\
\deeppoly & $N_{22}$ & $1.30$ & $10.8$ & $8.27$ & $5.96$ & $4.57$ & $18.0$ & $13.8$ & $3.13$ & $2.40$ & --- & --- \\
\deeppoly & $N_{23}$ & $1.30$ & $10.8$ & $8.30$ & $5.86$ & $4.50$ & $18.0$ & $13.8$ & $3.15$ & $2.42$ & --- & --- \\
\deeppoly & $N_{24}$ & $1.30$ & $10.6$ & $8.13$ & $5.79$ & $4.45$ & $18.0$ & $13.8$ & $3.13$ & $2.41$ & --- & --- \\
\deeppoly & $N_{25}$ & $6.34$ & $21.0$ & $3.31$ & $15.9$ & $2.51$ & $49.1$ & $7.73$ & $9.02$ & $1.42$ & --- & --- \\
\deeppoly & $N_{26}$ & $0.122$ & $3.87$ & $31.7$ & $0.841$ & $6.89$ & $40.7$ & $334$ & $1.12$ & $9.15$ & --- & --- \\
\deeppoly & $N_{27}$ & $0.122$ & $3.84$ & $31.6$ & $0.841$ & $6.92$ & $40.7$ & $335$ & $1.11$ & $9.13$ & --- & --- \\
\deeppoly & $N_{28}$ & $0.121$ & $3.99$ & $32.9$ & $0.841$ & $6.94$ & $40.7$ & $336$ & $1.11$ & $9.17$ & --- & --- \\
\deeppoly & $N_{29}$ & $0.764$ & $33.2$ & $43.4$ & $15.6$ & $20.4$ & NC & NC & $2.80$ & $3.67$ & --- & --- \\
\deeppoly & $N_{30}$ & $1.25$ & $28.7$ & $23.0$ & $4.57$ & $3.61$ & $85.7$ & $68.5$ & $3.12$ & $2.49$ & --- & --- \\
\deeppoly & $N_{31}$ & $0.761$ & $33.2$ & $43.6$ & $15.9$ & $20.9$ & NC & NC & $2.81$ & $3.69$ & --- & --- \\
\deeppoly & $N_{32}$ & $16.1$ & $323$ & $20.0$ & $56.3$ & $3.49$ & NC & NC & $89.8$ & $5.57$ & --- & --- \\
\deeppoly & $N_{33}$ & $593$ & $25{,}971$ & $43.8$ & $1{,}475$ & $2.48$ & NC & NC & $3{,}804$ & $6.41$ & --- & --- \\
\deeppoly & $N_{34}$ & $595$ & $26{,}178$ & $44.0$ & $1{,}492$ & $2.51$ & NC & NC & $3{,}821$ & $6.42$ & --- & --- \\
\deeppoly & $N_{35}$ & $8{,}017$ & NC & NC & $14{,}754$ & $1.84$ & NC & NC & $97{,}168$ & $12.1$ & --- & --- \\
\deeppoly & $N_{36}$ & $8{,}363$ & NC & NC & $14{,}717$ & $1.76$ & NC & NC & $99{,}311$ & $11.9$ & --- & --- \\
\deeppoly & $N_{37}$ & $45{,}465$ & NC & NC & NC & NC & NC & NC & $492{,}269$ & $10.8$ & --- & --- \\
\deeppoly & $N_{38}$ & $2{,}003$ & $30{,}063$ & $15.0$ & $3{,}437$ & $1.71$ & NC & NC & $17{,}751$ & $8.86$ & --- & --- \\
\deeppoly & $N_{40}$ & $189{,}889$ & NC & NC & NC & NC & NC & NC & NC & NC & --- & --- \\
\deeppoly & $N_{49}$ & $157$ & $5{,}084$ & $32.5$ & $361$ & $2.31$ & NC & NC & $1{,}107$ & $7.07$ & --- & --- \\
\deeppoly & $N_{50}$ & $354$ & $13{,}971$ & $39.5$ & $637$ & $1.80$ & NC & NC & $2{,}427$ & $6.86$ & --- & --- \\
\deeppoly & $N_{51}$ & $1{,}662$ & NC & NC & $2{,}694$ & $1.62$ & NC & NC & $18{,}316$ & $10.8$ & --- & --- \\
\deeppoly & $N_{52}$ & $487$ & $19{,}902$ & $40.8$ & $1{,}491$ & $3.03$ & NC & NC & $3{,}626$ & $7.44$ & --- & --- \\
\deeppoly & $N_{54}$ & $908{,}540$ & NC & NC & NC & NC & NC & NC & NC & NC & --- & --- \\
\deeppoly & \textbf{Geomean} & $5.42$ & $42.3$ & $30.7$ & $14.3$ & $5.69$ & $5.68$ & $22.3$ & $20.4$ & $6.57$ & --- & --- \\
\midrule
\skippoly & $N_{1}$ & $0.0315$ & $1.34$ & $42.5$ & --- & --- & --- & --- & --- & --- & --- & --- \\
\skippoly & $N_{2}$ & $0.0358$ & $1.34$ & $37.4$ & --- & --- & --- & --- & --- & --- & --- & --- \\
\skippoly & $N_{3}$ & $0.0728$ & $4.30$ & $59.1$ & --- & --- & --- & --- & --- & --- & --- & --- \\
\skippoly & $N_{4}$ & $0.0728$ & $4.30$ & $59.1$ & --- & --- & --- & --- & --- & --- & --- & --- \\
\skippoly & $N_{5}$ & $0.122$ & $9.74$ & $80.0$ & --- & --- & --- & --- & --- & --- & --- & --- \\
\skippoly & $N_{6}$ & $0.180$ & $4.29$ & $23.8$ & --- & --- & --- & --- & --- & --- & --- & --- \\
\skippoly & $N_{7}$ & $0.367$ & $9.93$ & $27.0$ & --- & --- & --- & --- & --- & --- & --- & --- \\
\skippoly & $N_{8}$ & $1.22$ & $6.41$ & $5.24$ & --- & --- & --- & --- & --- & --- & --- & --- \\
\skippoly & $N_{9}$ & $1.22$ & $6.25$ & $5.11$ & --- & --- & --- & --- & --- & --- & --- & --- \\
\skippoly & $N_{10}$ & $1.22$ & $6.35$ & $5.19$ & --- & --- & --- & --- & --- & --- & --- & --- \\
\skippoly & $N_{11}$ & $1.48$ & $3.39$ & $2.28$ & --- & --- & --- & --- & --- & --- & --- & --- \\
\skippoly & $N_{12}$ & $0.405$ & $19.8$ & $49.0$ & --- & --- & --- & --- & --- & --- & --- & --- \\
\skippoly & $N_{13}$ & $0.0765$ & $2.90$ & $38.0$ & --- & --- & --- & --- & --- & --- & --- & --- \\
\skippoly & $N_{14}$ & $0.0767$ & $2.88$ & $37.6$ & --- & --- & --- & --- & --- & --- & --- & --- \\
\skippoly & $N_{15}$ & $0.931$ & $22.5$ & $24.2$ & --- & --- & --- & --- & --- & --- & --- & --- \\
\skippoly & $N_{16}$ & $11.9$ & $245$ & $20.6$ & --- & --- & --- & --- & --- & --- & --- & --- \\
\skippoly & $N_{17}$ & $39.6$ & $734$ & $18.5$ & --- & --- & --- & --- & --- & --- & --- & --- \\
\skippoly & $N_{18}$ & $7.42$ & $157$ & $21.2$ & --- & --- & --- & --- & --- & --- & --- & --- \\
\skippoly & $N_{19}$ & $0.122$ & $3.12$ & $25.7$ & --- & --- & --- & --- & --- & --- & --- & --- \\
\skippoly & $N_{20}$ & $0.177$ & $6.00$ & $33.9$ & --- & --- & --- & --- & --- & --- & --- & --- \\
\skippoly & $N_{21}$ & $0.887$ & $12.6$ & $14.2$ & --- & --- & --- & --- & --- & --- & --- & --- \\
\skippoly & $N_{22}$ & $2.50$ & $7.96$ & $3.18$ & --- & --- & --- & --- & --- & --- & --- & --- \\
\skippoly & $N_{23}$ & $2.50$ & $7.60$ & $3.04$ & --- & --- & --- & --- & --- & --- & --- & --- \\
\skippoly & $N_{24}$ & $2.51$ & $7.65$ & $3.05$ & --- & --- & --- & --- & --- & --- & --- & --- \\
\skippoly & $N_{25}$ & $12.3$ & $17.7$ & $1.43$ & --- & --- & --- & --- & --- & --- & --- & --- \\
\skippoly & $N_{26}$ & $0.118$ & $4.40$ & $37.4$ & --- & --- & --- & --- & --- & --- & --- & --- \\
\skippoly & $N_{27}$ & $0.117$ & $4.43$ & $37.7$ & --- & --- & --- & --- & --- & --- & --- & --- \\
\skippoly & $N_{28}$ & $0.117$ & $4.41$ & $37.8$ & --- & --- & --- & --- & --- & --- & --- & --- \\
\skippoly & $N_{29}$ & $0.750$ & $9.87$ & $13.2$ & --- & --- & --- & --- & --- & --- & --- & --- \\
\skippoly & $N_{30}$ & $1.21$ & $26.7$ & $22.1$ & --- & --- & --- & --- & --- & --- & --- & --- \\
\skippoly & $N_{31}$ & $0.752$ & $9.84$ & $13.1$ & --- & --- & --- & --- & --- & --- & --- & --- \\
\skippoly & $N_{32}$ & $15.5$ & $318$ & $20.5$ & --- & --- & --- & --- & --- & --- & --- & --- \\
\skippoly & $N_{33}$ & $713$ & NC & NC & --- & --- & --- & --- & --- & --- & --- & --- \\
\skippoly & $N_{34}$ & $715$ & NC & NC & --- & --- & --- & --- & --- & --- & --- & --- \\
\skippoly & $N_{35}$ & $8{,}377$ & NC & NC & --- & --- & --- & --- & --- & --- & --- & --- \\
\skippoly & $N_{36}$ & $8{,}705$ & NC & NC & --- & --- & --- & --- & --- & --- & --- & --- \\
\skippoly & $N_{37}$ & $46{,}627$ & NC & NC & --- & --- & --- & --- & --- & --- & --- & --- \\
\skippoly & $N_{38}$ & $2{,}071$ & NC & NC & --- & --- & --- & --- & --- & --- & --- & --- \\
\skippoly & $N_{40}$ & $194{,}404$ & NC & NC & --- & --- & --- & --- & --- & --- & --- & --- \\
\skippoly & $N_{49}$ & $159$ & $4{,}926$ & $30.9$ & --- & --- & --- & --- & --- & --- & --- & --- \\
\skippoly & $N_{50}$ & $357$ & $14{,}583$ & $40.8$ & --- & --- & --- & --- & --- & --- & --- & --- \\
\skippoly & $N_{51}$ & $1{,}765$ & NC & NC & --- & --- & --- & --- & --- & --- & --- & --- \\
\skippoly & $N_{52}$ & $492$ & $21{,}015$ & $42.7$ & --- & --- & --- & --- & --- & --- & --- & --- \\
\skippoly & $N_{54}$ & $946{,}266$ & NC & NC & --- & --- & --- & --- & --- & --- & --- & --- \\
\skippoly & \textbf{Geomean} & $6.69$ & $18.6$ & $18.3$ & --- & --- & --- & --- & --- & --- & --- & --- \\
\midrule
\boxpoly & $N_{1}$ & $0.0396$ & $3.71$ & $94.8$ & $0.522$ & $13.3$ & --- & --- & $0.533$ & $13.6$ & $0.295$ & $7.52$ \\
\boxpoly & $N_{2}$ & $0.0429$ & $3.79$ & $90.1$ & $0.785$ & $18.6$ & --- & --- & $0.560$ & $13.3$ & $0.293$ & $6.94$ \\
\boxpoly & $N_{3}$ & $0.0742$ & $12.6$ & $171$ & $1.76$ & $23.8$ & --- & --- & $1.28$ & $17.4$ & $0.400$ & $5.42$ \\
\boxpoly & $N_{4}$ & $0.0785$ & $12.4$ & $163$ & $1.84$ & $24.1$ & --- & --- & $1.25$ & $16.4$ & $0.399$ & $5.23$ \\
\boxpoly & $N_{5}$ & $0.135$ & $28.9$ & $217$ & $2.90$ & $21.8$ & --- & --- & $2.27$ & $17.1$ & $0.565$ & $4.25$ \\
\boxpoly & $N_{6}$ & $0.116$ & $12.6$ & $109$ & $2.89$ & $25.1$ & --- & --- & $1.37$ & $11.9$ & $0.483$ & $4.19$ \\
\boxpoly & $N_{7}$ & $0.211$ & $28.6$ & $135$ & $4.66$ & $22.1$ & --- & --- & $2.51$ & $11.9$ & $0.840$ & $3.83$ \\
\boxpoly & $N_{8}$ & $0.602$ & $13.6$ & $22.7$ & $6.37$ & $10.6$ & --- & --- & $1.64$ & $2.73$ & $1.58$ & $2.61$ \\
\boxpoly & $N_{9}$ & $0.489$ & $13.9$ & $28.6$ & $5.97$ & $12.3$ & --- & --- & $1.67$ & $3.42$ & $1.23$ & $2.53$ \\
\boxpoly & $N_{10}$ & $0.449$ & $13.3$ & $29.8$ & $6.65$ & $14.8$ & --- & --- & $1.63$ & $3.63$ & $1.11$ & $2.46$ \\
\boxpoly & $N_{11}$ & $0.818$ & $5.47$ & $6.69$ & $5.53$ & $6.77$ & --- & --- & $1.38$ & $1.68$ & $1.63$ & $1.99$ \\
\boxpoly & $N_{12}$ & $0.551$ & $41.1$ & $74.8$ & $7.83$ & $14.2$ & --- & --- & $0.736$ & $1.34$ & $2.98$ & $5.39$ \\
\boxpoly & $N_{13}$ & $0.0895$ & $4.36$ & $48.8$ & $0.970$ & $10.8$ & --- & --- & $0.630$ & $7.05$ & $0.391$ & $4.36$ \\
\boxpoly & $N_{14}$ & $0.0770$ & $4.19$ & $54.7$ & $0.903$ & $11.6$ & --- & --- & $0.641$ & $8.37$ & $0.420$ & $5.37$ \\
\boxpoly & $N_{15}$ & $0.771$ & $41.7$ & $54.2$ & $8.95$ & $11.6$ & --- & --- & $0.987$ & $1.28$ & $5.24$ & $6.81$ \\
\boxpoly & $N_{16}$ & $1.89$ & $151$ & $80.3$ & $32.0$ & $17.0$ & --- & --- & $3.27$ & $1.74$ & $7.10$ & $3.75$ \\
\boxpoly & $N_{17}$ & $21.4$ & $219$ & $10.3$ & $68.2$ & $3.19$ & --- & --- & $30.2$ & $1.41$ & $57.3$ & $2.68$ \\
\boxpoly & $N_{18}$ & $5.05$ & $369$ & $73.1$ & $36.1$ & $7.14$ & --- & --- & NC & NC & $15.6$ & $3.09$ \\
\boxpoly & $N_{19}$ & $0.0875$ & $9.15$ & $105$ & $1.55$ & $17.8$ & --- & --- & $0.955$ & $11.0$ & $0.483$ & $5.51$ \\
\boxpoly & $N_{20}$ & $0.129$ & $16.7$ & $130$ & $2.36$ & $18.2$ & --- & --- & $1.51$ & $11.7$ & $0.545$ & $4.22$ \\
\boxpoly & $N_{21}$ & $0.428$ & $36.8$ & $86.0$ & $5.58$ & $13.0$ & --- & --- & $3.12$ & $7.29$ & $1.33$ & $3.06$ \\
\boxpoly & $N_{22}$ & $0.876$ & $14.8$ & $16.9$ & $7.33$ & $8.37$ & --- & --- & $1.75$ & $2.00$ & $1.84$ & $2.10$ \\
\boxpoly & $N_{23}$ & $1.21$ & $14.9$ & $12.3$ & $7.30$ & $6.04$ & --- & --- & $1.90$ & $1.57$ & $1.99$ & $1.64$ \\
\boxpoly & $N_{24}$ & $1.04$ & $15.5$ & $14.8$ & $6.97$ & $6.68$ & --- & --- & $1.98$ & $1.90$ & $1.91$ & $1.83$ \\
\boxpoly & $N_{25}$ & $6.22$ & $24.5$ & $3.94$ & $17.2$ & $2.76$ & --- & --- & $4.93$ & $0.792$ & $9.35$ & $1.50$ \\
\boxpoly & $N_{26}$ & $0.115$ & $4.36$ & $38.1$ & $1.08$ & $9.36$ & --- & --- & $0.655$ & $5.72$ & $0.478$ & $4.17$ \\
\boxpoly & $N_{27}$ & $0.128$ & $4.34$ & $33.8$ & $1.48$ & $11.5$ & --- & --- & $0.702$ & $5.48$ & $0.932$ & $7.26$ \\
\boxpoly & $N_{28}$ & $0.0998$ & $4.39$ & $44.0$ & $0.959$ & $9.61$ & --- & --- & $0.666$ & $6.68$ & $0.404$ & $4.05$ \\
\boxpoly & $N_{29}$ & $0.457$ & $5.87$ & $12.9$ & $2.01$ & $4.40$ & --- & --- & $0.778$ & $1.71$ & $1.55$ & $3.39$ \\
\boxpoly & $N_{30}$ & $1.17$ & $42.0$ & $35.9$ & $10.4$ & $8.86$ & --- & --- & $1.22$ & $1.04$ & $7.16$ & $6.10$ \\
\boxpoly & $N_{31}$ & $0.783$ & $5.87$ & $7.50$ & $16.3$ & $20.8$ & --- & --- & $0.934$ & $1.19$ & $14.8$ & $18.9$ \\
\boxpoly & $N_{32}$ & $4.52$ & $165$ & $37.2$ & $37.2$ & $8.36$ & --- & --- & $4.89$ & $1.10$ & $13.5$ & $3.04$ \\
\boxpoly & $N_{33}$ & $190$ & $10{,}189$ & $53.8$ & $831$ & $4.38$ & --- & --- & $194$ & $1.03$ & $550$ & $2.90$ \\
\boxpoly & $N_{34}$ & $556$ & $10{,}242$ & $18.5$ & $1{,}180$ & $2.12$ & --- & --- & $1{,}111$ & $2.00$ & $6{,}912$ & $12.4$ \\
\boxpoly & $N_{35}$ & $146$ & NC & NC & $419$ & $2.87$ & --- & --- & $218$ & $1.49$ & $423$ & $2.90$ \\
\boxpoly & $N_{36}$ & $5{,}011$ & NC & NC & $8{,}891$ & $1.77$ & --- & --- & $36{,}348$ & $7.25$ & NC & NC \\
\boxpoly & $N_{37}$ & $297$ & NC & NC & $763$ & $2.57$ & --- & --- & $412$ & $1.39$ & $743$ & $2.52$ \\
\boxpoly & $N_{38}$ & $84.3$ & NC & NC & $796$ & $8.66$ & --- & --- & $298$ & $3.53$ & $177$ & $2.10$ \\
\boxpoly & $N_{40}$ & $98{,}852$ & NC & NC & NC & NC & --- & --- & $663{,}940$ & $6.71$ & NC & NC \\
\boxpoly & $N_{49}$ & $50.2$ & $5{,}915$ & $118$ & $144$ & $2.87$ & --- & --- & $120$ & $2.36$ & $142$ & $2.82$ \\
\boxpoly & $N_{50}$ & $51.6$ & $13{,}204$ & $257$ & $164$ & $3.17$ & --- & --- & $97.9$ & $1.89$ & $111$ & $2.15$ \\
\boxpoly & $N_{51}$ & $86.2$ & NC & NC & $282$ & $3.05$ & --- & --- & $324$ & $3.77$ & $193$ & $2.25$ \\
\boxpoly & $N_{52}$ & $89.0$ & NC & NC & $509$ & $4.58$ & --- & --- & $211$ & $2.37$ & $212$ & $2.38$ \\
\boxpoly & $N_{54}$ & $509{,}415$ & NC & NC & NC & NC & --- & --- & NC & NC & NC & NC \\
\boxpoly & \textbf{Geomean} & $2.93$ & $33.6$ & $44.4$ & $14.5$ & $8.40$ & --- & --- & $7.67$ & $3.54$ & $5.20$ & $3.68$ \\
\midrule
\reusecert & $N_{1}$ & $0.0279$ & $1.08$ & $38.8$ & --- & --- & --- & --- & --- & --- & --- & --- \\
\reusecert & $N_{2}$ & $0.0327$ & $1.10$ & $33.6$ & --- & --- & --- & --- & --- & --- & --- & --- \\
\reusecert & $N_{3}$ & $0.0586$ & $3.16$ & $53.9$ & --- & --- & --- & --- & --- & --- & --- & --- \\
\reusecert & $N_{4}$ & $0.0588$ & $3.20$ & $54.4$ & --- & --- & --- & --- & --- & --- & --- & --- \\
\reusecert & $N_{5}$ & $0.0841$ & $7.00$ & $83.2$ & --- & --- & --- & --- & --- & --- & --- & --- \\
\reusecert & $N_{6}$ & $0.123$ & $3.16$ & $25.7$ & --- & --- & --- & --- & --- & --- & --- & --- \\
\reusecert & $N_{7}$ & $0.196$ & $7.63$ & $38.8$ & --- & --- & --- & --- & --- & --- & --- & --- \\
\reusecert & $N_{8}$ & $0.585$ & $4.03$ & $6.89$ & --- & --- & --- & --- & --- & --- & --- & --- \\
\reusecert & $N_{9}$ & $0.584$ & $4.45$ & $7.60$ & --- & --- & --- & --- & --- & --- & --- & --- \\
\reusecert & $N_{10}$ & $0.583$ & $4.03$ & $6.91$ & --- & --- & --- & --- & --- & --- & --- & --- \\
\reusecert & $N_{11}$ & $0.976$ & $2.05$ & $2.10$ & --- & --- & --- & --- & --- & --- & --- & --- \\
\reusecert & $N_{12}$ & $0.308$ & $14.8$ & $48.2$ & --- & --- & --- & --- & --- & --- & --- & --- \\
\reusecert & $N_{13}$ & $0.0596$ & $1.92$ & $32.2$ & --- & --- & --- & --- & --- & --- & --- & --- \\
\reusecert & $N_{14}$ & $0.0588$ & $1.95$ & $33.2$ & --- & --- & --- & --- & --- & --- & --- & --- \\
\reusecert & $N_{15}$ & $0.801$ & $15.0$ & $18.7$ & --- & --- & --- & --- & --- & --- & --- & --- \\
\reusecert & $N_{16}$ & $2.87$ & $46.2$ & $16.1$ & --- & --- & --- & --- & --- & --- & --- & --- \\
\reusecert & $N_{17}$ & $16.6$ & $66.1$ & $3.99$ & --- & --- & --- & --- & --- & --- & --- & --- \\
\reusecert & $N_{18}$ & $5.07$ & $96.8$ & $19.1$ & --- & --- & --- & --- & --- & --- & --- & --- \\
\reusecert & $N_{19}$ & $0.113$ & $2.28$ & $20.3$ & --- & --- & --- & --- & --- & --- & --- & --- \\
\reusecert & $N_{20}$ & $0.164$ & $4.20$ & $25.6$ & --- & --- & --- & --- & --- & --- & --- & --- \\
\reusecert & $N_{21}$ & $0.692$ & $9.40$ & $13.6$ & --- & --- & --- & --- & --- & --- & --- & --- \\
\reusecert & $N_{22}$ & $1.94$ & $4.92$ & $2.53$ & --- & --- & --- & --- & --- & --- & --- & --- \\
\reusecert & $N_{23}$ & $1.94$ & $5.05$ & $2.60$ & --- & --- & --- & --- & --- & --- & --- & --- \\
\reusecert & $N_{24}$ & $1.94$ & $5.07$ & $2.61$ & --- & --- & --- & --- & --- & --- & --- & --- \\
\reusecert & $N_{25}$ & $7.52$ & $11.3$ & $1.50$ & --- & --- & --- & --- & --- & --- & --- & --- \\
\reusecert & $N_{26}$ & $0.150$ & $3.56$ & $23.7$ & --- & --- & --- & --- & --- & --- & --- & --- \\
\reusecert & $N_{27}$ & $0.148$ & $3.57$ & $24.2$ & --- & --- & --- & --- & --- & --- & --- & --- \\
\reusecert & $N_{28}$ & $0.147$ & $3.54$ & $24.1$ & --- & --- & --- & --- & --- & --- & --- & --- \\
\reusecert & $N_{29}$ & $1.45$ & $6.92$ & $4.78$ & --- & --- & --- & --- & --- & --- & --- & --- \\
\reusecert & $N_{30}$ & $1.84$ & $17.0$ & $9.26$ & --- & --- & --- & --- & --- & --- & --- & --- \\
\reusecert & $N_{31}$ & $1.45$ & $6.96$ & $4.79$ & --- & --- & --- & --- & --- & --- & --- & --- \\
\reusecert & $N_{32}$ & $9.60$ & $72.7$ & $7.57$ & --- & --- & --- & --- & --- & --- & --- & --- \\
\reusecert & $N_{33}$ & $676$ & $3{,}604$ & $5.33$ & --- & --- & --- & --- & --- & --- & --- & --- \\
\reusecert & $N_{34}$ & $673$ & $3{,}648$ & $5.42$ & --- & --- & --- & --- & --- & --- & --- & --- \\
\reusecert & $N_{35}$ & $1{,}266$ & $8{,}573$ & $6.77$ & --- & --- & --- & --- & --- & --- & --- & --- \\
\reusecert & $N_{36}$ & $1{,}294$ & $8{,}561$ & $6.61$ & --- & --- & --- & --- & --- & --- & --- & --- \\
\reusecert & $N_{37}$ & $3{,}336$ & $40{,}146$ & $12.0$ & --- & --- & --- & --- & --- & --- & --- & --- \\
\reusecert & $N_{38}$ & $2{,}595$ & $4{,}891$ & $1.88$ & --- & --- & --- & --- & --- & --- & --- & --- \\
\reusecert & $N_{40}$ & $2{,}475$ & NC & NC & --- & --- & --- & --- & --- & --- & --- & --- \\
\reusecert & $N_{49}$ & $85.3$ & $1{,}921$ & $22.5$ & --- & --- & --- & --- & --- & --- & --- & --- \\
\reusecert & $N_{50}$ & $317$ & $4{,}971$ & $15.7$ & --- & --- & --- & --- & --- & --- & --- & --- \\
\reusecert & $N_{51}$ & $661$ & $5{,}321$ & $8.05$ & --- & --- & --- & --- & --- & --- & --- & --- \\
\reusecert & $N_{52}$ & $298$ & $6{,}618$ & $22.2$ & --- & --- & --- & --- & --- & --- & --- & --- \\
\reusecert & $N_{54}$ & NC & NC & NC & --- & --- & --- & --- & --- & --- & --- & --- \\
\reusecert & \textbf{Geomean} & $3.12$ & $32.0$ & $12.0$ & --- & --- & --- & --- & --- & --- & --- & --- \\
\midrule
\zid & $N_{1}$ & $0.0302$ & $1.42$ & $47.3$ & --- & --- & --- & --- & --- & --- & --- & --- \\
\zid & $N_{2}$ & $0.0306$ & $1.41$ & $46.2$ & --- & --- & --- & --- & --- & --- & --- & --- \\
\zid & $N_{3}$ & $0.0404$ & $3.14$ & $77.8$ & --- & --- & --- & --- & --- & --- & --- & --- \\
\zid & $N_{4}$ & $0.0401$ & $3.19$ & $79.8$ & --- & --- & --- & --- & --- & --- & --- & --- \\
\zid & $N_{5}$ & $0.0505$ & $5.49$ & $109$ & --- & --- & --- & --- & --- & --- & --- & --- \\
\zid & $N_{6}$ & $0.0430$ & $3.06$ & $71.2$ & --- & --- & --- & --- & --- & --- & --- & --- \\
\zid & $N_{7}$ & $0.0541$ & $5.54$ & $103$ & --- & --- & --- & --- & --- & --- & --- & --- \\
\zid & $N_{8}$ & $0.0975$ & $3.55$ & $36.5$ & --- & --- & --- & --- & --- & --- & --- & --- \\
\zid & $N_{9}$ & $0.0975$ & $3.51$ & $36.1$ & --- & --- & --- & --- & --- & --- & --- & --- \\
\zid & $N_{10}$ & $0.0980$ & $3.66$ & $37.4$ & --- & --- & --- & --- & --- & --- & --- & --- \\
\zid & $N_{11}$ & $0.272$ & $1.98$ & $7.28$ & --- & --- & --- & --- & --- & --- & --- & --- \\
\zid & $N_{12}$ & $0.375$ & $16.7$ & $44.4$ & --- & --- & --- & --- & --- & --- & --- & --- \\
\zid & $N_{13}$ & $0.0412$ & $2.00$ & $48.6$ & --- & --- & --- & --- & --- & --- & --- & --- \\
\zid & $N_{14}$ & $0.0406$ & $2.00$ & $49.1$ & --- & --- & --- & --- & --- & --- & --- & --- \\
\zid & $N_{15}$ & $0.557$ & $17.0$ & $30.5$ & --- & --- & --- & --- & --- & --- & --- & --- \\
\zid & $N_{16}$ & $0.757$ & $39.4$ & $51.9$ & --- & --- & --- & --- & --- & --- & --- & --- \\
\zid & $N_{17}$ & $7.40$ & $51.5$ & $6.95$ & --- & --- & --- & --- & --- & --- & --- & --- \\
\zid & $N_{18}$ & $3.22$ & $72.2$ & $22.4$ & --- & --- & --- & --- & --- & --- & --- & --- \\
\zid & $N_{19}$ & $0.0426$ & $2.49$ & $58.5$ & --- & --- & --- & --- & --- & --- & --- & --- \\
\zid & $N_{20}$ & $0.0501$ & $3.80$ & $76.0$ & --- & --- & --- & --- & --- & --- & --- & --- \\
\zid & $N_{21}$ & $0.0806$ & $6.40$ & $79.5$ & --- & --- & --- & --- & --- & --- & --- & --- \\
\zid & $N_{22}$ & $0.191$ & $3.75$ & $19.6$ & --- & --- & --- & --- & --- & --- & --- & --- \\
\zid & $N_{23}$ & $0.193$ & $3.92$ & $20.3$ & --- & --- & --- & --- & --- & --- & --- & --- \\
\zid & $N_{24}$ & $0.189$ & $3.85$ & $20.4$ & --- & --- & --- & --- & --- & --- & --- & --- \\
\zid & $N_{25}$ & $0.784$ & $4.78$ & $6.12$ & --- & --- & --- & --- & --- & --- & --- & --- \\
\zid & $N_{26}$ & $0.0662$ & $2.72$ & $41.1$ & --- & --- & --- & --- & --- & --- & --- & --- \\
\zid & $N_{27}$ & $0.0682$ & $2.77$ & $40.6$ & --- & --- & --- & --- & --- & --- & --- & --- \\
\zid & $N_{28}$ & $0.0662$ & $2.77$ & $41.9$ & --- & --- & --- & --- & --- & --- & --- & --- \\
\zid & $N_{29}$ & $0.129$ & $3.37$ & $26.1$ & --- & --- & --- & --- & --- & --- & --- & --- \\
\zid & $N_{30}$ & $0.574$ & $17.7$ & $30.8$ & --- & --- & --- & --- & --- & --- & --- & --- \\
\zid & $N_{31}$ & $0.130$ & $3.30$ & $25.4$ & --- & --- & --- & --- & --- & --- & --- & --- \\
\zid & $N_{32}$ & $0.885$ & $41.6$ & $46.9$ & --- & --- & --- & --- & --- & --- & --- & --- \\
\zid & $N_{33}$ & $123$ & $1{,}398$ & $11.4$ & --- & --- & --- & --- & --- & --- & --- & --- \\
\zid & $N_{34}$ & $120$ & $1{,}399$ & $11.7$ & --- & --- & --- & --- & --- & --- & --- & --- \\
\zid & $N_{35}$ & $87.8$ & $2{,}783$ & $31.7$ & --- & --- & --- & --- & --- & --- & --- & --- \\
\zid & $N_{36}$ & $88.2$ & $2{,}804$ & $31.8$ & --- & --- & --- & --- & --- & --- & --- & --- \\
\zid & $N_{37}$ & $130$ & $7{,}264$ & $55.9$ & --- & --- & --- & --- & --- & --- & --- & --- \\
\zid & $N_{38}$ & $42.3$ & $1{,}865$ & $44.3$ & --- & --- & --- & --- & --- & --- & --- & --- \\
\zid & $N_{40}$ & $156$ & $12{,}578$ & $80.7$ & --- & --- & --- & --- & --- & --- & --- & --- \\
\zid & $N_{49}$ & $14.0$ & $919$ & $66.6$ & --- & --- & --- & --- & --- & --- & --- & --- \\
\zid & $N_{50}$ & $27.1$ & $1{,}904$ & $70.8$ & --- & --- & --- & --- & --- & --- & --- & --- \\
\zid & $N_{51}$ & $35.6$ & $1{,}804$ & $50.9$ & --- & --- & --- & --- & --- & --- & --- & --- \\
\zid & $N_{52}$ & $45.0$ & $2{,}274$ & $51.1$ & --- & --- & --- & --- & --- & --- & --- & --- \\
\zid & $N_{54}$ & $2{,}521$ & $9{,}043$ & $3.59$ & --- & --- & --- & --- & --- & --- & --- & --- \\
\zid & \textbf{Geomean} & $0.811$ & $28.7$ & $35.4$ & --- & --- & --- & --- & --- & --- & --- & --- \\
\end{longtable}
\endgroup

\section{Specialization and Tensor Compilation}\label{app:specialization-ablation}
\begin{figure}[H]
\centering
\includegraphics[width=\linewidth]{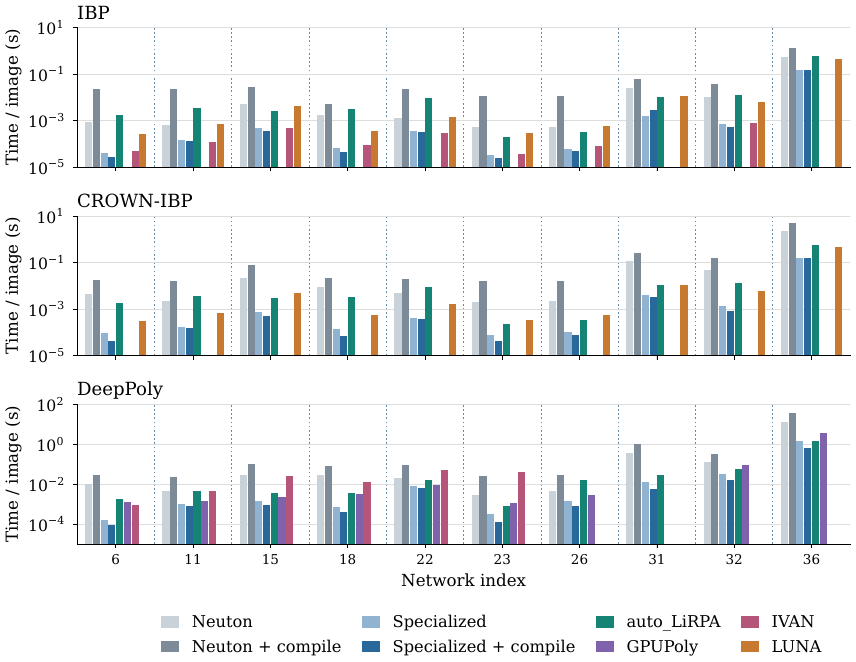}
\caption{Time per image, on a log scale, of Neuton's generic abstract interpreters and
\scepter's specialized abstract interpreters, each with and without \texttt{torch.compile}, along with
the baselines. Missing bars indicate baselines that do not support the analysis
or did not complete.}
\label{fig:specialization-ablation}
\end{figure}

To separate the effect of specialization from that of \texttt{torch.compile},
we compare Neuton and \scepter, each
with and without \texttt{torch.compile} on
ten networks. The analyses include IBP, CROWN-IBP, and \deeppoly. All other settings are as in
\S\ref{sec:spec-eval}. Fig.~\ref{fig:specialization-ablation} shows the time per
image of each implementation, together with the baselines that support each analysis.

\pheader{Specialization accounts for most of the speedup}
Specialization alone makes Neuton's interpreters $10.1\times$, $25.1\times$, and
$10.6\times$ faster for IBP, CROWN-IBP, and \deeppoly, respectively. Compiling
the specialized code adds only a further $1.15\times$, $1.45\times$, and
$1.80\times$.

\pheader{\scepter outperforms the baselines without \texttt{torch.compile}}
Without \texttt{torch.compile}, \scepter is faster than auto\_LiRPA, GPUPoly,
and IVAN, with geometric-mean speedups of up to $12.6\times$, $2.5\times$, and
$13.6\times$, respectively.

\pheader{\texttt{torch.compile} without specialization is not effective}
Compiling Neuton's generic interpreters with \texttt{torch.compile} makes them
at least $2.2\times$ slower on every configuration. Neuton interleaves structural 
and numerical computation (§2.x), so torch.compile attempts to trace both. Since the 
structural decisions are Python control flow, it cannot capture these decisions in a single graph. Hence, it
breaks each interpreter into hundreds of small graphs, with $436$--$1159$ graph breaks per network. Specialization
makes all structural decisions before verification and provides pure tensor computation
(\S\ref{sec:overview-solution}), so \texttt{torch.compile} sees the same code
for every image and compiles it without any graph breaks.

\end{document}